\documentclass{article}
\usepackage{amz_jfrr}
\usepackage{lipsum}  
\usepackage{todonotes}
\usepackage{url}
\usepackage{apalike}
\usepackage{setspace}
\usepackage{import}
\usepackage{booktabs}  
\usepackage{caption} 
\usepackage{pgfplots}
\usepackage{graphicx}
\usepackage{subcaption}
\usepgfplotslibrary{external}
\usetikzlibrary{pgfplots.groupplots,positioning}
\usepackage{tikz}
 
\usetikzlibrary{arrows.meta}

\pgfplotsset{
    compat=newest,
	tick label style={font=\small},
	label style={font=\small},
	legend style={font=\footnotesize},
    /pgfplots/legend image code/.code={%
        \draw[mark repeat=3,mark phase=2,#1]
            plot coordinates {
                (0cm,0cm)
                (0.2cm,0cm)
                (0.4cm,0cm)
            };
    },
}
\newlength\figureheight
\newlength\figurewidth
\usepackage{amsmath}
\usepackage{amssymb} 
\usepackage{bm} 
\usepackage{siunitx}
\usepackage{cleveref}

\usepackage{xcolor}
\usepackage[linesnumbered,ruled,vlined]{algorithm2e}

\SetCommentSty{mycommfont}

\SetKwInput{KwInput}{Input}                
\SetKwInput{KwOutput}{Output}              

\title{AMZ Driverless: The Full Autonomous Racing System}
\author{
\textbf{Juraj Kabzan}$^{\dag1}$, 
\textbf{Miguel I. Valls}$^{1}$, 
\textbf{Victor J.F. Reijgwart}$^{1}$, 
\textbf{Hubertus F.C. Hendrikx}$^{1}$,
\\ \textbf{Claas Ehmke}$^{1}$,
\textbf{Manish Prajapat}$^{1}$, 
\textbf{Andreas B{\"u}hler}$^{1}$, 
\textbf{Nikhil Gosala}$^{1}$, 
\textbf{Mehak Gupta}$^{1}$,
\\\textbf{Ramya Sivanesan}$^{1}$, 
\textbf{Ankit Dhall}$^{1}$, 
\textbf{Eugenio Chisari}$^{1}$, 
\textbf{Napat  Karnchanachari}$^{1}$, 
\textbf{Sonja Brits}$^{1}$,
\\ \textbf{Manuel Dangel}$^{1}$, 
\textbf{Inkyu Sa}$^{2}$, 
\textbf{Renaud Dub\'e}$^{2}$, 
\textbf{Abel Gawel}$^{2}$, 
\textbf{Mark Pfeiffer}$^{2}$, 
\\ \textbf{Alexander Liniger}$^{3}$,
\textbf{John Lygeros}$^{3}$ 
\textbf{and} \textbf{Roland Siegwart}$^{2}$\\
ETH Zurich\\
Zurich, Switzerland \\
\thanks{$^{\dag}$ Correspondence: \texttt{kabzanj@gmail.com}}
\thanks{$^{1}$ Authors are with AMZ Driverless, ETH Z\"urich.}%
\thanks{$^{2}$ Authors are with the Autonomous Systems Lab (ASL), ETH Z\"urich.}%
\thanks{$^{3}$ Authors are with the Automatic Control Laboratory (IfA), ETH Z\"urich.}%
}
\begin{document}

\maketitle

\begin{abstract}
This paper presents the algorithms and system architecture of an autonomous racecar.
The introduced vehicle is powered by a software stack designed for robustness, reliability, and extensibility. In order to autonomously race around a previously unknown track, the proposed solution combines state of the art techniques from different fields of robotics. Specifically, perception, estimation, and control are incorporated into one high-performance autonomous racecar. This complex robotic system, developed by AMZ Driverless and ETH Zurich, finished 1st overall at each competition we attended: Formula Student Germany 2017, Formula Student Italy 2018 and Formula Student Germany 2018. We discuss the findings and learnings from these competitions and present an experimental evaluation of each module of our solution.
\end{abstract}

\section*{Supplementary Material}

For a supplementary video visit: \url{https://www.youtube.com/watch?v=hYpToBIMpVQ} \newline
For open-source software and datasets visit: \url{https://github.com/AMZ-Driverless}\newline

\newpage
\section{Introduction}

Self-driving vehicles promise significantly improved safety, universal access, convenience, efficiency, and reduced costs compared to conventional vehicles. In order to fulfill the Society of Automation Engineers (SAE) level 4 autonomy\footnote{Taxonomy and Definitions for Terms Related to Driving Automation Systems for On-Road Motor Vehicles (J3016\_201609), \url{https://www.sae.org/standards/content/j3016_201609/}}, no driver attention must be required, even in emergency situations and under challenging weather conditions. Although a major part of autonomous driving on the public roads will happen in standard situations, a crucial aspect to reach full autonomy is the ability to operate a vehicle close to its limits of handling, i.e. in avoidance maneuvers or in case of slippery surfaces. 

\begin{figure}[ht]
    \centering
    \includegraphics[width=0.80\linewidth]{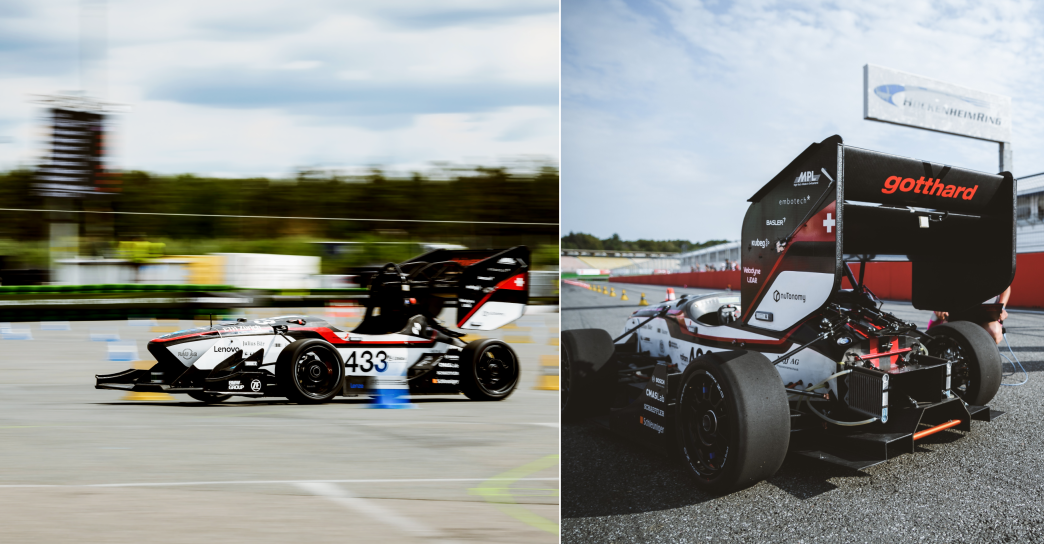}
    \caption{\emph{gotthard} driverless at the Formula Student Driverless competition which took place in August 2018 at the famous Hockenheimring, Germany.}
    \label{fig:gotthard_main}
\end{figure}

In autonomous car racing, vehicles race without a human in the loop as shown in \Cref{fig:gotthard_main}. Much alike traditional motorsports, autonomous racing provides a platform to develop and validate new technologies under challenging conditions. Self-driving racecars provide a unique opportunity to test software required in autonomous transport, such as redundant perception, failure detection, and control in challenging conditions. Testing such systems on closed tracks mitigates the risks of human injury.

This paper describes an entire autonomous racing platform, covering all required software modules reaching from environment perception to vehicle dynamics control. Starting with the perception pipeline, the developed system works using either LiDAR, vision or both.
Next, the motion estimation subsystem fuses measurements from five different sensors, while rejecting outliers in the sensor readings.
For localization and mapping, we utilize a particle filter-based SLAM system, facilitating the detection and association of landmarks. Subsequently, we propose a custom approach to plan paths which takes the most likely track boundaries into account, given a map and/or on-board perception. Lastly, we present a control framework that directly minimizes lap time while obeying the vehicle's traction limits and track boundary constraints.

The presented system has been thoroughly tested in context of the world's largest autonomous racing competition, Formula Student Driverless (described in \Cref{sec:main concept}). Our team, AMZ Driverless, secured the overall first place at each competition we participated in (see \Cref{tab:fs_leaderboards}). This demonstrates that the system can reliably race and finish on previously unknown tracks.

\begin{table}[ht]
\begin{center}
\begin{tabular}{@{}lllllllll@{}}
\toprule
\multicolumn{3}{c}{FS Germany 2017}            & \multicolumn{3}{c}{FS Italy 2018}              & \multicolumn{3}{c}{FS Germany 2018}            \\ \midrule\midrule
Team                & Fastest & Total & Team                & Fastest & Total & Team                & Fastest & Total \\
\textbf{Zurich ETH} & 55.53       & 557.89     & \textbf{Zurich ETH} & -           & 776.28     & \textbf{Zurich ETH} & 28.48       & 285.93     \\
Hamburg TU          & 105.44      & DNF        & Munchen UAS         & -           & 874.51     & Hamburg TU          & 48.93       & 581.24     \\
                    &             &            & Karlsruhe KIT       & -           & DNF        & Darmstadt TU        & 233.31      & 2336.60    \\
                    &             &            &                     &             &            & Karlsruhe KIT       & 124.79      & DNF        \\
                    &             &            &                     &             &            & Augsburg UAS        & 146.64      & DNF        \\
                    &             &            &                     &             &            & Munchen UAS         & 81.14       & DNF        \\
                    &             &            &                     &             &            & Trondheim NTNU      & 138.46      & DNF        \\ \bottomrule
\end{tabular}
\end{center}
\caption{Overview of the fastest lap and total time in seconds of each team that finished at least one lap, at each competition attended by AMZ Driverless. DNF is used when a team did not finish all ten laps. No official fastest lap times for FS Italy 2018 are available. The aim of every team is to achieve the lowest total time.}
\label{tab:fs_leaderboards}
\end{table}

This paper presents a revision of our previous works \cite{paper_fluela,cit:gotthard-icra-paper,keypoints_2019arXiv190202394D} and extends these with a description of the full autonomous system, including path planning (\Cref{sec:boundary_estimation}), control (\Cref{sec:control}) and the testing framework (\Cref{sec:testing_framework}). Furthermore, this paper provides a holistic evaluation of the system (\Cref{sec:overall_system_results}).

In summary, the contributions of this paper are;
 \begin{itemize}
    \item A complete autonomous racing system, covering all aspects of environment perception, state estimation, SLAM, planning, and vehicle dynamics control\footnote{For a video showing the entire pipeline in action visit \url{https://www.youtube.com/watch?v=hYpToBIMpVQ}}
    \item A demonstration of the system's capabilities on a full scale autonomous racecar, by reaching lateral accelerations of up to $1.5$g ($\SI{14.7}{\meter\per\square\second}$)
    \item A performance evaluation of each subsystem, after thorough testing in context of real-world autonomous racing competitions
    \item An open-source implementation of our automated simulation framework as well as an extensive data set for development and testing of autonomous racing applications \footnote{\url{https://github.com/AMZ-Driverless}}
\end{itemize}

The paper is organized as follows; \Cref{sec:related work} presents related studies for autonomous racing. The problem statement and overall concept are addressed in \Cref{sec:main concept}. Perception, motion estimation and mapping are described in \Cref{sec:perception} and \ref{sec:motion estimation and mapping} respectively and are followed by the vehicle control strategy in \Cref{sec:control}. Each of these 3 sections directly presents its associated real-world experimental results. Our simulation testing framework is presented in \Cref{sec:testing_framework}. In \Cref{sec:overall_system_results}, we provide an overview of the performance of the system as a whole and share our insights and suggestions for designing an autonomous racecar. Finally, this paper is concluded in \Cref{sec:conclusion}. 

\subsection*{Related Work}\label{sec:related work}

In 2004, 2005 and 2007, DARPA organized a series of Grand Challenges designed to radically speed up autonomous driving development. The first two challenges consisted of autonomous races and were extensively covered in the literature \cite{darpa_grand_challenge_editorial_part_1,darpa_grand_challenge_editorial_part_2,darpa_grand_challenge_editorial_part_3}. The system presented in this paper differs in that the car is designed to travel over a narrow unknown race track, whose boundaries are indicated only by cones.

After the DARPA challenges ended, the bulk of autonomous driving research shifted toward the industry. Yet several systems papers on autonomous racing are still being published, such as in~\cite{georgia_tech_autorally} describing the AutoRally platform. The autonomous system presented in this paper differs in that it runs on a full size, road racecar and has been tested at speeds exceeding $\SI{15}{\meter\per\second}$.

In the remaining paragraphs, a concise overview is provided of the state of the art for each sub-system presented in this paper.

Starting with perception, a system has been developed that can detect cones, estimate their 3D poses and colour using either LiDAR, vision or both. This follows from the fact that the tracks used to validate this paper's autonomous racecar are demarcated solely by colored cones and redundancy against single sensor failure was identified as a key requirement for robustness. This is explained in more detail in \Cref{sec:main concept}.

The LiDAR based perception system breaks the problem into a ground removal step similar to \cite{cit:adaptive-ground-removal}. It then uses Euclidean clustering and classification based on geometric priors to detect the cones and their poses, as presented in our first paper \cite{paper_fluela}. As a last step, the color patterns on sufficiently nearby cones can be identified using LiDAR intensity returns processed within a convolutional neural network (CNN). We presented this full pipeline in our previous paper \cite{cit:gotthard-icra-paper}. This paper therefore only includes a short summary for completeness and to provide context for further experimental results. 

To be able to drive through a track of cones, first they need to be detected on the image. Before the advent of large datasets and general purpose GPUs, classical computer vision techniques such as  \cite{viola_rapid_2001} for human face detection using smart, hand-crafted features and pedestrian detector \cite{dalal_histograms_2005} using histogram of oriented gradients (HoG) and support vector machines (SVM) were commonly used. Recently, CNNs such as the Single-shot detector \cite{liu_ssd_2016} and YOLO \cite{yolov1} have shown great promise in detecting objects more efficiently in a single pass of the image, unlike previous methods. In our work, we use a modified version of YOLOv2 \cite{yolov2_RedmonF16} for the purpose of cone detection in our pipeline. Further, to apply these detections for robotics based applications, it is essential to know the 3D position of these objects. CNNs have been proposed to predict view-points \cite{tulsiani_viewpoints_2015} for different objects; their work aims to capture the relation between keypoints on an object and their respective view-points. More recently, end-to-end pose estimation has been proposed where a 6 DoF pose is the output. In our work, instead of directly producing the pose of an object from a CNN model, we use a CNN for locating the keypoints in the image \cite{keypoints_2019arXiv190202394D} and subsequently use information about the cone's appearance and geometry to estimate it's pose via 2D-3D correspondences.

The Motion Estimation subsystem estimates the linear and angular velocities by fusing measurements from six different sensors. Several filter families exist, including the Extended Kalman Filter (EKF), Unscented Kalman Filter and Particle Filter \cite{Thrun:2005:PR:1121596}. An EKF was chosen given it's computational efficiency and accuracy when dealing with mildly non-linear systems perturbed by Gaussian noise \cite{cit:filters}. Outliers are rejected using the chi-squared test and a variance based drift detector. We previously discussed these outlier rejection approaches in \cite{paper_fluela} and \cite{cit:gotthard-icra-paper}. Consequently, this paper focuses on a more detailed description of the process and measurements models that we used, followed by more detailed experimental results.

A broad range of techniques exist to concurrently map an environment and localize a robot within it \cite{cadena_past_2016}. Since the race tracks in our experiments are initially unknown, the use of an online SLAM method is required. Many promising online graph-based SLAM methods have been proposed in the literature, such as \cite{lsd-slam:_2014,cartographer,orbslam2,rtab-map_2019}. However, the tracks we race on consist almost solely of colored cones placed atop wide open surfaces. The environment is thus sparse and comprised of landmarks that are distinguishable but that cannot be identified uniquely. This problem lends itself well to Particle Filter based SLAM systems such as FastSLAM \cite{Montemerlo2010}. This was illustrated in \cite{paper_fluela}, where we applied FastSLAM to autonomous racing and showed that it efficiently and reliably provides sufficiently accurate maps. We extended this work by switching to FastSLAM 2.0 \cite{Montemerlo2003} and by estimating cone colors in addition to their poses in \cite{cit:gotthard-icra-paper}. In this paper we therefore focus on providing more detailed experimental results and only briefly discusses the SLAM pipeline itself.

The planning problem statement for our autonomous racing use case differs from the general planning problem presented in the literature. Because the track is unknown when driving the first lap and the perception range is limited, there is no known goal position. Many classical search techniques such as those presented in \cite{Frazzoli} can therefore not be directly used. In \cite{tanzmeister2013road} a search technique is presented that does not require knowledge of a goal position and instead picks the best trajectory out of a set of feasible candidates. Since the environment in our problem is entirely built up of cones, we propose to discretize the space into a Delauney graph \cite{Delaunay1} by connecting only the middle points of each cone pair. Paths are sampled from this graph and the most likely track middle line is chosen based on a quadratic cost function that takes the path's geometry, the rules of the competition, and the uncertainty of cone colors and positions into account. This middle line is then tracked with a pure pursuit controller \cite{coulter1992implementation} to complete the first lap.

Once the first lap is completed, the SLAM subsystem provides a map including the boundaries of the entire track. Therefore, our focus lies in control techniques that are able to use this knowledge to push a car to its performance limits.
In the literature, several control techniques for autonomous racing are known. They can be roughly categorized into two categories, predictive \cite{liniger_scale_rc_cars_2015,verschueren2014towards,Funke_MPC,rosolia2017autonomous,Liniger2019viab} and non-predictive controllers \cite{zhang2001sliding,kritayakirana2012autonomous,klomp2014non}. Since non-predictive racing controllers are not able to plan the motion of the car, it is necessary to compute the racing line beforehand. An approach which was shown to work well is \cite{funke_up_2012}, where a set of clothoid curves was used to build a path. Alternatively, also methods from lap time optimization could be used \cite{casanova2000minimum,kelly2010time}. On the other hand, predictive racing controllers, are able to plan the optimal motion of the car by re-planning in a receding horizon fashion. Due to not requiring any pre-optimization and thereby being flexible with respect to the track we decided to use a predictive motion planning technique. More precisely we extended the real-time Model Predictive Control (MPC) method proposed in \cite{liniger_scale_rc_cars_2015}, which is able to directly find an optimal trajectory based on the performance envelope of the car. In this paper, we extend the method extensively to work with full-sized cars. The main difference is a novel vehicle model which is able to deal with zero velocity and includes the behavior of low-level controllers. 

\section{Main Concept}\label{sec:main concept}

The presented work was developed for the Formula Student Driverless competition. Therefore, the competition's rules, goals, and constraints lay the basis for the problem formulation and thus are detailed in the present section.

\subsection{Problem Statement -- The Race}\label{subsec:dynamic_discipline_description}

Recognizing the interest in autonomous driving and the associated technological challenges, Formula Student Germany organized the first driverless competition in 2017 followed by other countries in 2018. Formula Student (FS) is an international engineering competition, in which multidisciplinary student teams compete with a self-developed racecar every year. 

\begin{figure}[ht]
  \centering
  \includegraphics[width=0.7\textwidth]{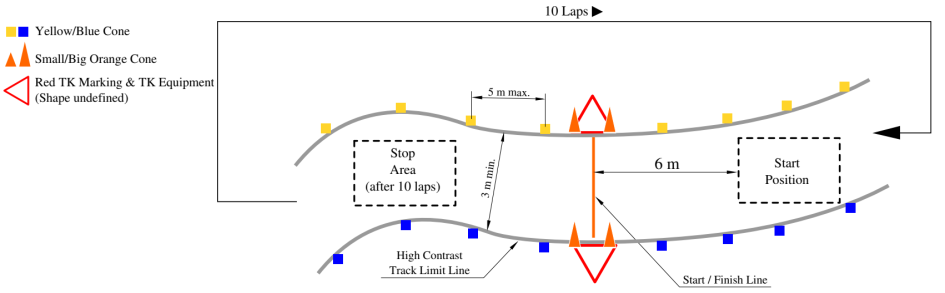}
  \caption{The track layout of a trackdrive discipline \cite{FSG_handbook}. Blue and yellow cones mark the track boundaries.}
  \label{fig:track_trackdrive}
\end{figure}

The main race, \textit{trackdrive}, consists of completing ten laps, as fast as possible, around an unknown track defined by small $228 \times 335$\,\si{mm} cones. Blue and yellow cones are used to distinguish the left and the right boundary respectively. The track is an up to $500$\si{m} long closed circuit, with a minimum track width of $3$\si{m} and the cones in the same boundary can be up to $5$\si{m} apart. By the rule book, this track contains straights, hairpins, chicanes, multiple turns, and decreasing radius turns among others. An exemplary track layout schematic is provided in \Cref{fig:track_trackdrive}.

It is important to note that the track is completely unknown before starting the race which increases the challenge considerably. In addition, all computations and sensing are required to happen on-board.

\subsection{Hardware Concept -- The Racecar}

The AMZ driverless racecar is an electric 4WD racecar with a full aerodynamic package, lightweight design and high drivetrain efficiency designed by AMZ\footnote{www.amzracing.ch} in 2016, see \Cref{fig:gotthard_main}. The car weighs only $190$ \si{\kilogram} and is able to accelerate from $0-100$\si{\km\per\hour} in $1.9$\si{\second}, faster than any road car or Formula1 car.

In order to make the vehicle race fully autonomously, sensors allowing for environment perception need to be added.
As all the perception algorithms, decision making and control will have to be executed on board, also additional computational units have to be added. 
The main system is required to be robust and reliable, which calls for redundancy in the pipeline. 
Therefore, two independent perception pipelines working with multiple sensor modalities were developed, allowing a robust operation of object detection (namely cones) and state estimation.

The first sensor modality is a 3D LiDAR sensor which is placed on the front wing. This maximizes the number of point returns per cone, and thus allows to perceive cones which are further away. A LiDAR was chosen over radars because of its ability to detect small non-metallic objects (such as plastic cones) and its high accuracy in depth. The choice of the LiDAR sensor is based on its physical parameters like the horizontal and vertical resolution and fields-of-view (FoV). In our application, the vertical resolution turns out to be the most important parameter since it limits the number of returns per cone which directly relates to the distance at which cones can be perceived.

\begin{figure}[ht]
    \centering	
	\begin{subfigure}{0.4\textwidth}
        \centering
        \includegraphics[width=0.8\textwidth]{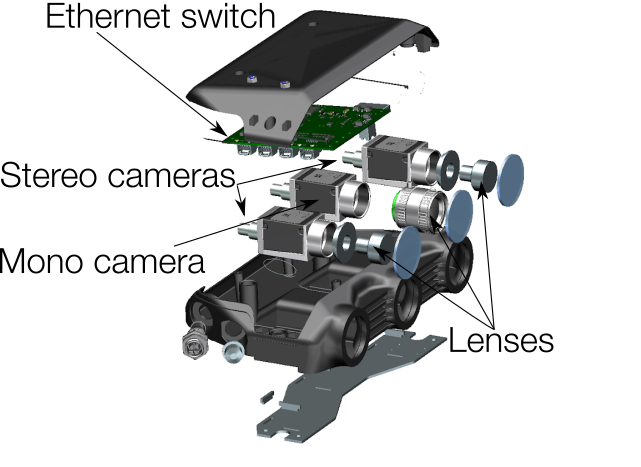}
        \caption{}
        \label{fig:explosion_cameras}
    \end{subfigure}
	\hspace{1cm}
	\begin{subfigure}{0.4\textwidth}
    	\centering
        \includegraphics[width=1.1\textwidth]{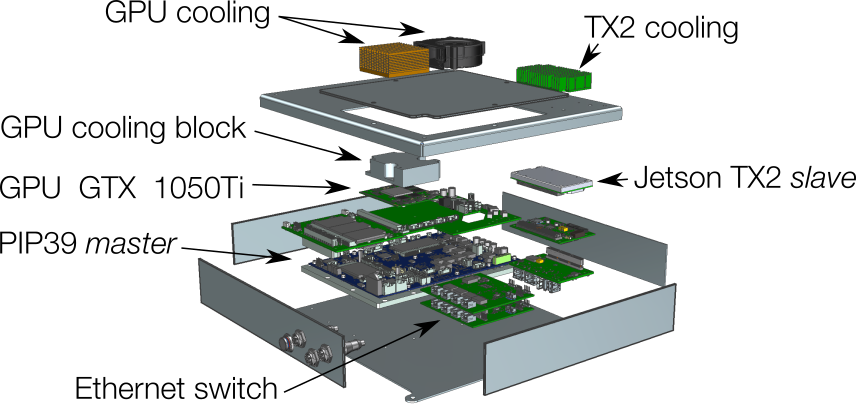}
        \caption{Our computing system consists of a \emph{master} computer, a \emph{slave} computer and an industrial ethernet switch.}
        \label{fig:cb_explosion}
    \end{subfigure}
    \caption{Self-developed camera and computing system housing.}
\end{figure}

\begin{figure}[ht]
	\centering
    \includegraphics[width=0.5\textwidth, angle=0]{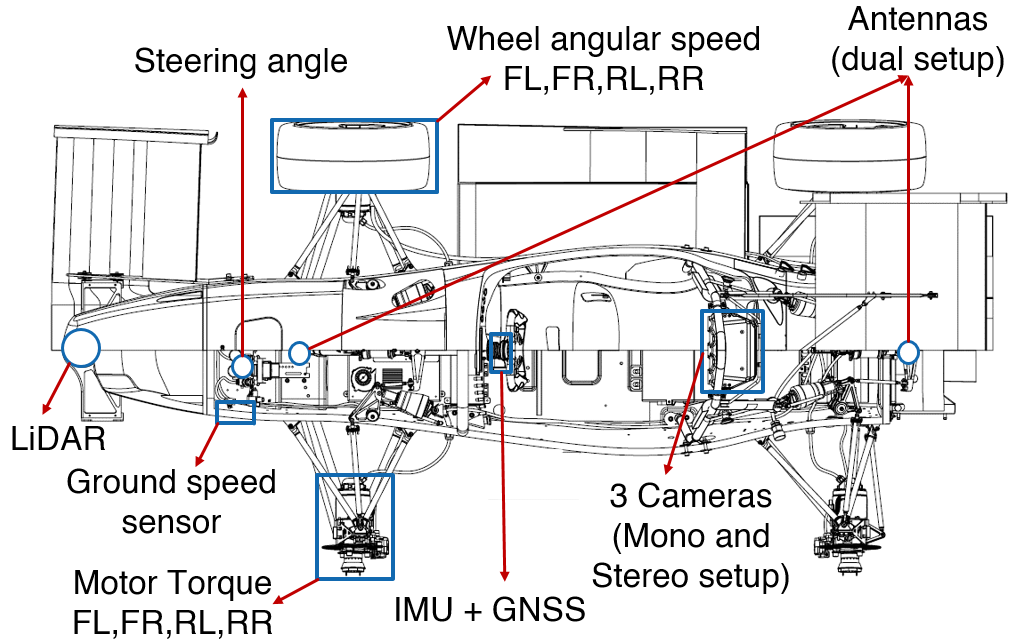}
    \caption{Location of all sensors providing data for motion estimation and perception.}
    \label{fig:VE_sensorplacement}
    \vspace{-0.5cm}
\end{figure}

The second sensor modality is image-based. Three CMOS~\cite{allen1987cmos} cameras with global shutter are used in a stereo and mono setup, see \Cref{fig:explosion_cameras}. The stereo setup is intended to deal with cones in close proximity whereas the mono camera is used for detecting cones further away.
To ensure that the position estimates of cones close to the car is highly accurate, the stereo setup uses lenses with short focal lengths (focal length of $5.5$\si{mm} and horizontal FoV of \ang{64.5} each). For accurately estimating the cone positions in the range of $7-15$\si{\meter}, a lens with focal length of $12$\si{mm} and horizontal FoV of \ang{54.5} was chosen for the monocular camera.
Considering the vantage point, the cameras were placed on the main roll hoop, above the driver's seat in the car, see \Cref{fig:VE_sensorplacement}. This offers the advantage that the occlusion among cones is reduced to a minimum and even the cones placed one behind the other (in line of sight) can be perceived sufficiently well.

Besides cameras and a LiDAR, which are used to determine the position of cones, the AMZ driverless racecar is equipped with additional sensors to estimate the car's velocity. The sensor setup is shown in ~\Cref{fig:VE_sensorplacement}.
Each of the four hub mounted motors incorporates a resolver, which provides smooth angular speed measurements. The motor controllers provide motor torques and the car is equipped with an industrial grade, laser based non-contact optical ground speed sensor (GSS). The steering angle is measured using an encoder placed in the steering rack. An accurate heading is crucial for localization and mapping of an autonomous vehicle, therefore, a dual antenna Inertial Navigation System (INS) is used which is capable of providing heading measurements with an accuracy of $0.08$\si{\degree}. The INS also includes an Inertial Measurement Unit (IMU).

Similar to the LiDAR-camera setup, this redundant sensor setup can cope with single sensor failures. The need for such redundancy  was confirmed throughout the testing seasons where many failures where observed.

\begin{figure}[ht]
    \centering
	\begin{subfigure}{0.4\textwidth}
	    \centering
        \includegraphics[width=0.4\textwidth]{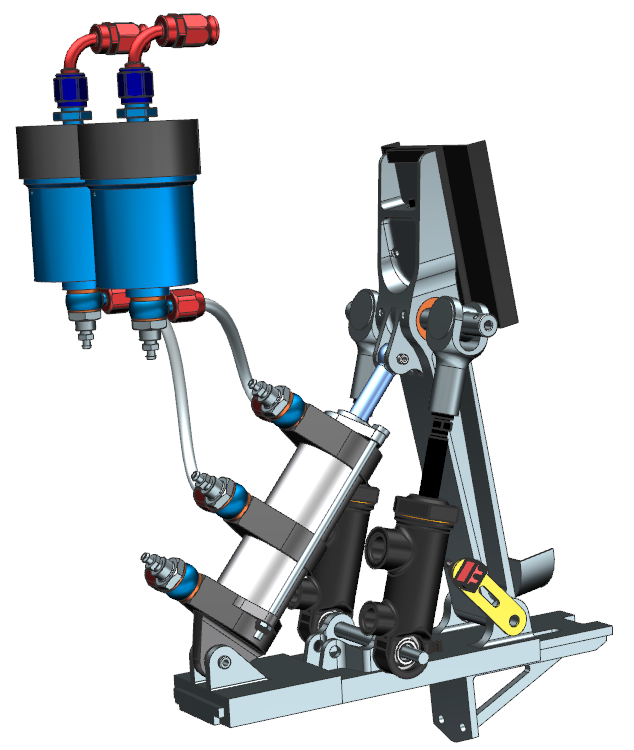}
        \caption{An Emergency Braking System (EBS) was added behind the braking pedal.}
        \label{fig:ebs}
    \end{subfigure}
	\hspace{1cm}
	\begin{subfigure}{0.5\textwidth}
	    \centering
        \includegraphics[width=0.7\textwidth]{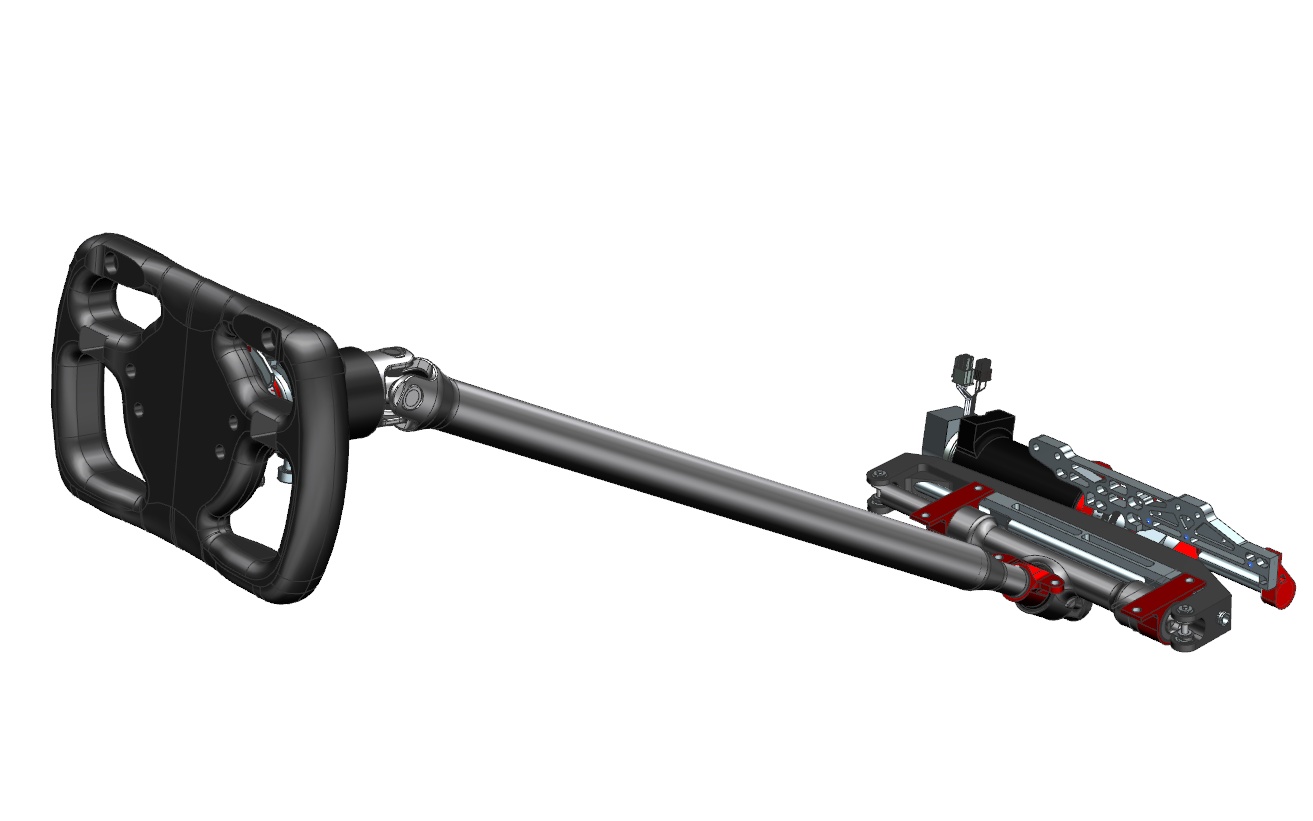}
        \caption{A steering motor was added below the axle. }
        \label{fig:steering}
    \end{subfigure}
  \caption{Actuators were installed to reproduce a human driver capabilities.}
\label{fig:ebs_steering}
\end{figure}

As a steering actuator, a motor was added under the front axle. This motor is able to steer from full left to full right in less than $0.9$\si{s}. The used autonomous racecar is capable of braking by recuperation through the motors. This generates a sufficient deceleration in order to race without a driver while not using the mechanical brakes actively. An Emergency Braking System (EBS) was installed behind the pedals as shown in \Cref{fig:ebs}. This system is only able to either fully brake or release the mechanical brakes and is used only in case of emergency. It can be triggered from either a Remote Emergency System (RES) or from an on-board computer.

To execute  the required computations, two computing units were installed (see \Cref{fig:cb_explosion}). Both units have a CPU and a GPU. One is a rugged industrial computer --- from here on called \emph{master} computer --- and the other is an efficient embedded unit called \emph{slave}. The hardware and software were designed such that in case of \textit{slave} failure (running only non-critical parts of the software pipeline), the system can still race. An Electronic Control Unit (ECU) is the only system that interacts with the actuators, also handling the low level controls and safety checks. In \Cref{fig:Main_Concept} one can see which parts of the pipeline run in which unit. 

\subsection{Software Concept}

To reach the full potential of the car, it is estimated that the track must be known for at least 2\si{s} ahead of the vehicle's current postion. This horizon, which at $80$\si{\km \per \hour} is above $40$\si{\m} is not achievable with the on-board sensor setup. Thus, a map of the track is built, such that once the track is known, its shape ahead can also be anticipated by estimating the car's pose in the map. This defines our two modes of operation, \textit{SLAM} mode and \textit{Localization} mode.

In \textit{SLAM} mode, the vehicle path has to be computed from the local perception and a map of the track has to be generated.  The main challenges are the limited perception horizon as the track has to be inferred from the perceived cones only. In \textit{Localization} mode, the goal is to race as fast as possible in the previously mapped track. 
Given the above requirements, due to the need to detect cones, map the track, and control the car to its full potential, the software stack is divided in three main modules, \textit{Perception}, \textit{Motion Estimation and Mapping} and \textit{Control}.

\textbf{Perception}: Its goal is to perceive cones and estimate their color and position as well as the associated uncertainties. This will be the input to the \textit{Motion Estimation and Mapping} module. To increase robustness, the perception system is designed to withstand any single failure and thus two fully independent pipelines are implemented at a hardware and software level, namely the LiDAR and camera pipelines. More details are provided in \Cref{sec:perception}.

\textbf{Motion Estimation and Mapping}: Its goal is to map the track, detect the boundaries and estimate the physical state of the car. The inputs are the observed cone positions from \textit{Perception} (camera and LiDAR pipeline) and the remaining sensors' measurements to estimate the velocity, see \Cref{sec:motion estimation and mapping}. 
For this purpose, three elements are needed. First, a velocity estimation that produces a fault tolerant estimate. Second, a SLAM algorithm, that uses cone position estimates from \textit{Perception} as landmarks, and the integrated velocity estimate as a motion model. Finally, and only for the \textit{SLAM} mode, a boundary estimation algorithm is used to classify cones as right or left boundary despite colour misclassification. Further insights are provided in \Cref{sec:motion estimation and mapping}.

\textbf{Control}: Its goal is to safely operate the car within the limits of handling while minimizing lap time. Taking the track boundaries and the state estimates as inputs, the control actions (steering, throttle and brake) are computed. Stability and robustness to external disturbances are important factors for safe operation. The low-level control distributes the torque among the four wheels, the safety system checks for consistency and unsafe states and is able to trigger the EBS. See \Cref{sec:control} for further details.

\begin{figure}[ht]
  \centering
  \includegraphics[width=0.5\textwidth]{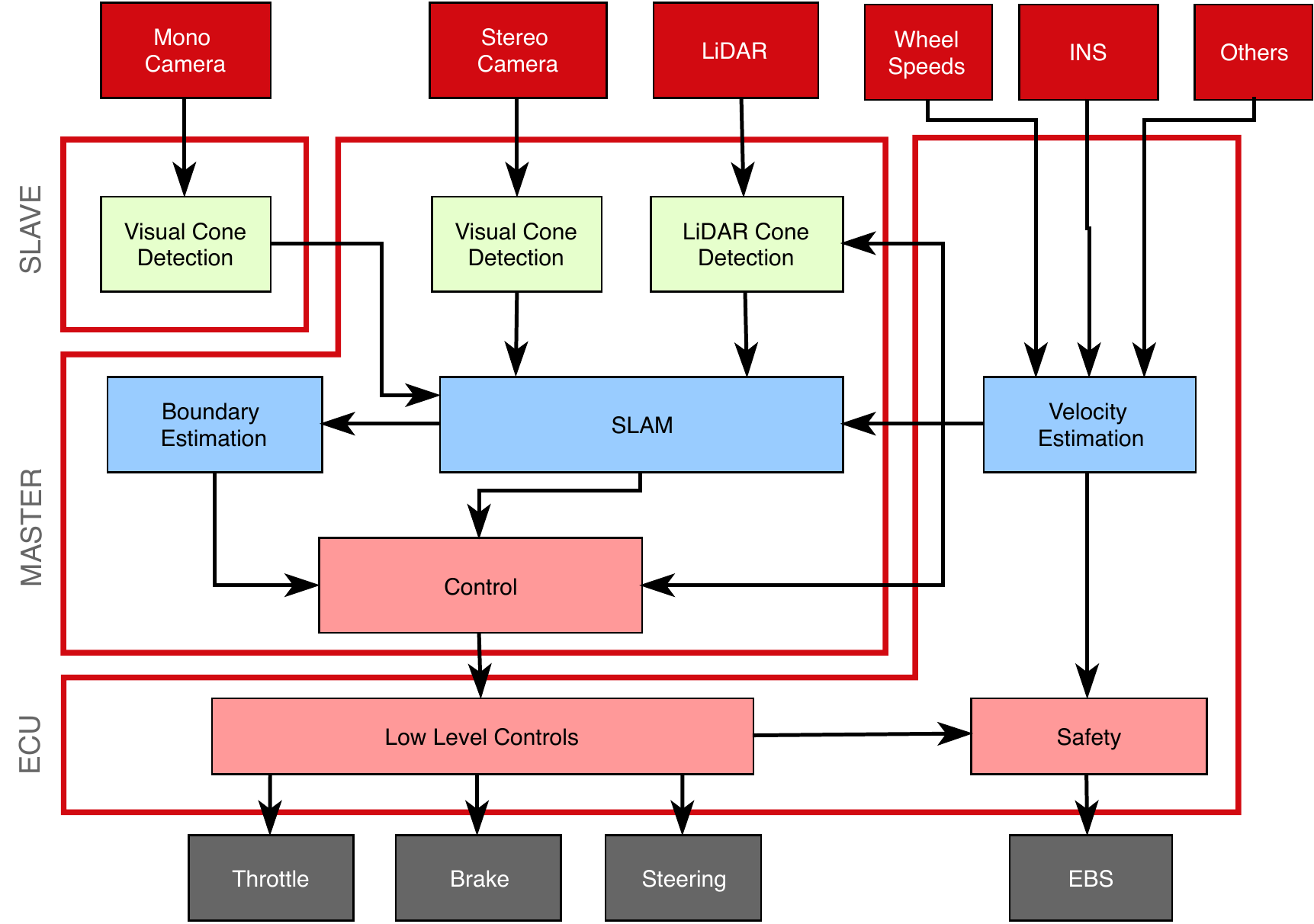}
  \caption{Software-hardware architecture of the autonomous system. From top to bottom the sensors are depicted in red, perception submodules in green, motion estimation and mapping submodules in blue, control submodules in light red and actuators in grey. The three red bounding boxes show the computing platform where the submodule is executed.  }
  \label{fig:Main_Concept}
\end{figure}

The software-hardware architecture of the system can be seen in \Cref{fig:Main_Concept}. Going from sensors to actuators passing through all software modules and the computing system they run on.

\section{Perception}\label{sec:perception}

The goal of the perception pipeline is to efficiently provide accurate cone position and color estimates as well as their uncertainties in real-time. Using these cone positions as landmarks, the \textit{SLAM} module builds a map aiding the car to navigate autonomously.
In order to perceive the environment reliably, a combination of sensors is used for redundancy and robustness. A LiDAR based pipeline is presented to detect cones based on its geometric information and a pattern recognition method based on LiDAR intensity returns to estimate color of the cones with a high confidence. In parallel, a camera-based multi-object detection algorithm is implemented alongside algorithms to accurately estimate the 3D positions of the cones, both from stereo and monocular images.

\subsection{LiDAR-based Sensing \& Estimation} \label{subsec:lidar_estimation}

We capitalize on the LiDAR information in two ways. First, we use the 3D information in the point cloud to detect cones on the track and find their position with respect to the car. Furthermore, we use the intensity data to differentiate between the various colored cones.

\subsubsection{The Cone Detection Algorithm}

\begin{figure}[ht]
    \centering
    \includegraphics[width=0.70\linewidth]{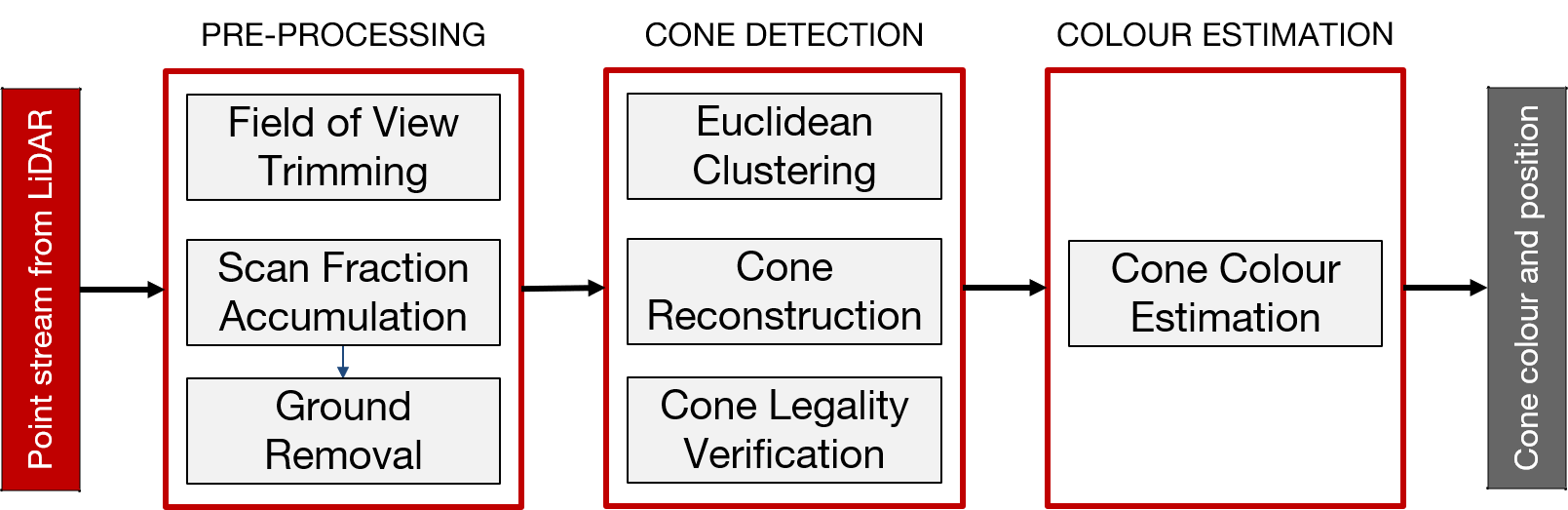}
    \setlength{\abovecaptionskip}{10pt}
    \caption{The LiDAR pipeline used to detect cones and estimate their color. The pipeline is composed of three major phases, namely \emph{Pre-processing}, \emph{Cone Detection}, and \emph{Color Estimation}. The pipeline accepts raw point clouds from the LiDAR sensor and outputs the location and color of the cones with respect to the reference frame of the car.}
    \label{fig:lidar-pipeline}
\end{figure}

The cones demarcating the race track are detected and classified using the pipeline shown in~\Cref{fig:lidar-pipeline}. The three major phases, pre-processing, cone detection, and color estimation make up the cone detection algorithm, which are explained next.

\subsubsection*{Pre-processing}

Due to the placement of the LiDAR sensor on the car, only cones in front of the car can be perceived, while the rear-view is occluded by the racecar. Thus the points behind the sensor are filtered out. The LiDAR sensor cannot inherently estimate motion which can lead to large distortions in the point cloud of a single scan and the appearance of \emph{ghost cones} if not accounted for. The scanned fractions are thus undistorted by using the velocity estimates of the car.

\begin{figure}[ht]
	\centering
    \begin{subfigure}{0.45\textwidth}
    	\centering
      	\includegraphics[width=0.8\linewidth]{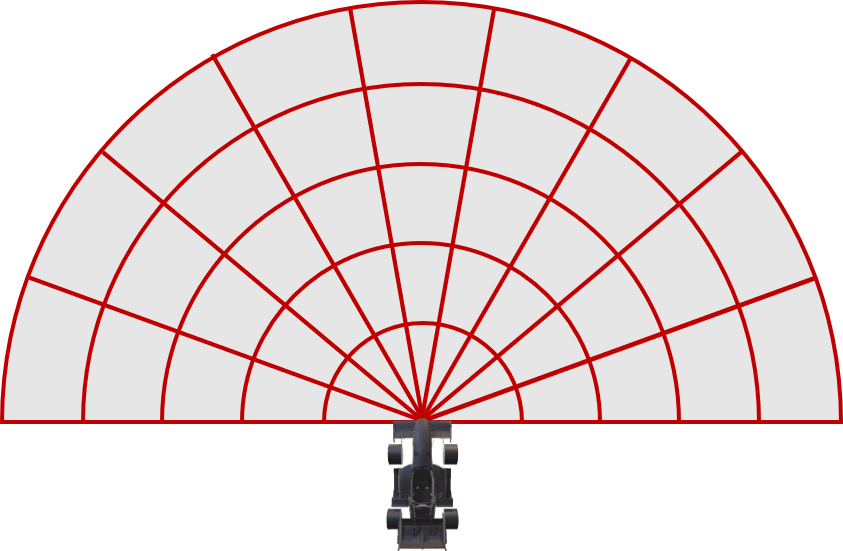}
        \caption{The FoV of the LiDAR is divided into multiple segments and bins.}
        \label{fig:lidar-ground-removal-top}
    \end{subfigure}
    \hspace{1cm}
	\begin{subfigure}{0.45\textwidth}
		\centering
      	\includegraphics[width=0.8\linewidth]{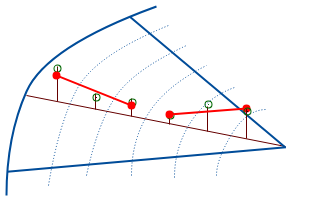}  
        \caption{Isometric view of the adaptive ground removal algorithm where the green circles represent the lowest LiDAR point in each bin and the red lines represent the estimated ground plane. The ground plane is regressed through the lowest points of all the bins in each segment and all points in the vicinity of these regressed ground lines are classified as ground and are subsequently removed.
        }
            \label{fig:lidar-ground-removal-side}
    \end{subfigure}
    \setlength{\abovecaptionskip}{10pt}
    \caption{View of LiDAR which is placed on top of the front wing.}
\end{figure}

An adaptive ground removal algorithm \cite{cit:adaptive-ground-removal} that adapts to changes in the inclination of the ground using a regression based approach is used to estimate the ground plane and distinguish between the ground and non-ground points, after which the ground points are discarded \cite{cit:gotthard-icra-paper}. The ground removal algorithm works by dividing the FoV of the LiDAR into angular segments and splitting each segment into radial bins (see \Cref{fig:lidar-ground-removal-top}). A line is then regressed through the lowermost points of all bins in a segment. All the points that are within a threshold distance to this line are classified as ground points and are removed (see \Cref{fig:lidar-ground-removal-side}).

\subsubsection*{Cone Detection}

\begin{figure}[ht]
    \centering
    \includegraphics[width=0.58\linewidth]{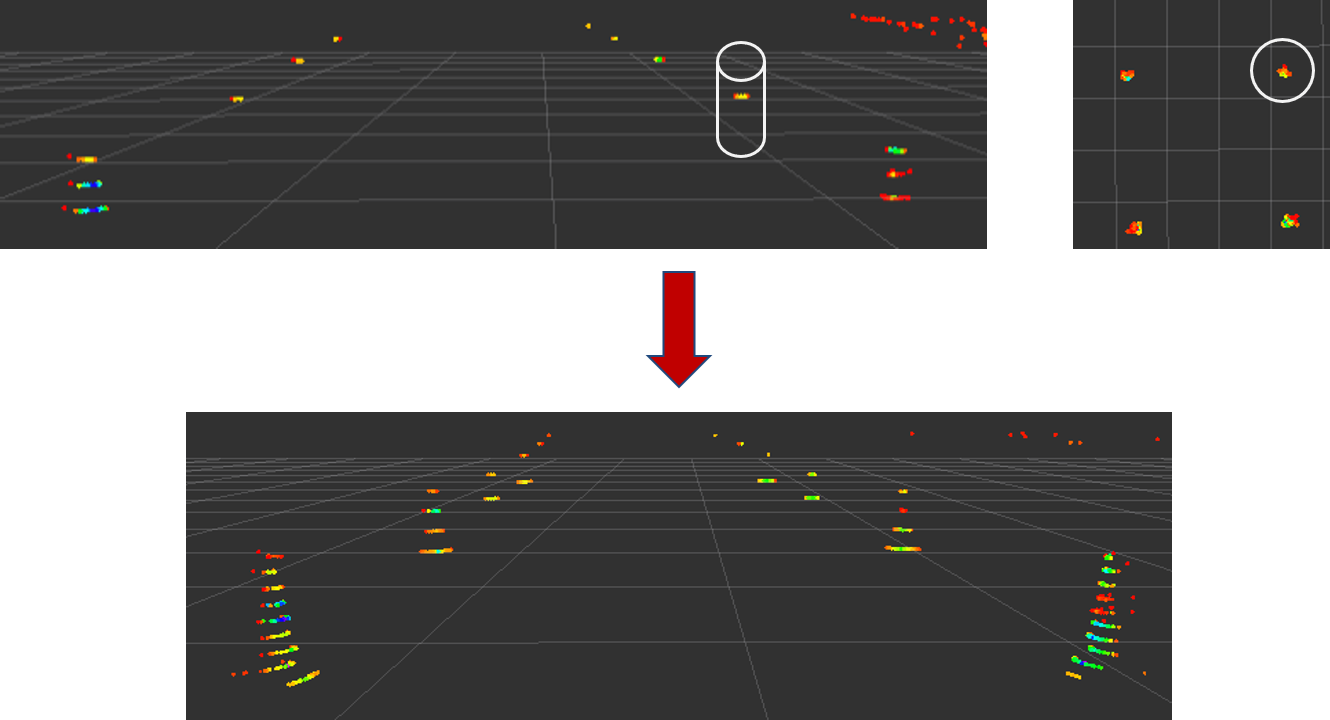}
    \caption{An illustration of the cone reconstruction process. The top-left image shows the LiDAR point cloud after executing the adaptive ground removal algorithm.  One can note the sparsity of the point cloud and the dearth of points in a cone. The cones are reconstructed by retrieving a cylindrical area around each cluster center (top-right) resulting in the bottom image wherein the presence of cones is more evident.}
    \label{fig:lidar-cone-reconstruction}
\end{figure}

The aforementioned ground removal algorithm removes a substantial amount of cone points in addition to those of the ground. This significantly reduces the already small number of return points that can be used to detect and identify cones.
This is addressed by first clustering the point cloud after ground removal using  Euclidean-distance based clustering algorithm and then reconstructing a cylindrical area around each cluster center using points from the point cloud before ground removal. This recovers most of the falsely removed points improving cone detection and color estimation (see \Cref{fig:lidar-cone-reconstruction}). The reconstructed clusters are then passed through a rule-based filter that checks whether the number of points in that cluster is in accordance with the expected number of points in a cone at that distance computed with
\begin{align}
\label{eq:lidar-cone-filter}
E(d) = \Bigg(\frac{1}{2} \frac{h_{c}}{2 d \tan(\frac{r_v}{2})} \frac{w_{c}}{2 d \tan(\frac{r_h}{2})}\Bigg) \, ,
\end{align}
where $d$ represent distance, $h_{c}$ and $w_{c}$ are the height and width of the cone respectively, and $r_v$ and $r_h$ are the vertical and horizontal angular resolutions of the LiDAR respectively.
The clusters that successfully pass through this filter are considered to be cones and are forwarded to the color estimation module.

\subsubsection*{Cone Pattern Estimation}

\begin{figure}[ht]
    \centering
    \includegraphics[width=0.80\linewidth]{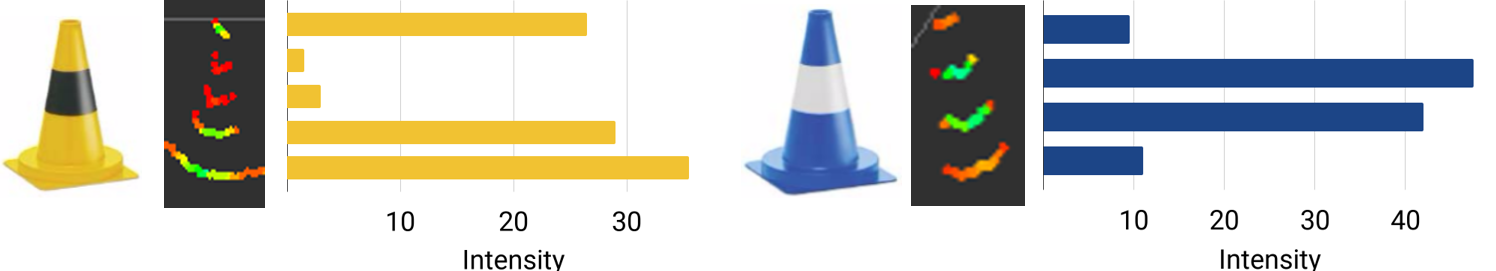}
    \caption{The intensity gradients for the pre-defined yellow and blue cones along with their point cloud returns. When reading the intensity values along the vertical axis, one denotes \textit{high-low-high} and \textit{low-high-low} patterns for the yellow and blue cones respectively. These differing intensity patterns are used to differentiate between yellow and blue cones.}
    \label{fig:lidar-cone-intensity-gradient}
\end{figure}

\begin{figure}[ht]
    \centering
    \includegraphics[width=0.80\linewidth]{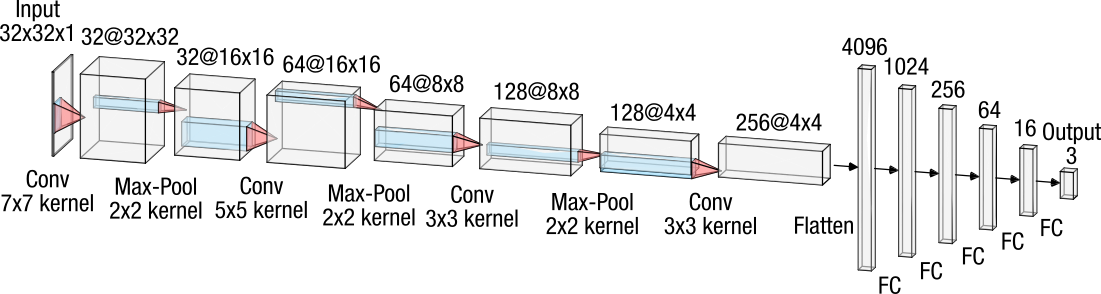}
    \caption{An illustration of our CNN architecture for estimating the color of the  cones which is composed of four convolutional and five fully-connected layers. The networks accepts a 32x32 grayscale image and outputs the probability of the cone belonging to each of the three classes, namely \textit{blue}, \textit{yellow}, and \textit{unknown}.}
    \label{fig:lidar-cnn-architecture}
\end{figure}

According to the rule book \cite{FSG_handbook}, a yellow cone has a \textit{yellow-black-yellow} pattern whereas a blue cone has a \textit{blue-white-blue} pattern which results in differing LiDAR intensity pattern as one moves along the vertical axis of the cone as shown in Figure \ref{fig:lidar-cone-intensity-gradient}. This intensity gradient pattern is leveraged in the cone color estimation using a CNN architecture. The CNN consists of four convolutional and five fully connected layers as shown in \Cref{fig:lidar-cnn-architecture}.
The network accepts a $32\times 32$ grayscale image of the cone as input and outputs the probability of the cone being \textit{blue}, \textit{yellow}, and \textit{unknown}. The input image is created by mapping the 3D bounding box of the cone cluster to a $32\times 32$ image where the cluster center is mapped to the center of the image and all the other points are appropriately scaled to fit in it. The pixel values store the intensities of points in the point cloud which are then scaled by a constant factor to increase the disparity between the intensities of various layers and make the difference more apparent. The CNN uses an asymmetric cross-entropy loss function that penalizes misclassifications (eg. blue cone classified as yellow cone) much more severely than incorrect classifications (eg. blue cone classified as unknown) which results in the network being more cautious towards ambiguous inputs and reduces the overall probability of misclassifying the color of cones. Furthermore, dropout and batch-normalization are used to prevent complex co-adaptations between neurons, control over-fitting, and improve the generalization of the network. However, the sparsity of the point cloud combined with the small size of the cones limits the color estimation range to \SI{4.5}{m}, as cone returns beyond this distance do not allow for a distinction between the differing intensity patterns.

\subsection{Camera-based Sensing \& Estimation} \label{subsec:camera_estimation}

To make the perception pipeline robust to sensor failure, a camera based sensing and estimation pipeline is deployed in parallel and independent of the LiDAR one. Considering that the vision pipeline is  developed for a real time system, it becomes crucial to detect and estimate color and positions of multiple cones with as little latency as possible and with minimum utilization of computational resources. Hence, this section proposes a real-time computer vision pipeline to detect and estimate 3D position using a single image (monocular pipeline) and a pair of images (stereo pipeline). 

\subsubsection{The Cone Detection Algorithm}

\begin{figure}[ht]
    \centering
    \includegraphics[width=0.5\textwidth]{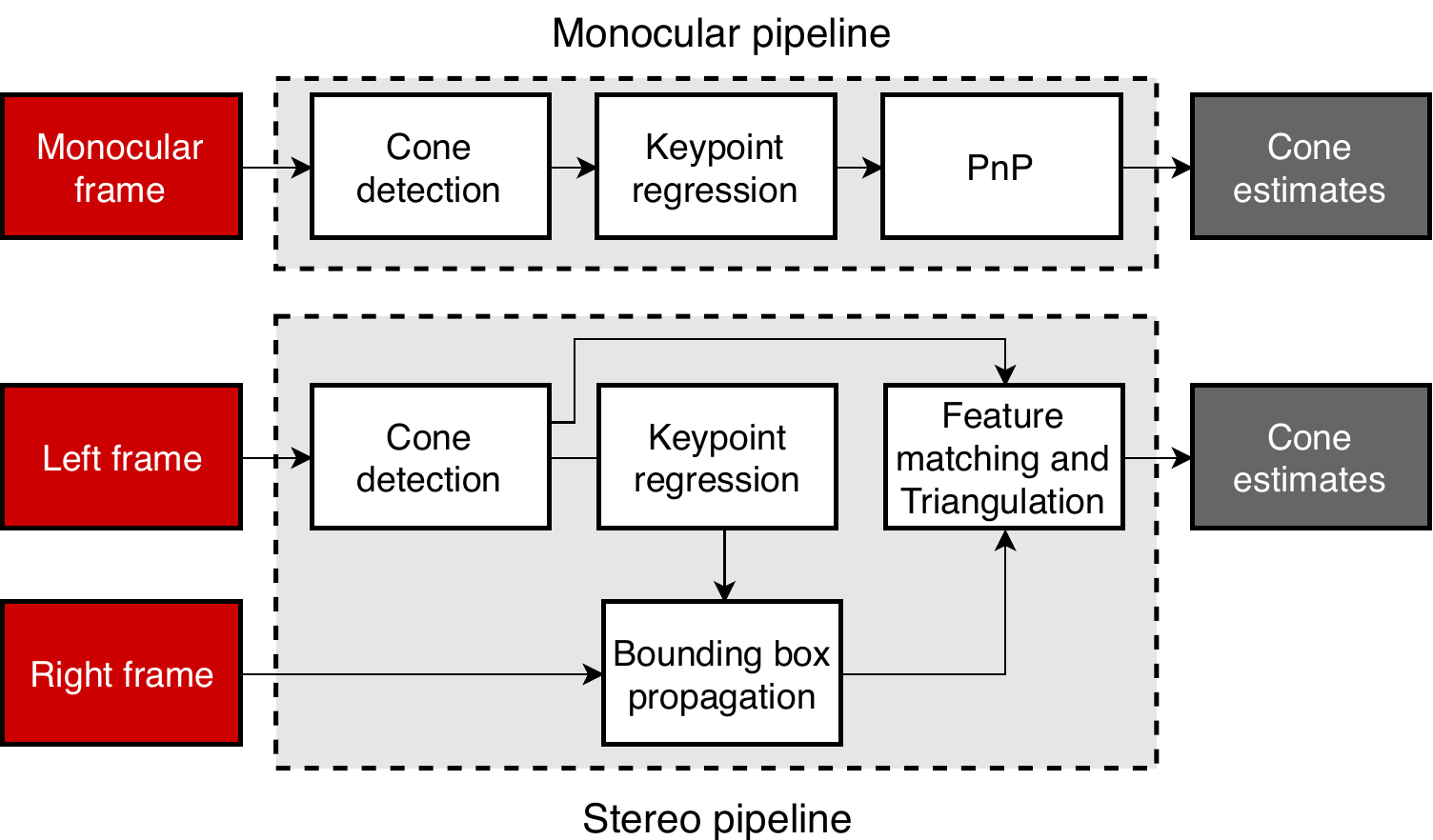}
    \caption{Pipeline of cone detection and position estimation using monocular and stereo cameras.}
    \label{fig:vision-pipeline-arch}
\end{figure}

Cones on the track are detected and their positions are estimated using the pipeline shown in \Cref{fig:vision-pipeline-arch}. The two pipelines have three major phases, multiple object detection Neural Network (NN), key-point regression and pose estimation.  For the monocular pipeline, keypoints are detected in a bounding box using ``keypoint regression''. The 3D pose of a cone is then estimated via the Perspective-n-Point (PnP) algorithm. For the stereo pipeline, features are matched in corresponding bounding boxes and triangulated to get an 3D position estimates of cones. 

\subsubsection{Multiple Object Detection}
\label{subsubsection:object-detection}

To estimate the 3D positions of multiple cones from a single image, it is necessary to detect these cones and classify them as blue, yellow or orange. A real-time and powerful object detector in the form of YOLOv2\cite{yolov2_RedmonF16} was trained on these three different types of cones. On being fed with the acquired images, this YOLOv2 returns the bounding box positions around the detected cones along with the confidence scores for each detection. This object detector network was chosen due to its robust outputs along with the ability to be fine-tuned with less data (using the ImageNet dataset \cite{imagenet}) and due to the existence of pre-trained weights which act as good priors during training. 

\subsubsection{Keypoint Regression}

\begin{figure}[h]
    \centering
    \includegraphics[width=0.35\textwidth]{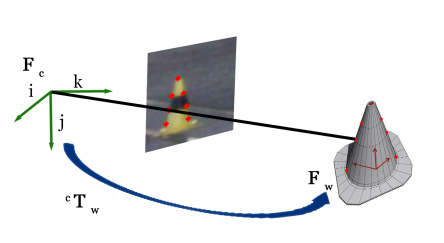}
    \caption{Schematic illustrating the matching of a 2D-3D correspondence and estimation of a transformation between the camera and world frames.}
    \label{fig:cone_pnp}
\end{figure}

Next, one needs to retrieve the 3D positions from objects on the image. This in itself is not solvable using a single image, because of ambiguities due to scale and limited information. The ill-posed problem can be solved by leveraging additional prior geometric information of the objects along with the 2D information obtained from the image. A ``keypoint regression'' scheme is used, that exploits this prior information about the object's shape and size to regress and find specific feature points on the image that match their 3D correspondences whose locations can be measured from a frame of reference $\mathcal{F}_w$ \cite{keypoints_2019arXiv190202394D} as shown in \Cref{fig:cone_pnp}.

\subsubsection*{From Patches to Features: Classical Computer Vision Techniques}

\begin{figure}[ht]
    \centering
    \begin{subfigure}{0.15\textwidth}
		\centering
        \includegraphics[scale=0.6]{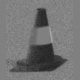}
        \caption{}
    \end{subfigure}
    \begin{subfigure}{0.15\textwidth}
		\centering
        \includegraphics[scale=0.6]{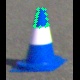}
        \caption{}
    \end{subfigure}
    \begin{subfigure}{0.15\textwidth}
		\centering
        \includegraphics[scale=0.6]{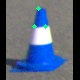}
        \caption{}
    \end{subfigure}
    \begin{subfigure}{0.15\textwidth}
		\centering
      \includegraphics[scale=0.6]{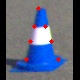}
        \caption{}
    \end{subfigure}
    \begin{subfigure}{0.15\textwidth}
		\centering
        \includegraphics[scale=0.6]{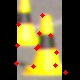}
        \caption{}
    \end{subfigure}
    \caption{Classical computer vision approaches for keypoint extraction. (a) ``b'' channel of LAB color space (b) Contour fitting on upper triangle (c) 3 defining vertices of cone (d) Extracted keypoints (e) Erroneous output when multiple cones are present.}
    \label{fig:stereo_cv_keypoint}
\end{figure}

 Initially, classical computer vision approaches were explored to extract these keypoints. Firstly, the bounding box is converted from the RGB color space to the LAB color space. In the LAB color space, the ``L'' channel represents the lightness whereas the ``a'' and ``b'' channel represent the green-red and blue-yellow components of the image respectively. In this color space, the ``b'' channel of the bounding box is extracted. An ``Otsu'' threshold is then applied on this transformed bounding box to obtain a binary image. Three vertices of the triangle were identified by performing contour fitting on the upper triangle of the binary image. Exploiting the 3D prior information of the cones, these three vertices were extended to obtain further four more keypoints. \Cref{fig:stereo_cv_keypoint} shows the extracted seven keypoints. The last image in \Cref{fig:stereo_cv_keypoint} shows the erroneous keypoints obtained in an edge case scenario obtained using this approach. Therefore, to make the keypoint extraction robust in such cases, the neural network described below is used. 

\subsubsection*{From Patches to Features: Keypoint Regression}

\begin{figure}[ht]
    \centering
	\begin{subfigure}{0.3\textwidth}
        \centering
        \includegraphics[width=1.3\textwidth]{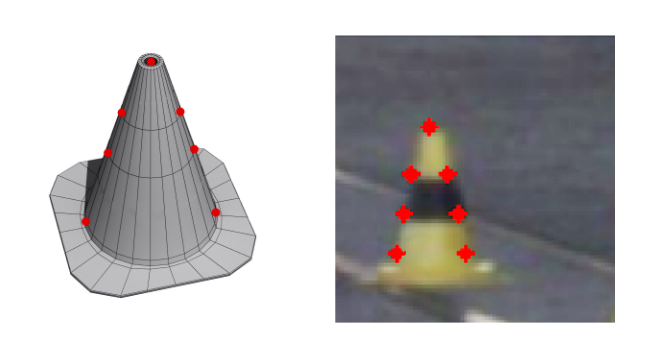}
        \caption{Regressed keypoints are represented using red markers.}
        \label{fig:model_and_kp}
    \end{subfigure}
	\hspace{1cm}
	\begin{subfigure}{0.6\textwidth}
        \centering
        \includegraphics[width=0.9\textwidth]{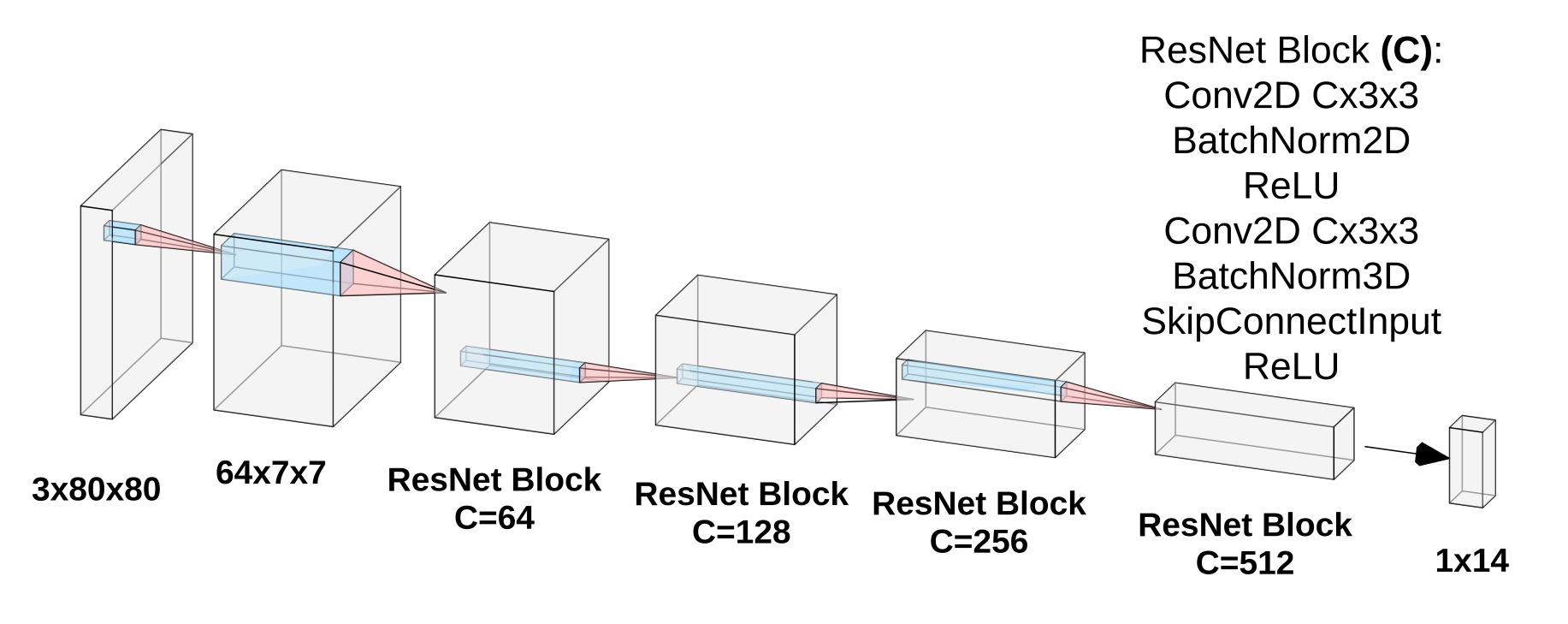}
        \caption{The ``keypoint network'' \cite{keypoints_2019arXiv190202394D} takes a sub-image patch of $80 \times 80 \times 3$ as input and maps it to coordinates of the keypoints.}
        \label{fig:network_arch}
    \end{subfigure}
  \caption{Keypoints are regressed from a detected box with a custom CNN and used to estimate the 3D position of a cone.}
\label{fig:keypoint_network}
\end{figure}

The customized CNN \cite{keypoints_2019arXiv190202394D} detects ``corner-like'' features given a $80 \times 80$ image patch input. \Cref{fig:model_and_kp}, shows the keypoints in 3D and 2D. We measure their 3D location relative to the base of the cone ($\mathcal{F}_w$). The keypoints are chosen such that they have a similar flavor of ``corners-like'' features from classical computer vision. The patches passed for the keypoint regression are bounding boxes detected by the object detector in the previous sub-module. The output vector is interpreted as the $(X, Y)$ coordinates of the keypoints for a given input patch. A part of ResNet \cite{he2016deep} is used as the backbone. The architecture applies \textit{same} convolutions using a $3 \times 3$ kernel via a residual block (see PyTorch \cite{pytorch} for more details). Using residual blocks reduces the chance of overfitting as they are capable of learning an identity transformation. The input tensor is passed through a convolution layer (with batch-norm) and non-linearity in the form of rectified linear units (ReLU). Batch-normalization makes the network robust to bad initialization. Following the batch-norm block, the signal passes through the following residual blocks $C= 64$, $C = 128$, $C = 256$ and $C = 512$ (refer to \Cref{fig:network_arch}). A final fully-connected linear layer predicts the location of the keypoints. 

With more convolution layers, the output feature tensor has more channels and smaller spatial dimensions. The final feature maps contain more global information than local information \cite{unet}. For regression, however, location specific information is essential. Using a ResNet-based architecture prevents drastic reduction of spatial resolution as the input volume is processed deeper into the network.

\subsubsection{Monocular 3D Position Estimation}\label{subsubsec:mono-3d}

The camera frame is defined as $\mathcal{F}_c$ and the world frame as $\mathcal{F}_w$. We choose $\mathcal{F}_w$ to be at the base of the cone, for convenience of calculation of the 3D location of the keypoints.

Detected keypoints are used as 2D points on the image to make correspondence with the respective points on the 3D model of a cone. Using the correspondences and the camera intrinsics, PnP is used to estimate the pose of every detected cone. This works by estimating the transformation $^{c}\mathcal{T}_w$ between the camera coordinate system and the world coordinate system. Since, the world coordinate system is located at the base of the cone, lying at an arbitrary location in $\mathbb{R}^3$ this transformation is exactly the pose that needs to be estimated. 

To estimate the position of the cone accurately, the non-linear version of PnP is used to obtain the transformation. RANSAC PnP is used instead of vanilla PnP, for the occasional incorrect or noisy correspondences. PnP is performed on the keypoint regressor output and the pre-computed 3D correspondences. Using the above pipeline, the pose transformation of multiple objects can be estimated using just a single image.

\subsubsection{Stereo 3D Position Estimation}

The pipeline for cone detection and pose estimation using stereo cameras is illustrated in \Cref{fig:vision-pipeline-arch}.  Unlike a conventional stereo pipeline where object detection is done on both the images of the stereo pair, here it is done only on the left frame of the stereo pair using the YOLOv2 network as mentioned in \Cref{subsubsection:object-detection}. The bounding boxes in the left camera frame are then propagated to the right camera frame using stereo geometry. Finally SIFT features are matched between corresponding bounding boxes and triangulated for position estimates.

\subsubsection*{Bounding Box Propagation}\label{subsubsection:bounding-box-prop}

\begin{figure}[ht]
    \centering
    \includegraphics[width=100mm]{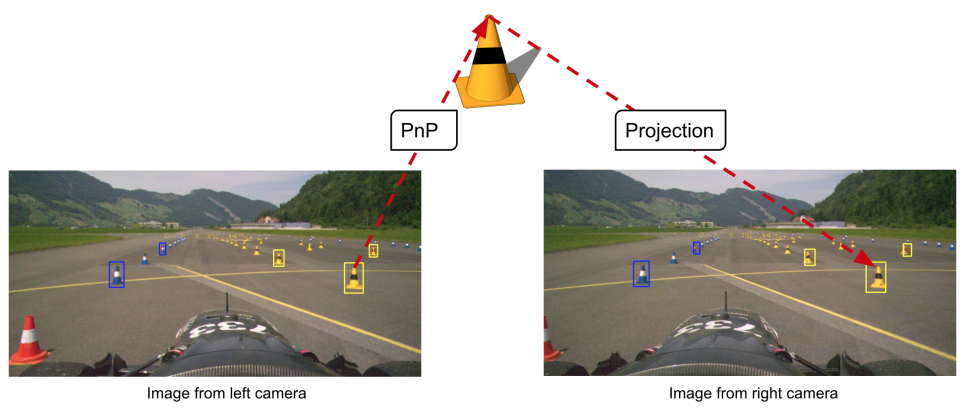}
    \caption{ Propagation of a bounding box from the left frame to the right frame.}
    \label{fig:stereo-prop-box}
\end{figure}

With the knowledge of the location of the cone in the left camera frame, an approximation of the position of the same cone in the right camera frame has to be estimated. Therefore, the extracted keypoints are projected into the 3D space using the PnP algorithm as described in \Cref{subsubsec:mono-3d}. Consequently, this 3D position is projected into the image plane of the right camera using the epipolar constraints and an approximate bounding box is drawn on the right camera frame. Therefore, using this approach, a bounding box detected in the left image frame is propagated spatially to the right camera frame which is depicted in \Cref{fig:stereo-prop-box}.

\subsubsection*{Feature Matching and Triangulation}\label{subsubsection:feat-matching}

\begin{figure}[ht]
    \centering
    \includegraphics[scale=0.4]{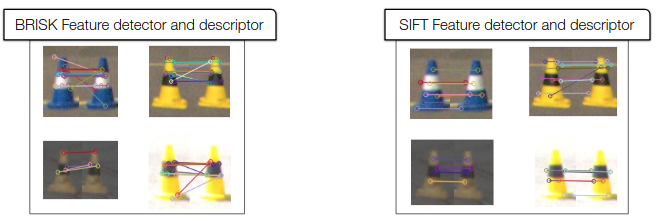}
    \caption{Comparison of feature matching obtained from BRISK and SIFT feature descriptors.}
    \label{fig:stereo-feature-descriptors}
\end{figure}

Although the bounding box has been approximated in the right camera frame, it is challenging to extract the position of the exact seven keypoints just from propagation of the bounding box. This is attributed to the offset caused by the 3D-2D projection errors. Therefore, there is a need of feature matching between the bounding box pair obtained from the left and the right image frames. BRISK and SIFT feature extraction techniques were explored and the SIFT feature extractor was selected as it was observed to perform better at admissible computational load. Fewer mismatched features were found in SIFT feature matching as compared to BRISK feature matching as  illustrated in \Cref{fig:stereo-feature-descriptors}. As a result, the median of positions estimated by triangulating SIFT features gives more accurate estimates of cone positions than BRISK features.

Given the feature descriptors of the bounding box of the left and right image frames, a ``KNN'' feature matching technique is applied to obtain the closest matched descriptors. Consequently, triangulation is performed on each of these matched descriptor pairs to obtain the cone's accurate 3D position estimate. For robustness against outliers, median filtering is applied on the obtained 3D position estimates and a single 3D position estimate is calculated.

\subsection{Perception Validation \& Results}
\subsubsection{LiDAR-based Results}
The color estimation module is validated by comparing the performance of the CNN and a rule-based approaches. In addition to that, we show a map created with our SLAM technique (see \Cref{sec:slam}) with only LiDAR cone and color observations in order to validate the color estimation over the period of one lap.

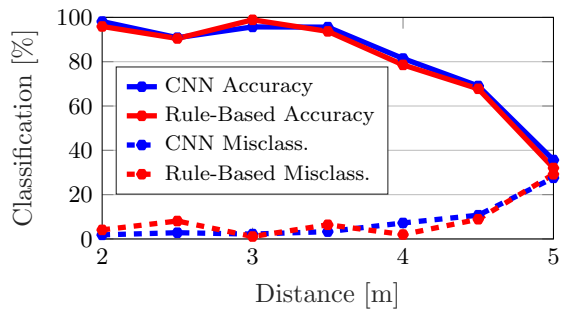
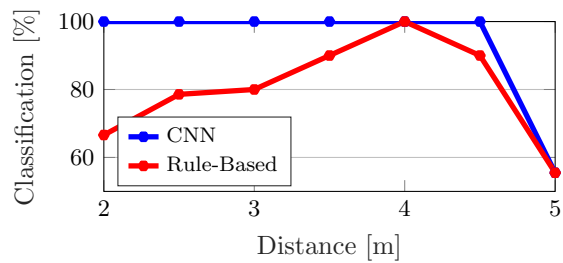
\begin{figure}[ht]
    \centering
	\begin{subfigure}{0.45\textwidth}
		\setlength{\figureheight}{4cm}
	    \setlength{\figurewidth}{6cm}
%
\begin{tikzpicture}

\begin{axis}[%
width=\figurewidth,
height=0.737\figureheight,
at={(0\figurewidth,0\figureheight)},
scale only axis,
xmin=2.000,
xmax=5.000,
xlabel style={font=\color{white!15!black}},
xlabel={Distance [m]},
ymin=0.000,
ymax=100.000,
ylabel style={font=\color{white!15!black}},
ylabel={Classification [\%]},
axis background/.style={fill=white},
title style={font=\bfseries},
ymajorgrids,
legend style={at={(0.03,0.2)}, anchor=south west, legend cell align=left, align=left, draw=white!15!black}
]
\addplot [color=blue, line width=2.0pt, mark=asterisk, mark options={solid, blue}]
  table[row sep=crcr]{%
2.000	98.140\\
2.500	90.84\\
3.000	95.700\\
3.500	95.600\\
4.000	81.440\\
4.500	69.040\\
5.000	35.630\\
};
\addlegendentry{CNN Accuracy}

\addplot [color=red, line width=2.0pt, mark=asterisk, mark options={solid, red}]
  table[row sep=crcr]{%
2.000	95.900\\
2.500	90.470\\
3.000	98.900\\
3.500	93.680\\
4.000	78.560\\
4.500	67.770\\
5.000	32.030\\
};
\addlegendentry{Rule-Based Accuracy}

\addplot [color=blue, dashed, line width=2.0pt, mark=asterisk, mark options={solid, blue}]
  table[row sep=crcr]{%
2.000	1.850\\
2.500	2.820\\
3.000	2.200\\
3.500	3.300\\
4.000	7.220\\
4.500	10.720\\
5.000	27.590\\
};
\addlegendentry{CNN Misclass.}

\addplot [color=red, dashed, line width=2.0pt, mark=asterisk, mark options={solid, red}]
  table[row sep=crcr]{%
2.000	4.100\\
2.500	8.100\\
3.000	1.100\\
3.500	6.40\\
4.000	2.000\\
4.500	8.900\\
5.000	29.200\\
};
\addlegendentry{Rule-Based Misclass.}

\end{axis}
\end{tikzpicture}%
        \caption{Case 1: The network is evaluated using test data which are similar to the images that the model was trained on.}
        \label{fig:lidar-color-validation-same}
    \end{subfigure}
	\hspace{1cm}
	\begin{subfigure}{0.45\textwidth}
		\setlength{\figureheight}{4.0cm}
	    \setlength{\figurewidth}{6cm}
%
\begin{tikzpicture}

\begin{axis}[%
width=\figurewidth,
height=0.565\figureheight,
at={(0\figurewidth,0\figureheight)},
scale only axis,
xmin=2.000,
xmax=5.000,
xlabel style={font=\color{white!15!black}},
xlabel={Distance [m]},
ymin=50.000,
ymax=100.000,
ylabel style={font=\color{white!15!black}},
ylabel={Classification [\%]},
axis background/.style={fill=white},
title style={font=\bfseries},
ymajorgrids,
legend style={at={(0.03,0.03)}, anchor=south west, legend cell align=left, align=left, draw=white!15!black}
]
\addplot [color=blue, line width=2.0pt, mark=asterisk, mark options={solid, blue}]
  table[row sep=crcr]{%
2.000	100.000\\
2.500	100.000\\
3.000	100.000\\
3.500	100.000\\
4.000	100.000\\
4.500	100.000\\
5.000	55.500\\
};
\addlegendentry{CNN}

\addplot [color=red, line width=2.0pt, mark=asterisk, mark options={solid, red}]
  table[row sep=crcr]{%
2.000	66.600\\
2.500	78.570\\
3.000	80.00\\
3.500	90.000\\
4.000	100.000\\
4.500	90.000\\
5.000	55.500\\
};
\addlegendentry{Rule-Based}
\end{axis}
\end{tikzpicture}%
        \caption{Case 2: The network is evaluated using test data which are different from the images the model was trained on.}
        \label{fig:lidar-color-validation-different}
    \end{subfigure}
  \caption{Comparison of classification performance of the CNN and Rule-Based approaches.}
\label{fig:lidar_experiment}
\end{figure}

\Cref{fig:lidar_experiment} compares the performance of the CNN and a rule-based approaches on cones of the same type (data which are similar to one used for training) as well as different type as those used to train the CNN respectively. The rule-based approach works by computing the direction of change in the intensity gradient along the vertical axis. A positive change followed by a negative one implies that the cone is blue in color whereas a negative change followed by a positive one represents a yellow cone. For test data with cones of the same type, the CNN and the rule-based approaches yield comparable results with both achieving a prediction accuracy of 96\% for cones that are close-by, as illustrated in \Cref{fig:lidar-color-validation-same}. However, the superior performance of the CNN is evident when both approaches are evaluated using test data that consists of cones of a different type. These cones have different point cloud intensities than usual due to the presence of an additional \textit{FSG} sticker on the cone. The rule based approach reports a large number of misclassifications whereas the CNN is hardly affected by it.  Furthermore, due to the use of an asymmetric cross-entropy loss function, despite the fact that classification accuracy is low at large distances, the miss-classification rate is small.

\Cref{fig:lidar-slam-map} shows a map created by SLAM (see \Cref{sec:slam}) using the cone position and color estimates from LiDAR alone. All the cones in the map are accurately colored, thus proving the accuracy and reliability of the color estimation module. The  cones near the start line are deliberately not colored in order to prevent any potential misclassifications, caused by orange cones, during the start.

\subsubsection{Camera-based Results}

\begin{figure*}[ht]
    \centering
	\begin{subfigure}{0.45\textwidth}
		\setlength{\figureheight}{3cm}
	    \setlength{\figurewidth}{6cm}
        \input{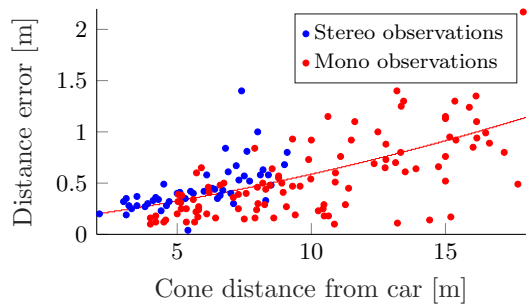}
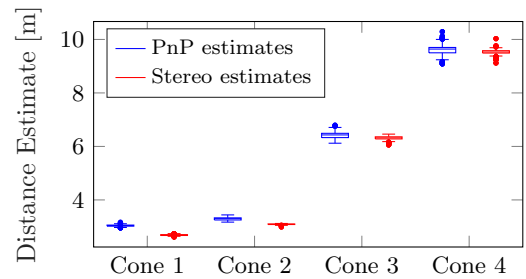
        \caption{The distance errors between LiDAR and vision estimated cone position.}
        \label{fig:experiment-camera-lidar}
    \end{subfigure}
	\hspace{1cm}
	\begin{subfigure}{0.45\textwidth}
		\setlength{\figureheight}{3cm}
	    \setlength{\figurewidth}{6cm}
%
\begin{tikzpicture}

\begin{axis}[%
width=0.951\figurewidth,
height=\figureheight,
at={(0\figurewidth,0\figureheight)},
scale only axis,
unbounded coords=jump,
xmin=0.50,
xmax=8.50,
xtick={1.50,3.50,5.50,7.50,9.50},
xticklabels={{Cone 1},{Cone 2},{Cone 3},{Cone 4},{Cone 5}},
ymin=2.24,
ymax=10.67,
ylabel style={font=\color{white!15!black}},
ylabel={Distance Estimate [m]},
axis background/.style={fill=white},
title style={font=\bfseries},
legend style={at={(0.03,0.97)}, anchor=north west, legend cell align=left, align=left, draw=white!15!black}
]
\addplot [color=blue, dashed, forget plot]
  table[row sep=crcr]{%
1.00	3.06\\
1.00	3.11\\
};
\addplot [color=red, dashed, forget plot]
  table[row sep=crcr]{%
2.00	2.70\\
2.00	2.74\\
};
\addplot [color=blue, dashed, forget plot]
  table[row sep=crcr]{%
3.00	3.33\\
3.00	3.44\\
};
\addplot [color=red, dashed, forget plot]
  table[row sep=crcr]{%
4.00	3.10\\
4.00	3.12\\
};
\addplot [color=blue, dashed, forget plot]
  table[row sep=crcr]{%
5.00	6.49\\
5.00	6.71\\
};
\addplot [color=red, dashed, forget plot]
  table[row sep=crcr]{%
6.00	6.36\\
6.00	6.46\\
};
\addplot [color=blue, dashed, forget plot]
  table[row sep=crcr]{%
7.00	9.70\\
7.00	10.00\\
};
\addplot [color=red, dashed, forget plot]
  table[row sep=crcr]{%
8.00	9.58\\
8.00	9.69\\
};
\addplot [color=blue, dashed, forget plot]
  table[row sep=crcr]{%
1.00	2.97\\
1.00	3.02\\
};
\addplot [color=red, dashed, forget plot]
  table[row sep=crcr]{%
2.00	2.64\\
2.00	2.67\\
};
\addplot [color=blue, dashed, forget plot]
  table[row sep=crcr]{%
3.00	3.17\\
3.00	3.25\\
};
\addplot [color=red, dashed, forget plot]
  table[row sep=crcr]{%
4.00	3.05\\
4.00	3.08\\
};
\addplot [color=blue, dashed, forget plot]
  table[row sep=crcr]{%
5.00	6.12\\
5.00	6.33\\
};
\addplot [color=red, dashed, forget plot]
  table[row sep=crcr]{%
6.00	6.18\\
6.00	6.28\\
};
\addplot [color=blue, dashed, forget plot]
  table[row sep=crcr]{%
7.00	9.24\\
7.00	9.50\\
};
\addplot [color=red, dashed, forget plot]
  table[row sep=crcr]{%
8.00	9.38\\
8.00	9.49\\
};
\addplot [color=blue, forget plot]
  table[row sep=crcr]{%
0.88	3.11\\
1.12	3.11\\
};
\addplot [color=red, forget plot]
  table[row sep=crcr]{%
1.88	2.74\\
2.12	2.74\\
};
\addplot [color=blue, forget plot]
  table[row sep=crcr]{%
2.88	3.44\\
3.12	3.44\\
};
\addplot [color=red, forget plot]
  table[row sep=crcr]{%
3.88	3.12\\
4.12	3.12\\
};
\addplot [color=blue, forget plot]
  table[row sep=crcr]{%
4.88	6.71\\
5.12	6.71\\
};
\addplot [color=red, forget plot]
  table[row sep=crcr]{%
5.88	6.46\\
6.12	6.46\\
};
\addplot [color=blue, forget plot]
  table[row sep=crcr]{%
6.88	10.00\\
7.12	10.00\\
};
\addplot [color=red, forget plot]
  table[row sep=crcr]{%
7.88	9.69\\
8.12	9.69\\
};
\addplot [color=blue, forget plot]
  table[row sep=crcr]{%
0.88	2.97\\
1.12	2.97\\
};
\addplot [color=red, forget plot]
  table[row sep=crcr]{%
1.88	2.64\\
2.12	2.64\\
};
\addplot [color=blue, forget plot]
  table[row sep=crcr]{%
2.88	3.17\\
3.12	3.17\\
};
\addplot [color=red, forget plot]
  table[row sep=crcr]{%
3.88	3.05\\
4.12	3.05\\
};
\addplot [color=blue, forget plot]
  table[row sep=crcr]{%
4.88	6.12\\
5.12	6.12\\
};
\addplot [color=red, forget plot]
  table[row sep=crcr]{%
5.88	6.18\\
6.12	6.18\\
};
\addplot [color=blue, forget plot]
  table[row sep=crcr]{%
6.88	9.24\\
7.12	9.24\\
};
\addplot [color=red, forget plot]
  table[row sep=crcr]{%
7.88	9.38\\
8.12	9.38\\
};
\addplot [color=blue, forget plot]
  table[row sep=crcr]{%
0.75	3.02\\
0.75	3.06\\
1.25	3.06\\
1.25	3.02\\
0.75	3.02\\
};
\addplot [color=red, forget plot]
  table[row sep=crcr]{%
1.75	2.67\\
1.75	2.70\\
2.25	2.70\\
2.25	2.67\\
1.75	2.67\\
};
\addplot [color=blue, forget plot]
  table[row sep=crcr]{%
2.75	3.25\\
2.75	3.33\\
3.25	3.33\\
3.25	3.25\\
2.75	3.25\\
};
\addplot [color=red, forget plot]
  table[row sep=crcr]{%
3.75	3.08\\
3.75	3.10\\
4.25	3.10\\
4.25	3.08\\
3.75	3.08\\
};
\addplot [color=blue, forget plot]
  table[row sep=crcr]{%
4.75	6.33\\
4.75	6.49\\
5.25	6.49\\
5.25	6.33\\
4.75	6.33\\
};
\addplot [color=red, forget plot]
  table[row sep=crcr]{%
5.75	6.28\\
5.75	6.36\\
6.25	6.36\\
6.25	6.28\\
5.75	6.28\\
};
\addplot [color=blue]
  table[row sep=crcr]{%
6.75	9.50\\
6.75	9.70\\
7.25	9.70\\
7.25	9.50\\
6.75	9.50\\
};
\addlegendentry{PnP estimates}

\addplot [color=red]
  table[row sep=crcr]{%
7.75	9.49\\
7.75	9.58\\
8.25	9.58\\
8.25	9.49\\
7.75	9.49\\
};
\addlegendentry{Stereo estimates}

\addplot [color=blue, forget plot]
  table[row sep=crcr]{%
0.75	3.04\\
1.25	3.04\\
};
\addplot [color=red, forget plot]
  table[row sep=crcr]{%
1.75	2.69\\
2.25	2.69\\
};
\addplot [color=blue, forget plot]
  table[row sep=crcr]{%
2.75	3.30\\
3.25	3.30\\
};
\addplot [color=red, forget plot]
  table[row sep=crcr]{%
3.75	3.09\\
4.25	3.09\\
};
\addplot [color=blue, forget plot]
  table[row sep=crcr]{%
4.75	6.43\\
5.25	6.43\\
};
\addplot [color=red, forget plot]
  table[row sep=crcr]{%
5.75	6.32\\
6.25	6.32\\
};
\addplot [color=blue, forget plot]
  table[row sep=crcr]{%
6.75	9.64\\
7.25	9.64\\
};
\addplot [color=red, forget plot]
  table[row sep=crcr]{%
7.75	9.53\\
8.25	9.53\\
};
\addplot [color=black, draw=none, mark size=0.8pt, mark=*, mark options={solid, blue}, forget plot]
  table[row sep=crcr]{%
1.00	2.96\\
1.00	2.96\\
1.00	2.96\\
1.00	2.96\\
1.00	2.96\\
1.00	2.96\\
1.00	2.96\\
1.00	2.96\\
1.00	2.96\\
1.00	2.97\\
1.00	2.97\\
1.00	2.97\\
1.00	3.11\\
1.00	3.11\\
1.00	3.11\\
1.00	3.11\\
1.00	3.11\\
1.00	3.11\\
1.00	3.16\\
1.00	3.16\\
1.00	3.16\\
};
\addplot [color=black, draw=none, mark size=0.8pt, mark=*, mark options={solid, red}, forget plot]
  table[row sep=crcr]{%
2.00	2.62\\
2.00	2.62\\
2.00	2.62\\
2.00	2.62\\
2.00	2.62\\
2.00	2.62\\
2.00	2.63\\
2.00	2.63\\
2.00	2.74\\
2.00	2.74\\
2.00	2.74\\
2.00	2.74\\
2.00	2.74\\
2.00	2.74\\
2.00	2.74\\
};
\addplot [color=black, draw=none, mark size=0.8pt, mark=*, mark options={solid, blue}, forget plot]
  table[row sep=crcr]{%
nan	nan\\
};
\addplot [color=black, draw=none, mark size=0.8pt, mark=*, mark options={solid, red}, forget plot]
  table[row sep=crcr]{%
4.00	2.99\\
4.00	2.99\\
4.00	2.99\\
4.00	3.04\\
4.00	3.04\\
4.00	3.04\\
4.00	3.04\\
4.00	3.04\\
4.00	3.04\\
4.00	3.05\\
4.00	3.05\\
4.00	3.05\\
};
\addplot [color=black, draw=none, mark size=0.8pt, mark=*, mark options={solid, blue}, forget plot]
  table[row sep=crcr]{%
5.00	6.75\\
5.00	6.75\\
5.00	6.79\\
5.00	6.79\\
5.00	6.79\\
};
\addplot [color=black, draw=none, mark size=0.8pt, mark=*, mark options={solid, red}, forget plot]
  table[row sep=crcr]{%
6.00	6.06\\
6.00	6.06\\
6.00	6.06\\
6.00	6.06\\
6.00	6.06\\
6.00	6.09\\
6.00	6.09\\
6.00	6.09\\
6.00	6.13\\
6.00	6.13\\
6.00	6.13\\
6.00	6.14\\
6.00	6.14\\
6.00	6.14\\
6.00	6.15\\
6.00	6.15\\
6.00	6.15\\
6.00	6.16\\
6.00	6.16\\
6.00	6.16\\
};
\addplot [color=black, draw=none, mark size=0.8pt, mark=*, mark options={solid, blue}, forget plot]
  table[row sep=crcr]{%
7.00	9.09\\
7.00	9.09\\
7.00	9.09\\
7.00	9.09\\
7.00	9.09\\
7.00	9.09\\
7.00	9.12\\
7.00	9.12\\
7.00	9.12\\
7.00	9.17\\
7.00	9.17\\
7.00	9.17\\
7.00	9.17\\
7.00	9.17\\
7.00	9.17\\
7.00	10.01\\
7.00	10.01\\
7.00	10.01\\
7.00	10.03\\
7.00	10.03\\
7.00	10.03\\
7.00	10.03\\
7.00	10.05\\
7.00	10.05\\
7.00	10.05\\
7.00	10.06\\
7.00	10.06\\
7.00	10.06\\
7.00	10.12\\
7.00	10.12\\
7.00	10.12\\
7.00	10.29\\
7.00	10.29\\
7.00	10.29\\
};
\addplot [color=black, draw=none, mark size=0.8pt, mark=*, mark options={solid, red}, forget plot]
  table[row sep=crcr]{%
8.00	9.12\\
8.00	9.12\\
8.00	9.12\\
8.00	9.24\\
8.00	9.24\\
8.00	9.24\\
8.00	9.26\\
8.00	9.26\\
8.00	9.26\\
8.00	9.28\\
8.00	9.28\\
8.00	9.28\\
8.00	9.31\\
8.00	9.31\\
8.00	9.31\\
8.00	9.32\\
8.00	9.32\\
8.00	9.33\\
8.00	9.33\\
8.00	9.33\\
8.00	9.33\\
8.00	9.33\\
8.00	9.33\\
8.00	9.36\\
8.00	9.36\\
8.00	9.36\\
8.00	9.36\\
8.00	9.36\\
8.00	9.36\\
8.00	9.36\\
8.00	9.36\\
8.00	9.36\\
8.00	9.36\\
8.00	9.36\\
8.00	9.72\\
8.00	9.72\\
8.00	9.73\\
8.00	9.73\\
8.00	9.73\\
8.00	9.76\\
8.00	9.76\\
8.00	9.76\\
8.00	10.03\\
8.00	10.03\\
8.00	10.03\\
};
\end{axis}
\end{tikzpicture}%
        \caption{The variance of 3D depth estimation obtained through PnP and through triangulation.}
        \label{fig:experiment-stereo-mono}
    \end{subfigure}
  \caption{Comparison of different 3D estimation techniques and comparison of lidar and vision cone observation pipeline.}
\end{figure*}

We compare the 3D positions estimates from the vision pipeline with those from an accurate range sensor such as the LiDAR, as can be seen in \Cref{fig:experiment-camera-lidar}. As expected, the error increases with distance to the cone. However, it is less than $\SI{1.0}{\meter}$ at a distance of \SI{15}{\meter}.

An experiment was conducted to compare the 3D positions of cones estimated via PnP and those estimated via triangulation. For this experiment setup, cones were placed in front of the car at various distances and then images were actively acquired. The estimated variance of each cone was observed over time and plotted as a box plot in \Cref{fig:experiment-stereo-mono}. This figure shows that the variance in the position estimates is higher for positions estimated via PnP (mono pipeline) and this is mitigated through triangulation (stereo pipeline). 

\begin{figure*}[ht]
    \centering
	\begin{subfigure}{0.45\textwidth}
        \centering
    	\setlength{\figureheight}{5cm}
        \setlength{\figurewidth}{5cm}
%
\definecolor{mycolor1}{rgb}{1.00000,0.75300,0.00000}%
\definecolor{mycolor2}{rgb}{0.00000,1.00000,1.00000}%
\begin{tikzpicture}

\begin{axis}[%
width=0.951\figurewidth,
height=\figureheight,
at={(0\figurewidth,0\figureheight)},
scale only axis,
xmin=-20.00000,
xmax=20.00000,
ymin=-30.00000,
ymax=20.00000,
xmajorgrids,
ymajorgrids,
ylabel style={font=\color{white!15!black}},
ylabel={Y [m]},
xlabel style={font=\color{white!15!black}},
xlabel={X [m]},
legend style={at={(0.71,0.02)}, anchor=south east, legend cell align=left, align=left, draw=white!15!black}
]
\addplot[only marks, mark=x, mark options={solid, mycolor1}, mark size=3.0pt, color=mycolor1] table[row sep=crcr]{%
x	y\\
7.60455	-2.95506\\
9.15855	-0.70806\\
11.51055	0.92994\\
13.29555	2.86194\\
14.45055	4.52094\\
15.20655	6.87294\\
15.01755	9.37194\\
14.32455	11.72394\\
13.06455	14.07594\\
11.44755	15.75594\\
8.86455	17.47794\\
6.47055	18.31794\\
3.88755	18.04494\\
1.97655	17.12094\\
-0.96345	15.58794\\
-3.79845	13.84494\\
-6.46545	11.66094\\
-8.71245	10.71594\\
-11.29545	9.98094\\
-13.73145	9.03594\\
-15.89445	7.22994\\
-17.36445	5.23494\\
-18.22545	2.67294\\
-17.46945	-2.15706\\
-15.81045	-4.08906\\
-13.35345	-5.28606\\
-11.06445	-6.54606\\
-9.00645	-7.36506\\
-7.36845	-9.17106\\
-6.54945	-11.92206\\
-6.00345	-14.58906\\
-4.95345	-17.17206\\
-3.18945	-18.05406\\
-1.17345	-18.45306\\
1.01055	-18.18006\\
2.64855	-16.58406\\
3.71955	-14.63106\\
4.26555	-12.74106\\
};\addlegendentry{/$\color{blue}\scalebox{1.15}{\ensuremath \times}$ Mapped cones}

\addplot[forget plot, only marks, mark=x, mark options={solid, blue}, mark size=3.0pt, color=blue] table[row sep=crcr]{%
x	y\\
3.55155	-3.77406\\
4.97955	-1.4220\\
6.93255	0.61494\\
8.88555	2.75694\\
10.71255	4.73094\\
11.76255	5.78094\\
12.41355	7.35594\\
12.41355	8.93094\\
12.22455	10.29594\\
11.13255	12.71094\\
9.89355	13.86594\\
8.15055	15.06294\\
6.36555	15.60894\\
4.60155	15.37794\\
3.02655	14.95794\\
0.25455	13.44594\\
-2.51745	11.74494\\
-4.93245	9.70794\\
-7.80945	8.44794\\
-9.97245	7.71294\\
-12.17745	6.74694\\
-13.71045	5.46594\\
-14.84445	3.89094\\
-15.60045	2.39994\\
-15.70545	0.42594\\
-15.09645	-0.91806\\
-13.92045	-2.15706\\
-12.32445	-3.20706\\
-10.20345	-4.15206\\
-7.80945	-5.30706\\
-5.66745	-7.21806\\
-4.38645	-9.59106\\
-3.73545	-11.79606\\
-3.48345	-13.41306\\
-2.07645	-15.34506\\
0.06555	-15.21906\\
0.88455	-13.74906\\
1.61955	-11.48106\\
};

\addplot[only marks, mark=*, mark options={}, mark size=2pt, color=mycolor2] table[row sep=crcr]{%
x	y\\
2.81655	-6.35706\\
5.63055	-7.05006\\
6.53355	-4.82406\\
9.43155	-2.89206\\
9.05355	12.29094\\
0.52755	6.62094\\
-16.37745	9.83394\\
2.31255	-9.19206\\
4.81155	-10.41006\\
2.66955	-6.86106\\
5.52555	-7.61706\\
};\addlegendentry{Mapped but no color}

\addplot [color=black, line width=2.0pt, forget plot]
  table[row sep=crcr]{%
0.59055	-6.147060\\
7.83555	-8.03706\\
};

\end{axis}
\end{tikzpicture}%
        \caption{A map output from SLAM using position and color information from LiDAR alone.}
        \label{fig:lidar-slam-map}
    \end{subfigure}
	\hspace{1cm}
	\begin{subfigure}{0.45\textwidth}
        \centering
        \setlength{\figureheight}{5cm}
        \setlength{\figurewidth}{5cm}
%
\definecolor{mycolor1}{rgb}{1.00000,0.70000,0.00000}%
\begin{tikzpicture}

\begin{axis}[%
width=0.951\figurewidth,
height=\figureheight,
at={(0\figurewidth,0\figureheight)},
scale only axis,
xmin=-12.00,
xmax=60.00,
ymin=-25.00,
ymax=15.00,
ylabel style={font=\color{white!15!black}},
ylabel={Y [m]},
xlabel style={font=\color{white!15!black}},
xlabel={X [m]},
xmajorgrids,
ymajorgrids,
axis background/.style={fill=white},
title style={font=\bfseries},
legend style={at={(0.65,0.02)}, anchor=south east, legend cell align=left, align=left, draw=white!15!black}
]

\addplot [only marks, color=blue, mark size=3.0pt, draw=none, mark=x, mark options={solid, blue}, forget plot]
  table[row sep=crcr]{%
6.67	8.24\\
5.38	5.53\\
4.09	2.44\\
4.67	-1.00\\
6.70	-4.18\\
10.49	-6.01\\
15.37	-7.07\\
19.26	-7.85\\
23.48	-8.72\\
27.45	-7.36\\
31.98	-7.22\\
36.18	-8.37\\
39.24	-11.42\\
42.38	-13.57\\
45.29	-10.51\\
47.06	-6.84\\
47.30	-3.25\\
45.81	-0.12\\
42.52	2.33\\
36.77	2.50\\
33.11	1.01\\
29.77	-0.28\\
26.31	-0.62\\
23.49	1.44\\
20.38	4.30\\
17.24	6.53\\
13.65	8.31\\
10.40	9.96\\
};

\addplot [only marks, color=mycolor1, draw=none, mark=x,mark size=3.0pt,  mark options={solid, mycolor1}]
  table[row sep=crcr]{%
4.26	11.20\\
1.93	8.69\\
0.40	6.65\\
-0.41	3.80\\
-0.44	-0.03\\
0.65	-3.27\\
3.02	-6.58\\
5.49	-9.04\\
9.81	-9.82\\
14.04	-10.12\\
16.90	-10.67\\
20.42	-11.99\\
24.34	-12.26\\
28.85	-11.07\\
33.27	-11.36\\
39.76	-17.25\\
35.82	-14.35\\
44.37	-17.68\\
47.34	-14.36\\
49.57	-11.55\\
51.37	-8.46\\
51.62	-5.68\\
50.72	-1.96\\
50.31	1.38\\
48.19	4.14\\
43.21	6.10\\
37.92	5.79\\
33.87	6.18\\
29.97	5.63\\
26.84	4.49\\
23.83	6.50\\
20.23	10.15\\
15.45	12.66\\
10.37	14.11\\
6.64	13.74\\
};
\addlegendentry{/$\color{blue}\scalebox{1.15}{\ensuremath\times}$ Mapped cones}

\addplot [only marks, color=mycolor1, draw=none, mark size=3.5pt, mark=o, mark options={solid, mycolor1}]
  table[row sep=crcr]{%
0.89	6.74\\
0.00	3.80\\
0.0	0.01\\
1.07	-3.18\\
3.31	-6.43\\
5.70	-8.87\\
9.96	-9.73\\
14.15	-10.06\\
16.92	-10.59\\
20.39	-11.92\\
24.33	-12.26\\
28.75	-11.04\\
32.99	-11.35\\
35.77	-14.47\\
39.76	-17.14\\
44.35	-17.46\\
47.03	-14.16\\
49.41	-11.35\\
51.19	-8.34\\
51.44	-5.56\\
50.50	-1.90\\
50.06	1.43\\
47.95	4.11\\
42.96	5.99\\
37.83	5.65\\
33.85	5.99\\
30.02	5.46\\
26.66	4.41\\
23.61	6.36\\
20.13	10.04\\
15.51	12.69\\
10.46	14.08\\
6.88	13.61\\
4.55	11.03\\
2.31	8.66\\
};
\addlegendentry{/$\color{blue}\scalebox{1.0}{\ensuremath\bigcirc}$ Leica cones}

\addplot [forget plot, only marks, color=blue, draw=none, mark size=3.5pt, mark=o, mark options={solid, blue}]
  table[row sep=crcr]{%
5.58	5.36\\
4.40	2.36\\
5.04	-1.05\\
7.00	-4.21\\
10.70	-6.05\\
15.44	-7.08\\
19.29	-7.91\\
23.42	-8.83\\
27.43	-7.43\\
31.81	-7.35\\
35.93	-8.51\\
38.94	-11.49\\
42.13	-13.52\\
45.25	-10.29\\
46.95	-6.66\\
47.13	-3.09\\
45.63	-0.02\\
42.25	2.29\\
36.60	2.40\\
32.97	0.98\\
29.68	-0.31\\
26.32	-0.66\\
23.50	1.32\\
19.95	4.66\\
17.13	6.52\\
13.83	8.28\\
10.49	9.960\\
6.87	8.03\\
};

\addplot [color=black, line width=2.0pt]
  table[row sep=crcr]{%
1.0	-9.5\\
11.5 -4.\\
};

\end{axis}
\end{tikzpicture}%
        \caption{Comparison of stereo 3D estimate of stereo system and the ground truth.}
        \label{fig:slam_leica_vision}
    \end{subfigure}
  \caption{Maps created with SLAM using inputs from either only vision or LiDAR based pipelines. In both figures, the black line indicates the starting line.}
\end{figure*}
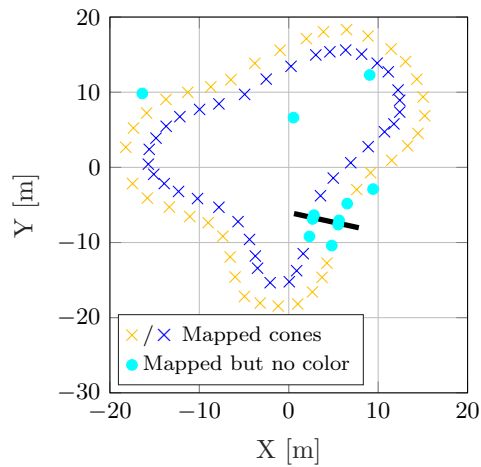
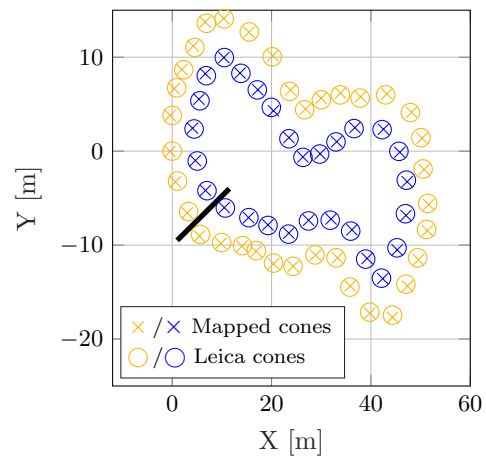

Further, we illustrate the quality of a map generated using cones from the vision pipeline. We measured a ground truth of cone positions with the Leica Totalstation, shown in \Cref{fig:slam_leica_vision}, as well as created a map with our SLAM algorithm (see \Cref{sec:slam}) using only vision-pipeline cone estimates. These two maps were then aligned using the Iterative Closest Point (ICP) \cite{chetverikov2002trimmed}. The Root Mean Square Error (RMSE) between cone estimates and its corresponding ground-truth is estimated to be \SI{0.25}{\meter}. 

\section{Motion Estimation and Mapping}\label{sec:motion estimation and mapping}

\subsection{Velocity Estimation}

Robust and accurate velocity estimation plays a critical role in an autonomous racecar. The estimated velocity is used to compensate the motion distortion in the LiDAR pipeline, propagate the state in the SLAM algorithm, as well as for the \textit{control} module. Velocity estimation needs to combine data from various sensors with a vehicle model in order to be robust against sensor failure and to compensate for model mismatch and sensor inaccuracies. This problem is addressed with a $9$ state Extended Kalman Filter (EKF), which fuses data from six different sensors.

\subsubsection{Estimator Design}\label{sec:kf}

\begin{figure}[h]
    \centering
    \includegraphics[width=0.7\linewidth]{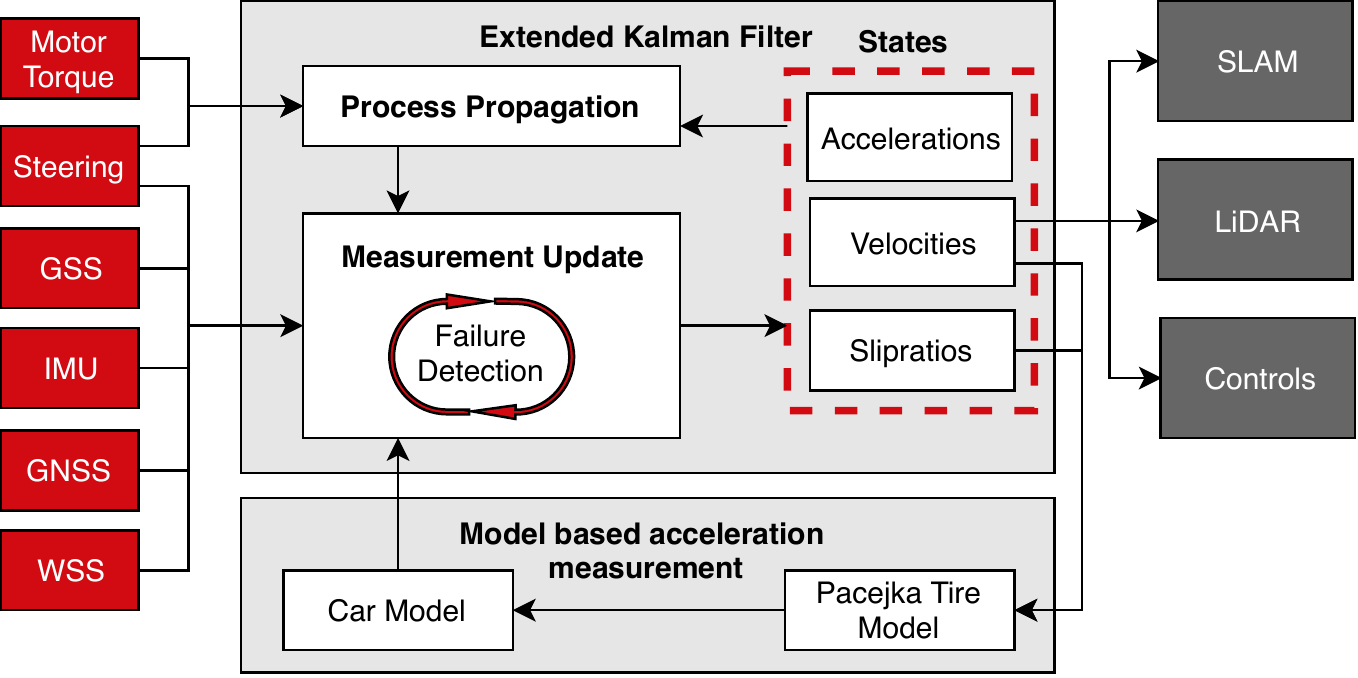}
    \setlength{\abovecaptionskip}{3pt}
    \setlength{\belowcaptionskip}{2pt}
	\caption{A simplified velocity estimation architecture.}
    \label{fig:VE_architecture}
    \vspace{0.1mm}
\end{figure}

State estimation is a common task known in robotics. The main challenge is to fuse data from different measurements, each having different bias, noise and failure characteristics, together with the prior of the robot motion model. We propose to use an Extended Kalman Filter (EKF), the well-known estimator for mildly non-linear systems with Gaussian process and sensor noise, due to its computational efficiency, and accurate estimates \cite{Thrun:2005:PR:1121596}. An overview of the velocity estimation algorithm is shown in~\Cref{fig:VE_architecture}.

\subsubsubsection{\textbf{Process Model}}\label{process_model}

The process model represents a prior distribution over the state vector. In our case, a constant acceleration process model is used since jerk is close to a zero mean Gaussian distribution. Hence, the velocities are propagated using the acceleration whereas slip ratios are propagated using the dynamics derived by time differentiation of the slip ratio (\Cref{sec:sr_equation}), resulting in the following process model,
\begin{align}\label{eq:process_model}
\begin{split}
\dot{\mathbf{v}} &= \mathbf{a} + r \left[ v_{{y}}, \, -v_{{x}} \right]^T+ \mathbf{n_{{v}}}\, ,\\
\dot{r} &= f_{{M}}(\mathbf{sr}, \mathbf{v}, r, \delta) + n_{{r}}\, , \\
\dot{\mathbf{a}} &= \mathbf{n_{{a}}}\, , \\
\dot{\mathbf{sr}} &= R \left(\text{diag}(I_{\omega}\mathbf{v}_{w x})^{-1} \mathbf{t}_{{M}}\right) + \left(\frac{C_\sigma R}{I_{\omega}} - a_{{{x}}}\right)(\text{diag}(\mathbf{v}_{w x})^{-1}\mathbf{sr}) - a_{{x}}\text{diag}(\mathbf{v}_{w x})^{-1}\mathbf{1}_{4\times 1} + \mathbf{n_{{sr}}}\, ,\\
\end{split}
\end{align}
where $f_{{M}}(\cdot)$ describes the yaw moment generated from tire forces which are in turn estimated using a linear function of longitudinal and lateral slip (see Chapter I \cite{cit:Moustapha2013}). $\mathbf{t}_M \in \mathbb{R}^{4}$ and $\delta$ are the motor torques and steering angle respectively, and are the inputs to the process model. $C_{\sigma}$ is the longitudinal tire stiffness \cite{cit:Moustapha2013} providing a linear relationship between slip ratio and longitudinal force and is estimated from experiments. $R$ is the radius of the wheel. $\mathbf{v}_{w x} \in \mathbb{R}^{4\times1}$ are the longitudinal velocity of the wheel hubs, $I_{\omega}$ is the moment of inertia of the wheel, and $\mathbf{n}_{\{\cdot\}}$ is independent and identically distributed (i.i.d.) Gaussian noise. All these build a state vector $\mathbf{x} = [\mathbf{v}^T, r, \mathbf{a}^T, \mathbf{sr}^T]^{T} \in \mathbb{R}^9$ with $\mathbf{v}=[v_{{x}}, v_{{y}}]^T$ and $r$ being the linear and angular velocities, respectively and $\mathbf{a}=[a_{{x}}, a_{{y}}]^T$ represent the linear accelerations. Finally, the four wheel slip ratios are denoted by $\mathbf{sr} = [\text{sr}_{FL}, \text{sr}_{FR}, \text{sr}_{RL}, \text{sr}_{RR}]^T$ where subscript $F/R$ means front or rear axle of the car respectively and $R/L$ denotes the right or left side of the car respectively. $\text{diag}(\mathbf{x})$ represents a diagonal matrix consisting from elements of vector $\mathbf{x}$ and $\mathbf{1}_{4\times 1} \in \mathbb{R}^{4\times1}$ is a vector of ones. This process model captures the fact that the probability of slippage increases with increased motor torques. 

Note that acceleration is part of the state vector. This is chosen in order to fuse it from multiple measurements which makes it robust and redundant with respect to the acceleration measurement alone. To reduce the uncertainty due to the change in surface inclination and time varying bias, the acceleration is fused from two sources, the tire slip model and the Inertia Measurement Unit (IMU).

\subsubsubsection{\textbf{Measurement Model}}

Since every sensor is running at their own frequency, the measurements arrive asynchronously. In addition, we assume that the sensor noise is uncorrelated among different sensors. This allows us to run the correction steps for each sensor independently.  As shown in \Cref{fig:VE_architecture}, the EKF is updated using velocity measurements $\mathbf{h_{v}(x)}$ of GSS and GNSS measurements, acceleration measurements $\mathbf{h_{a}(x)}$ from IMU as well as the tire slip model\cite{pacejka1992magic}, yaw rate measurements $h_{r}(\mathbf{x})$ from gyroscope and wheel speed measurements $\mathbf{h_{\omega}(x)}$ from the resolvers. This sums up to the following measurement model
\begin{align}\label{eq:sensor_model}
\begin{split}
\mathbf{z_{{v}}} &= \mathbf{h_{v}(x)} =   \mathbf{R}(\theta_{{{s}}}) (\mathbf{v} + r[-{p}_{{s, y}}, {p}_{{s, x}}]^T) + \mathbf{{n}_{z_{v}}} \, ,\\
z_r &= h_{r}(\mathbf{x}) = r + n_{z_{r}}\, ,\\
\mathbf{z_{a}} &= \mathbf{h_{a}(x)} = \mathbf{a} + \mathbf{n_{z_a}} \, ,\\
\mathbf{z_{\omega}} &= \mathbf{h_{\omega}(x)} = \frac{1}{R}\text{diag}(\mathbf{v}_{w x}) (\mathbf{sr}+\mathbf{1}_{4\times 1}) + \mathbf{n_{z_{\omega}}} \, ,\\
\end{split}
\end{align}
where $\mathbf{R}(\theta_{{s}})$ denotes the rotation matrix with $\theta_{{s}}$ being the orientation of the sensor in the car frame and $[{p}_{{s, x}},{p}_{{s, y}}]$ are the coordinates of the sensor also in the car frame.

To incorporate the slipratio into the wheelspeed measurement update, they are split into two components, linear hub velocity and slipratio in the measurement model explained in  ~\Cref{sec:sr_equation}. Both the components are updated simultaneously using wheelspeed measurement.

The Non-linear Observability analysis (\Cref{sec:observability_analysis}) shows that the system is still observable if either one of GSS or GNSS fails. All possible combinations of sensor configuration can be found in~\Cref{tab:ob_analysis}. If the IMU fails along with one of the velocity sensors (GSS or GNSS), then the yaw rate becomes unobservable. To avoid this, the model is converted from a full dynamic model to a partial kinematic one using a \textit{zero slip ratio measurement update} (ZSMU). High changes in wheel speeds are still captured as slip by the process model and the ZSMU later shrinks the slip ratio in the update step of the EKF, which is reliable at low car speeds ($<\SI{6}{\meter\per\second}$).

\begin{table}[h]
\centering
\begin{tabular}{@{}ccccc@{}}
\toprule
\multicolumn{4}{c}{Sensors}            & \multicolumn{1}{c}{Result}          \\ \midrule
GSS & IMU & GNSS & WSS & Observability                                        \\
$\checkmark$ & $\checkmark$ & $\checkmark$ & $\checkmark$ &$\checkmark$\\
$\times$ & $\checkmark$ & $\checkmark$ & $\checkmark$ &$\checkmark$\\
$\times$ & $\times$ & $\checkmark$ & $\checkmark$ &$\checkmark^*$\\
$\times$ & $\checkmark$ & $\times$ & $\checkmark$ &$\checkmark^*$\\
$\times$ & $\checkmark$ & $\checkmark$ & $\times$ &$\checkmark^*$\\
$\times$ & $\times$ & $\times$ & $\checkmark$ &$\checkmark^*$\\
$\times$ & $\times$ & $\times$ & $\times$ &$\times$\\ \bottomrule
\end{tabular}
\caption{Results are obtained by removing sensors successively. $\checkmark$ denotes that a sensor is available, $\times$ denotes that a sensor is not available, and $\checkmark^*$ denotes that the full state of the original system is not observable and the presented ZSMU is applied, to observe the reduced system.}
\label{tab:ob_analysis}
\end{table}

\subsubsection{Failure Detection}\label{sec:failure_detection}

Inaccurate and faulty sensor measurements result in temporarily or even permanently disturbed state estimates, given the recursive nature of the algorithm. It is therefore important to detect such sensor failures. The sensor faults can be classified as \textit{outlier} (e.g., spikes in measurements), \textit{drift} and \textit{null}. A null measurement is defined as not receiving an input from the sensor, which is addressed by updating the measurement using the respective callback.  
A Chi-square-based approach for outlier detection, and a variance based sensor isolation for drift detection are implemented. The details for the Chi-square-based approach can be found in~\cite{paper_fluela}.

The Chi-square test works flawlessly for outliers. But sensor drift is only detected after a long time when the error is large. Such failure cases are identified using a variance based sensor drift detection given by
 \begin{align}
 \sum_{i=1}^{n} ({z_{{i}}}-{\mu}_{{{z}}})^2  < k\vspace{-1mm} \, ,
 \end{align}
 where ${\mu}_{{{z}}}$ represents the mean of the sensor measurement. For each measurement, the variance is calculated using $n$ number of sensors which were used to measure that state variable. 
 If the variance is higher than the parameter $k$, the sensors are removed progressively, until the sensor with the highest contribution to the variance is rejected in the given time instant.
 The drift detection can also detect outliers but it requires measurement of a state variable from more than 2 sensors to work effectively. Hence, both outlier and drift detection are implemented alongside each other.

\subsection{Simultaneous Localization and Mapping}
\label{sec:slam}

The Formula Student competition poses multiple distinct challenges for mapping. Since the track is demarcated by cones only, a feature-based map representation is preferred over a grid-based one. Landmarks cannot be distinguished using unique descriptors, as the cones are described only by their position and color. As a result, the algorithm needs to handle uncertain data association. The SLAM algorithm must operate in real-time and the run-time should be predictable and even adjustable. Testing time is a major limiting factor for the whole system performance. Hence, it is desired to have an easily tunable algorithm that quickly achieves its full potential. Since environment conditions and sensor failures are unforeseeable, the algorithm should be able to detect malfunctioning sensors and proceed accordingly. Bearing the above criteria in mind, the fastSLAM 2.0~\cite{Montemerlo2003} algorithm was chosen. Its particle filter nature inherently provides multi-hypothesis data associations. Additionally, it is easy to trade-off its performance versus run-time by adjusting the number of particles. 
FastSLAM scales linearly in the number of landmarks and therefore provides superior computational performance compared to the quadratic complexity of an EKF-SLAM algorithm~\cite{Montemerlo2010}. 

\begin{figure}[ht]
   \centering
   \includegraphics[width=0.7\textwidth]{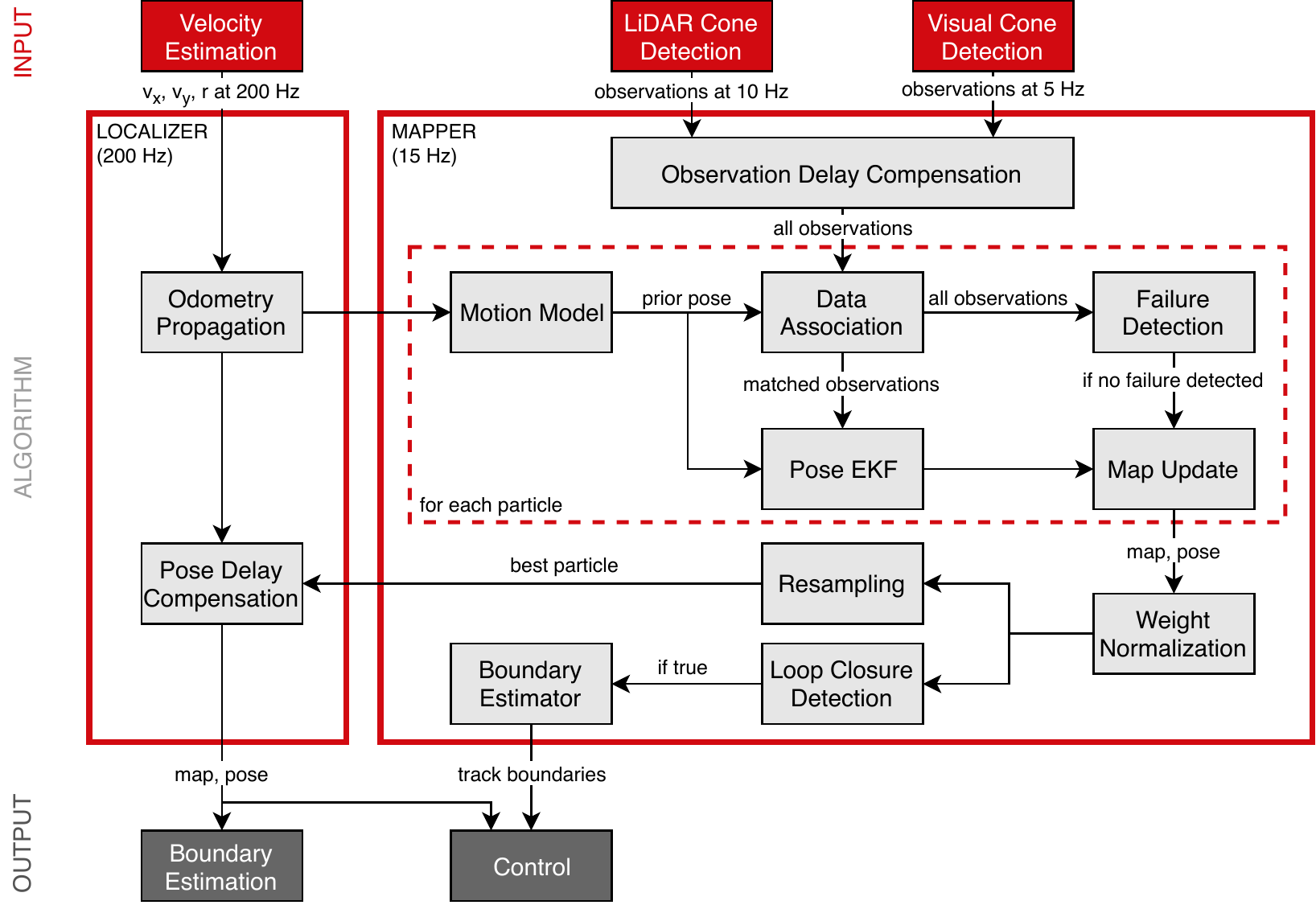}
   \caption{Detailed SLAM architecture used to fuse landmark observations from camera and LiDAR with velocity estimates into a coherent map and pose within the map. The dashed line visually demonstrates which parts of the algorithm are computed on a particle basis. }
   \label{fig:slam_arch}
\end{figure}

As illustrated in \Cref{fig:slam_arch}, the implementation is divided into two parts, a localizer which processes velocities and continuously integrates them giving a pose update at $200$\si{Hz} and a mapping algorithm. Processing the landmark observations of both perception pipelines and the cars velocity estimate result in a high frequency update of the corresponding map. The algorithm details can be found in \Cref{sec:slam_algo_details}.

\newpage
\subsubsubsection{\textbf{Failure Detection}}

To benefit most from the redundant sensor setup, a two-stage sensor failure detection is implemented. The goal is to avoid an (irreversible) map update due to observations coming from a malfunctioning sensor, penalizing measurements that do not match previous observations. 
In a first stage each landmark holds two counters, a recall counter $n_s$, which counts the number of times a landmark has been seen, and the second counter $n_m$, which is increased when a landmark is not seen, despite being in the sensors' FoV. 
If the confidence of a landmark as relation between both counters drops below a threshold, the landmark is assumed to be dubious.
The second stage of the failure detection is on a sensor level. 
A set of observations from a sensor at a given time (after delay compensation) is only accepted if enough observations match with landmarks that have a landmark confidence above 80\%, given there are any in the FoV. 

\subsubsubsection{\textbf{Lap Closure Detection and Post-processing}}

After one lap, sufficient information is collected, a static map is assumed and further observations will not increase the quality of the map significantly. The car actively detects the completion of this first lap, a process we call lap closure detection. Lap closure is detected with three basic rules. First, the spread of all particles, measured by the sample standard deviation of the position, has to be below a threshold. Second, the cars current heading has to be roughly equal to the heading at the beginning of the race and third, all particles have to be close to the starting point.

\subsubsubsection{\textbf{Localization}}

After finishing the first lap, a lap closure is detected and the operational mode switches from \textit{SLAM} to \textit{Localization} mode. The track boundaries are then identified and the map is not longer updated. The map of the most likely particle (highest weight) is taken as the map for subsequent laps. In order to localize the car in the created map, Monte Carlo Localization~\cite{dellaert99Monte} is used. A smooth pose update is given through the computation of the mean over all particles.

\subsection{Boundary Estimation}\label{sec:boundary_estimation}

The \textit{SLAM} mode has two main goals, first to map the environment using the SLAM algorithm described in \Cref{sec:slam}, and second to discover the track layout while remaining within the track boundaries. In this section, we discuss the second goal where we have to decide upon a path only given local cone observations. Note that the cones observations are probabilistic in both color and position. Additionally, false positive cones could be detected and the color may not be present at all. This makes this task challenging and means that our path planning algorithm needs to be able to deal with a high level of uncertainty.

In this section, we propose a path planning algorithm that uses the cone observations within the sensor range as guidance. Since the cones describe the track limits, it follows that the center line of the real track is in between two cones, which gives us a finite but potentially very large set of possible paths. This also means that there is no given goal position, which makes this a non-standard path planning problem. However, the cones naturally give rise to a graph search problem if an appropriate discretization is used.

\subsubsection{Path Planning Algorithm}  

Our graph search path planning algorithm consists of three steps: first the search space is discretized using a triangulation algorithm, second, a tree of possible paths is grown through the discretization. Finally, all paths are ranked by a cost function and the path with the lowest cost is picked, which gives us an estimate of the boundary and the center line. Due to the limited sensor range, we run the path planning algorithm in a receding horizon fashion, re-estimating a new center line every $\SI{50}{\milli\second}$ given new cone observations. In the following, these three algorithm steps are explained in detail.

The cones naturally form a discrete space, if the goal is to drive in between them. Thus, we discretize the $X$-$Y$ space by performing a Delaunay triangulation \cite{Delaunay1} \cite{Delaunay2}, which subdivides the $X$-$Y$ space into connected triangles. The vertices of the triangulation are the cone observations and the triangles are chosen such that the minimum internal angle of each triangle is maximized. Note that Delaunay triangulation is performed using the CGAL Library \cite{cgal}.

The second step of the algorithm consists of growing a tree of possible paths through the discretized space. Starting from the position of the car, the paths always connect to the next center points of the surrounding edges. The resulting paths describe all possible center lines given the observed cones and the current car positions. In each step, the tree is grown in a breath first manor. Furthermore, in order to avoid wasting memory and computation power, the worst paths (the ones with the highest cost) are pruned in a beam search fashion \cite{ComputingDictionaryBeam}. Thus, we only keep a fixed number of nodes. Finally, we only run a finite number of iteration before the algorithm is stopped. The number of iteration is determined experimentally to keep the computation time within the given limits and still guarantee a good coverage of all possible paths.

The final step of the algorithm computes a cost for each path generated in the second step and selects the one with the lowest cost as the most likely track center line. We choose a cost function which exploits information about the rules of the competition (see \Cref{subsec:dynamic_discipline_description}) including implications of these rules, as well as sensor knowledge (see \Cref{sec:perception}). The cost function uses five cost terms, that are explained in \Cref{tab:table_with_costs} and always apply to a specific path. 

\begin{table}[ht]
    \begin{center}
        \begin{tabular}{@{}cc@{}}
        \toprule
        \textbf{Cost} & \textbf{Reason}          \\ \midrule
        
        Maximum angle change &
        \begin{tabular}{@{}c@{}}Large angle changes from one path segment to the next\\ are unlikely because even sharp corners are\\ constructed using multiple cones\end{tabular} \\ \midrule
        
        Standard deviation of the track width & 
        \begin{tabular}{@{}c@{}}The width is regulated by the rules and is\\ unlikely to change a lot\end{tabular}\\\midrule
        
        \begin{tabular}{@{}c@{}}Standard deviation of the distance \\ between left as well as the right cones \end{tabular} &
        \begin{tabular}{@{}c@{}} Cones corresponding to the track are normally roughly space\\ equal. This is not necessarily true if both tight corners \\ and straights are present in the observed cones, however, \\this is unlikely given the limited sensor range and\\ the cost helps to remove outliers\end{tabular} \\\midrule
        
        Maximal wrong color probability &
        \begin{tabular}{@{}c@{}} If color information is given we know by the rules that\\ left cones should be blue and right cones  should be yellow, \\this cost uses the color probability given by the perception layer\\ (see Section \ref{sec:perception}) to penalize paths that do not respect\\ this rules. By using the color probability the cost becomes\\ automatically zero if there is no color \\information \end{tabular} \\\midrule
        
        \begin{tabular}{@{}c@{}}Squared difference between \\ path length and sensor range \end{tabular}  &
        \begin{tabular}{@{}c@{}}The cost penalizes too short and too long paths\\ and gives an emphasize to paths that are the same length \\ as the sensor range which is around \SI{10}{\meter} \end{tabular} \\
        
        \bottomrule
        \end{tabular}
        \caption{Overview of cost structure used to find the track boundaries.}
        \label{tab:table_with_costs}
    \end{center}
\end{table}

To get the cost of a path, the five costs are first computed, then normalized, squared, weighted and finally added together. The best path is then selected as the estimate of the center line as well as the corresponding boundaries. In a final step, the estimated center line path is then given to a pure pursuit path following controller \cite{coulter1992implementation}, which generates the input to follow the center line.

\subsection{Motion Estimation and Mapping Validation \& Results}
\subsubsection{Velocity Estimation Validation \& Results}

\begin{figure}[h]
	\centering
    \setlength{\figureheight}{4.5cm}
    \setlength{\figurewidth}{9cm}
    \input{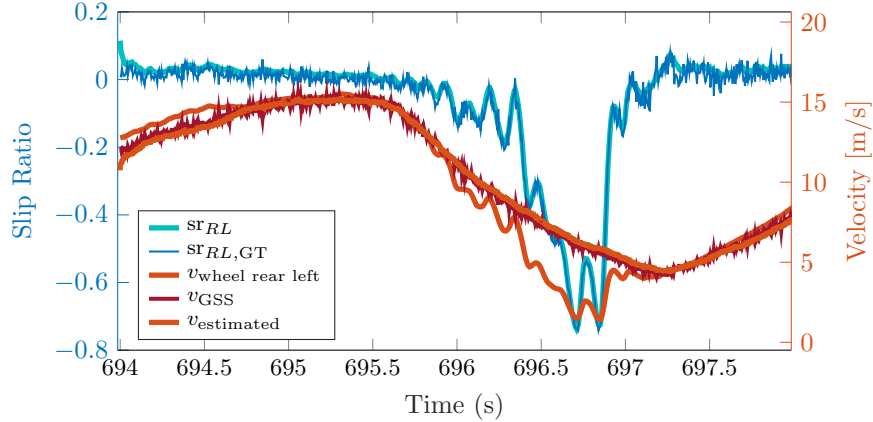}
    \caption{Estimated longitudinal velocity and slip ratio compared to ground truth data.}
    \label{fig:VE_long}
\end{figure}

\Cref{fig:VE_long} shows the estimated velocity, $v_\text{estimated}$, without the GSS compared to ground truth from GSS, $v_\text{GSS}$. Also, it shows the estimated slip ratio of rear left wheel, $\text{sr}_{RL}$, compared with slip ratio obtained from GSS and WSS, $\text{sr}_{RL,\text{GT}}$. It can be seen that the velocity estimate is accurate even when the wheels substantially slip. The very accurate velocity estimate is also shown when we compare the distance between the position of the car obtained by integrating the estimated velocity and the GPS position. The difference is less than $\SI{1.5}{\meter}$ over a $\SI{310}{\meter}$ long track, which results in a drift of less than $0.5$~\%.

\begin{figure}[h]
	\centering
    \setlength{\figureheight}{4.5cm}
    \setlength{\figurewidth}{9cm}
  	\input{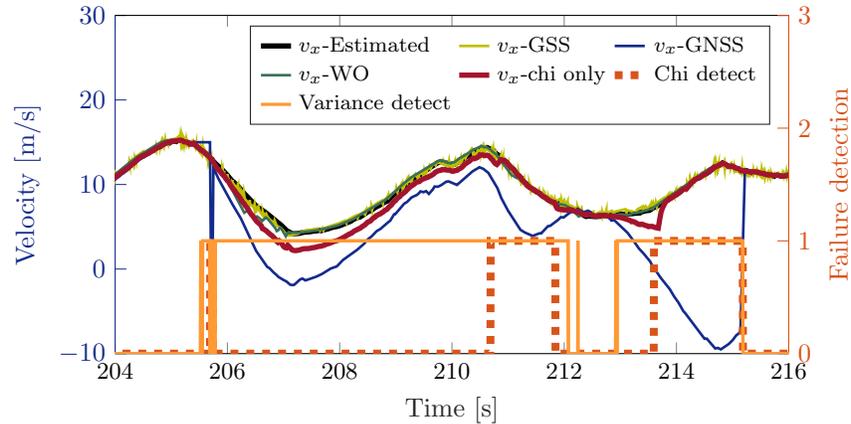}
  	\caption{Failure of the GNSS (Blue) is detected by the drift detection (Orange). $v_x$-Estimated (chi only) (Dark Brown) is the estimated velocity using the chi test (Brown dashed) which shows that the chi test without drift detection is unable to detect the full sensor failure.}
    \label{fig:VE_drift_detection}
\end{figure}

 The ability to detect and further reject sensor failures is depicted in \Cref{fig:VE_drift_detection}.  It can be observed that the chi-square based failure detection is able to reject the signal only when the failure is short-lived, whereas the drift failure detection is able to also discard continuous sensor failures. Using both techniques in conjunction ensures removal of most of the sensor failures.

\subsubsection{SLAM Validation \& Results}
 Multiple results were already presented in \Cref{subsec:lidar_estimation} as well as \Cref{subsec:camera_estimation} which show the estimated map with comparison to ground-truth measurements. By purely vision-based mapping, a RMSE of $\SI{0.25}{\meter}$ between estimated map and GT is achieved, whereas maps made with only LiDAR measurements gives a RMSE of $\SI{0.23}{\meter}$.

\begin{figure}[ht]
    \centering
    \setlength{\figureheight}{5cm}
    \setlength{\figurewidth}{9cm}
  	    \input{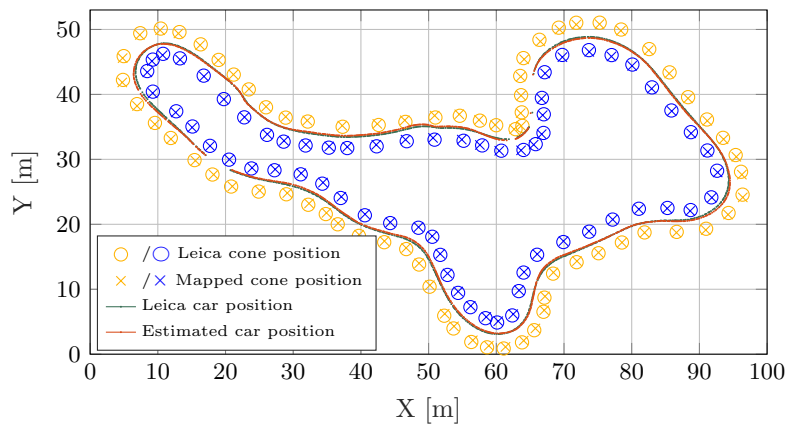}
    \label{fig:result-SLAM-map}
    \caption{Mapping and localization comparison against ground-truth measurements.}
\end{figure}

The $\SI{230}{\meter}$ long track shown in \Cref{fig:result-SLAM-map} is mapped with both LiDAR and vision pipeline. In addition, the car’s position is measured in order to verify the vehicle's particulate filter-based location. Therefore, a Totalstation’s measurement prism was attached to the car and measured for positional ground-truth. The estimated position differs only by a RMSE of $\SI{0.2}{\meter}$ from the tracked ground-truth.

\subsubsection{Boundary Estimation Validation \& Results}

\begin{figure}[ht]
    \centering
	\begin{subfigure}{0.45\textwidth}
    	\centering
    	\includegraphics[width=1.1\textwidth]{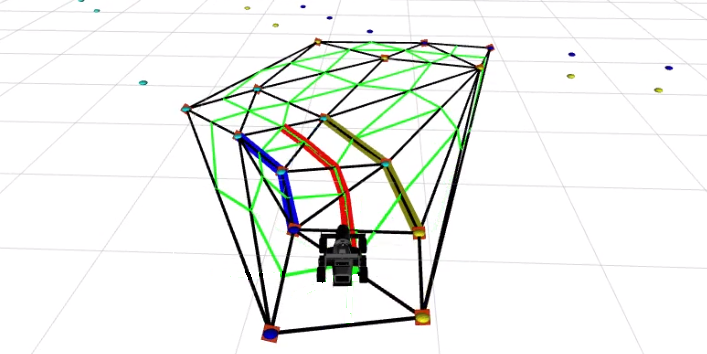}
    	\caption{Depicted in black is the Delaunay triangulation, in green the tree of possible paths, in red the chosen path, and in blue and yellow the resulting boundaries for the chosen path. }
    	\label{fig:Tree}
    \end{subfigure}
	\hspace{1cm}
	\begin{subfigure}{0.45\textwidth}
        \centering
    	\setlength{\figureheight}{3.5cm}
    	\setlength{\figurewidth}{5cm}
%
\definecolor{mycolor1}{rgb}{0.00000,0.44700,0.74100}%
\begin{tikzpicture}

\begin{axis}[%
width=0.951\figurewidth,
height=\figureheight,
at={(0\figurewidth,0\figureheight)},
scale only axis,
xmin=0.00000,
xmax=10.00000,
xlabel style={font=\color{white!15!black}},
xlabel={Distance from the car [m]},
ymin=0.00000,
ymax=0.05000,
ylabel style={font=\color{white!15!black}},
ylabel={Paths that left the track [$\%$]},
yticklabel style={
        /pgf/number format/fixed,
        /pgf/number format/precision=4
},
axis background/.style={fill=white},
title style={font=\bfseries},
legend style={legend cell align=left, align=left, draw=white!15!black}
]
\addplot[ybar interval, fill=mycolor1, fill opacity=0.60000, draw=black, area legend] table[row sep=crcr] {%
x	y\\
7.00000	0.00400\\
7.26000	0.00800\\
7.52000	0.00800\\
7.78000	0.01300\\
8.04000	0.01300\\
8.30000	0.01900\\
8.56000	0.01900\\
8.82000	0.03100\\
9.08000	0.03100\\
9.34000	0.03800\\
9.60000	0.03800\\
9.86000	0.04200\\
10.12000	0.04200\\
10.38000	0.04200\\
10.64000	0.04200\\
10.90000	0.04200\\
11.16000	0.04200\\
11.42000	0.04200\\
11.68000	0.04200\\
11.94000	0.04200\\
12.20000	0.04200\\
12.46000	0.04200\\
12.72000	0.04200\\
12.98000	0.04200\\
13.24000	0.04200\\
13.50000	0.04200\\
13.76000	0.04200\\
14.02000	0.04200\\
14.28000	0.04200\\
14.54000	0.04200\\
14.80000	0.04200\\
15.06000	0.04200\\
15.32000	0.04200\\
15.58000	0.04200\\
15.84000	0.04200\\
16.10000	0.04200\\
16.36000	0.04200\\
16.62000	0.04200\\
16.88000	0.04200\\
17.14000	0.04200\\
17.40000	0.04200\\
17.66000	0.04200\\
17.92000	0.04200\\
18.18000	0.04200\\
18.44000	0.04200\\
18.7000	0.04200\\
18.96000	0.04200\\
19.22000	0.04200\\
19.48000	0.04200\\
19.74000	1.00000\\
20.00000	1.00000\\
};

\end{axis}
\end{tikzpicture}%
        \caption{Percentage of paths outside of the real track at a certain distance from the car.}
        \label{fig:boundary_estimation_errors}
    \end{subfigure}
  \caption{Evaluation of boundary estimation.}
\label{fig:be_experiment}
\end{figure}
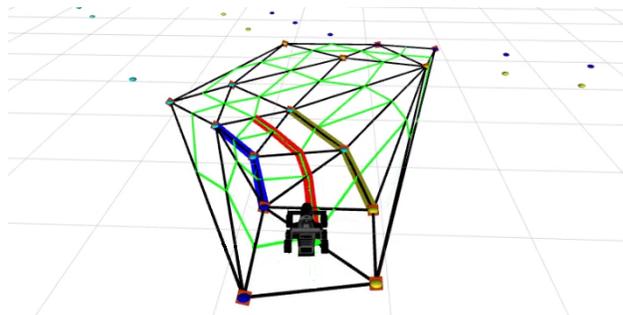
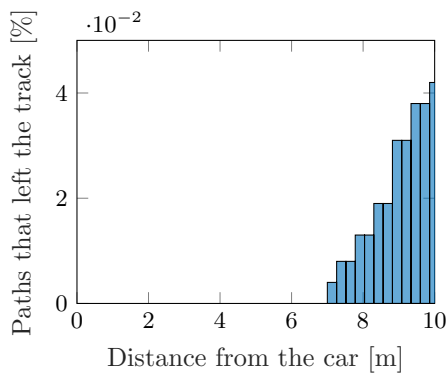

The result of all three steps of our proposed path planning algorithm are visualized in \Cref{fig:Tree}, where the car is able to find the proper path and track boundaries. The Delaunay triangulation given the observed cones as vertices is shown in black, and the tree of possible trajectories through the mid-points is shown in green. It is clearly visible that an exhaustive library of possible paths is generated using this approach, and that the cost function picked a path (shown in red) that is regularly spaced and of the correct length given the sensor range of the car. 

The boundary estimation algorithm is tested extensively under various track layouts as well as different road conditions. The system is robust when driving through rule compliant tracks. However, in case of distances larger than \SI{5}{\meter} between two consecutive cones, we observed an increased error in predicting the track. In \Cref{fig:boundary_estimation_errors} we show the classification success rate from the FSG trackdrive competition. For this track, only $4.2\%$ of all iterations yield a path which is not inside the real track. Furthermore, the miss predicted paths only leave the real track at more than \SI{7}{\meter} away from the car. At these distances, a miss-classified path does not cause the car to leave the track, since we run the algorithm in a receding horizon fashion, and only driving with \SI{3}{\meter\per\second} in the \textit{SLAM} mode. These results clearly show the robustness of the algorithm in the competing environment. 

\section{Control}\label{sec:control}

After the first successfully finished lap, the complete track is mapped using the SLAM algorithm (see \Cref{sec:slam}). Thus, the car now knows the track layout and can localize itself within the environment. Given this capability, we can race the car around a known track. Which brings us to our motion planning problem where the goal is to drive around the track as fast as possible. For this task, we extend the nonlinear MPC formulation of \cite{liniger_scale_rc_cars_2015}. The formulation aims to maximize the progress along a given reference path (in our case the center line) while respecting the vehicle model and track constraints. The two main advantages of the proposed MPC is the direct consideration of the vehicle limits when computing the command and that the algorithm does not need any pre-determined logic, only the track layout, and the vehicle model. Note, that we extend the formulation of \cite{liniger_scale_rc_cars_2015} in several ways, first, by using a novel vehicle model that is well behaved and accurate also when driving slow. Second, by including simple actuator dynamics and finally by including tire friction constraints. We also directly solve the nonlinear program instead of relying on a convex approximation scheme. 

In this section we first present our novel vehicle model which is suited for slow and fast driving while being tractable in a real-time MPC framework, furthermore, we explain the MPC formulation introduced in \cite{liniger_scale_rc_cars_2015}, with a focus on parts which are novel for this paper. 

\subsection{Vehicle Model} \label{sec:vehicle_model}
\begin{figure}[ht]
    \centering
    \def\svgscale{1.2}
    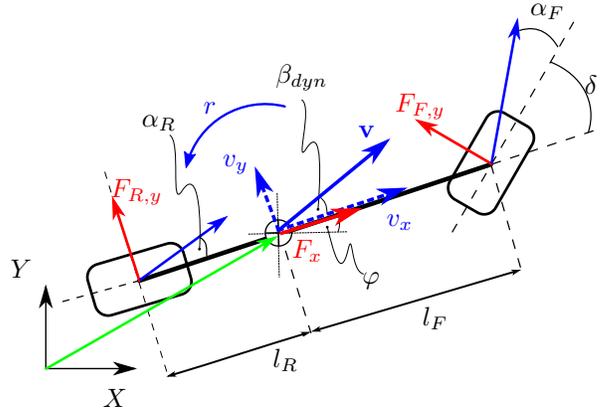
    \caption{Bicycle Model: Position vectors are in depicted in green, velocities in blue, and forces in red.}
    \label{fig:vehicle_model}
\end{figure}

The task of driving a vehicle at its operational limits is challenging due to the highly nonlinear behavior in this operation range. Therefore, we model the dynamics of our racecar as a dynamic bicycle model with nonlinear tire force laws. The model is able to match the performance of the car even in racing conditions, while at the same time being simple enough to allow the MPC problem to be solved in real-time. The used vehicle model is derived under the following assumptions, (i) the vehicle drives on a flat surface, (ii) load transfer can be neglected, (iii) combined slip can be neglected, and (iv) the longitudinal drive-train forces act on the center of gravity. The last three assumptions are valid since the used low-level controllers  are designed to deal with these effects. The resulting model is visualized in \Cref{fig:vehicle_model}. The equation of motion is given by,
\begin{align}
       \begin{bmatrix}
        \dot{X} \\
        \dot{Y} \\
        \dot{\varphi} \\
        \dot{v}_x \\
        \dot{v}_y \\
        \dot{r}
        \end{bmatrix}  = 
        \begin{bmatrix}
        v_x \cos{\varphi} - v_y \sin{\varphi} \\
        v_x \sin{\varphi} + v_y \cos{\varphi} \\
        r \\
        \frac{1}{m}(F_{R,x}- F_{F,y}\sin{\delta}+m v_y r) \\
        \frac{1}{m}(F_{R,y}+ F_{F,y}\cos{\delta}-m v_x r) \\
        \frac{1}{I_z}(F_{F,y}l_F\cos{\delta} - F_{R,y}l_R + \tau_{\text{TV}})\\
        \end{bmatrix}, 
\label{eq:vehicle_model}
\end{align}
where the car has a mass $m$ and an inertia $I_z$, $l_R$ and $l_F$ represent the distance form the center of gravity to the rear and the front wheel respectively, $F_{R,y}$ and $F_{F,y}$ are the lateral tire forces of the rear/front wheel, $F_x$ is the combined force produced by the drive-train and $\tau_{\text{TV}}$ the additional moment produced by the torque vectoring system. The state of the model $\tilde{\mathbf{x}} = [X,Y,\varphi ,v_x,v_y,r]^T$, consists of $(X,Y)$ and $\varphi$ the position and heading in a global coordinate system, as well as the $(v_x,v_y)$ the longitudinal and lateral velocities, and finally the yaw rate $r$. The control inputs $\tilde{\mathbf{u}} = [\delta, D]^T$ are the steering angle $\delta$ and driving command $D$. The driving command replicates the pedals of a driver and $D = 1$ corresponds to full throttle and $D = -1$ to full braking. We denote this model as $\dot{\tilde{\mathbf{x}}} = \tilde{f}_{\text{dyn}}(\tilde{\mathbf{x}}, \tilde{\mathbf{u}})$.

The forces $F_{a,y}$ represent the interaction between tires and track surface where the subscript $a$ refers to the front or rear tires, $a \in \{F,R\}$. A simplified Pacejka tire model \cite{pacejka1992magic} was chosen, which meets the trade-off between precision and computational requirements,
\begin{align}
\alpha_R &= \arctan{\Big(\frac{v_y - l_R r}{v_x}\Big)} \, , & \alpha_F &= \arctan{\Big(\frac{v_y + l_F r}{v_x} \Big)} - \delta \label{eq:slip_angles} \, ,\\
F_{R,y} &= D_R \sin{\Big(C_R \arctan{\Big(B_R\alpha_R\Big)}\Big)}\, , & F_{R,y} &= D_F \sin{\Big(C_F \arctan{\Big(B_F\alpha_F\Big)}\Big)}  \nonumber\, ,
\end{align}
where $B_a,C_a, D_a, a \in \{R,F\}$ are experimentally identified coefficients of the model and $\alpha_a,a \in \{R,F\} $ are the rear and front slip angles.

In addition to the lateral tire models, a drive-train model \eqref{eq:drive_train} describes the longitudinal force $F_x$ acting on the car depending on the applied driver command, $D \in [-1,1]$,
\begin{align} \label{eq:drive_train}
F_{x} &= C_{m} D - C_{r0} - C_{r2} v_x^2  \, ,
\end{align}
which consists of a motor model, $C_{m} D $, rolling resistance, $C_{r0}$, and drag, $ C_{r2} v_x^2$. The parameters $C_a, a \in \{m,r0,r2\}$ are identified from experiments. Note that this simplistic model is valid since the low-level controller takes care of the motor torque distribution to each wheel.

The final modelling component is the torque vectoring moment $\tau_{\text{TV}}$. The moment is generated by the low-level controller distributing the requested force independently to the four motors. By doing so, an additional torque acting on the car can be generated. To include this effect the logic of the low-level controller is included in the model \eqref{eq:vehicle_model}. The low-level controller, based on \cite{milliken1995race}, was developed for human drivers and is designed to mimic a kinematic car in order to achieve a more predictable behaviour. More precisely, the yaw moment is determined by a proportional controller, which acts on the error between the kinematic yaw rate $r_{\text{target}}$ and the measured yaw rate $r$. Thus the torque vectoring yaw moment $\tau_{\text{TV}}$ is given by,
\begin{align}
r_{\text{target}} &=  \delta \frac{v_x}{l_F+l_R} \, ,\\ 
\tau_{\text{TV}} &= (r_{\text{target}} - r) P_{\text{TV}}\, , \label{eq:TV}
\end{align}
where a small angle assumption on $\delta$ is used, and $P_{\text{TV}}$ is the proportional gain of the low level torque vectoring controller.

One problem of the dynamical bicycle model \eqref{eq:vehicle_model} is that the model is ill-defined for slow velocities due to the slip angles \eqref{eq:slip_angles}. However, slow velocities are important for race start and in sharp corners. For slow driving normally kinematic models are used which do not depend on slip angles. However, kinematic models are not suited for fast driving as they neglect the interaction of the tires and the ground. Therefore, to get the best of both models within one formulation we propose a novel vehicle model combining a dynamic and a kinematic vehicle model. To this end, we first formulate the kinematic model using the state of the dynamic model $\tilde{\mathbf{x}}$. The accelerations of the kinematic model are obtained through the differentiation of $v_{y,\text{kin}} = l_R r_{\text{target}}$ and $r_{\text{kin}} = \tan{(\delta)} \frac{v_x}{l_F+l_R}$, resulting in the following dynamics,

\begin{align}
        \begin{bmatrix}
        \dot{X} \\
        \dot{Y} \\
        \dot{\varphi} \\
        \dot{v}_{x} \\
        \dot{v}_{y} \\
        \dot{r}
        \end{bmatrix} = 
        \begin{bmatrix}
        v_x \cos{\varphi} - v_y \sin{\varphi} \\
        v_x \sin{\varphi} + v_y \cos{\varphi} \\
        r \\
        \frac{F_x}{m} \\
        (\dot{\delta} v_x + \delta \dot{v}_x)\frac{l_R}{l_R + l_F} \\
        (\dot{\delta} v_x + \delta \dot{v}_x)\frac{1}{l_R + l_F}\\
        \end{bmatrix} \, , 
\label{eq:kin_vehicle_model}
\end{align}
where we assumed $\delta$ to be small, what simplifies the following terms, $\cos{(\delta)}^2 \approx 1$ and $\tan{(\delta)} \approx \delta$. The resulting model \eqref{eq:kin_vehicle_model} is denoted as $\dot{\tilde{\mathbf{x}}} = \tilde{f}_{\text{kin}}(\tilde{\mathbf{x}}, \tilde{\mathbf{u}},\dot{\tilde{\mathbf{u}}})$. Thus, both models are formulated using the same states  which allows us to combine them. The resulting vehicle model is generated by linearly blended the two models, as follows
\begin{align}
    \dot{\tilde{\mathbf{x}}} &= \lambda \tilde{f}_{\text{dyn}}(\tilde{\mathbf{x}}, \tilde{\mathbf{u}}) + (1-\lambda) \tilde{f}_{\text{kin}}(\tilde{\mathbf{x}}, \tilde{\mathbf{u}},\dot{\tilde{\mathbf{u}}}) = \tilde{f}(\tilde{\mathbf{x}}, \tilde{\mathbf{u}},\dot{\tilde{\mathbf{u}}})\, , \nonumber \\
    \lambda &= \text{min} \left( \text{max} \left( \frac{v_x - v_{x,\text{blend min}}}{v_{x,\text{blend max}} - v_{x,\text{blend min}}},0\right),1\right) \, \label{eq:kindynmodel}.
\end{align}
The models are only combined in the velocity range of $v_x \in [v_{x,\text{blend min}},v_{x,\text{blend max}}]$; below we use purely the kinematic model while for velocities above $v_{x,\text{blend max}}$ we use purely the dynamic model. For our car we determined that this velocity range is from $v_{x,\text{blend min}} = 3$ to $v_{x,\text{blend max}} = 5$\,m/s.

Note that, the model now depends on the derivatives of the inputs $\dot{\tilde{\mathbf{u}}}$. We propose to use input dynamics to incorporate the input derivatives in the model. More precisely, we adopt the so-called $\Delta$ formulation in the MPC and assume that the inputs cannot be controlled directly but only their derivatives. Thus, the input dynamics are given by $\dot{D} = \Delta D$ and $\dot{\delta} = \Delta \delta$ which we summarize as $\dot{\tilde{\mathbf{u}}} = \Delta \tilde{\mathbf{u}}$, where $\Delta \tilde{\mathbf{u}} = [\Delta D, \Delta \delta]^T$ are the new control inputs. Thus the new model \eqref{eq:kindynmodel} is now properly defined. Furthermore, the input dynamics approximately model the real behavior of the steering system and the drivetrain, as well as allow to naturally constrain and penalize the input rates, among others implementing  the rate limit of the steering system described in Section \ref{sec:main concept}.

\subsubsection*{Tire Constraints}
One of the assumptions in the dynamic bicycle model \eqref{eq:vehicle_model}, is that the combined slip can be neglected. Combined slip occurs if the tire has to transform lateral as well as longitudinal forces. In such a case a real tire cannot produce the same maximum grip as in pure conditions. Simply said, a tire has a certain elliptic force budget it can transfer to the ground, often called the friction ellipse. This budget can be introduced as a constraint without explicitly modeling combined slip, which would require additional states. Thus, an ellipsoidal constraint on motor force at each tire $F_{R,x} = F_{F,x} = 0.5 C_m D$ (assuming equal front-rear force splitting) and $F_{a,y}$ is enforced,
\begin{align}
F_{R,y}^2 + (p_{\text{long}} F_{R,x})^2 &\leq (p_{\text{ellipse}} D_R)^2 \, , \nonumber \\
F_{F,y}^2 + (p_{\text{long}} F_{F,x})^2 &\leq (p_{\text{ellipse}} D_F)^2 \label{eq:friction_ellipse}\, ,
\end{align}
where $p_{\text{long}}$ and $p_{\text{ellipse}}$ are the tire specific ellipse parameters. The shape of the ellipse can greatly influence the driving style, e.g., with lower $p_{\text{long}}$ the vehicle is allowed to corner more while accelerating, whereas with higher $p_{\text{ellipse}}$ the tires are allowed to go closer to the limit. 

These tire friction constraints are meant to ensure that the tire budget is not violated.  Additionally, the low-level traction controller, which is based on \cite{Bohl2014model}, ensures the motor torques are correctly distributed also considering load changes. Therefore, the combination of the tire constraints and the low-level controller make sure  that the model assumptions are valid.

\subsection{Contouring Formulation}

The goal of the contouring formulation is to follow a reference path as fast as possible, in our case the center line of a track parametrized by the arc length. A third order spline is used to describe the path, as it offers a fast way to evaluate any point along the contour $(X_{\text{ref}}(\theta), Y_{\text{ref}}(\theta))$. To follow the path, the position of the car $(X,Y)$ has to be linked to the position on the path, or in other words the arc-length. We call this arc-length $\theta_\mathcal{P}$ and it can be computed by projecting the cars position $(X,Y)$ onto the reference path. However, computing the projection inside the MPC is computationally too expensive. We introduce $\theta$ to the model to approximates the true arc-length $\theta_\mathcal{P}$. To keep track of $\theta$ as well as the velocity and acceleration of the car relative to the reference path, a double integrator model is introduced, where $\dot \theta = v_\theta$ and $\dot{v}_\theta = \Delta v_\theta$. Note that the notation is used to highlight the similarity to the input dynamics of the vehicle model. 

\begin{figure}[ht]
	\centering
	\includegraphics[width=0.35\textwidth]{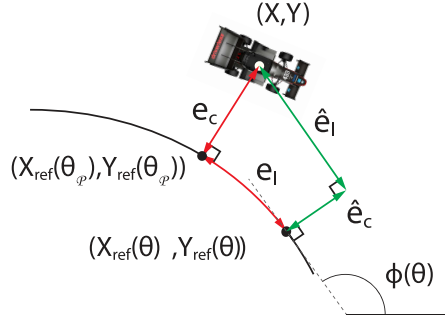}
	\caption{The approximate lead and lag errors with respect to the vehicle's position. }
	\label{fig:contouring}
\end{figure}

To ensure that $\theta$ is a good approximation of the true arc-length, a cost function is introduced which minimizes the approximation error along the path called the lag error $\hat{e}_l$ and the error perpendicular to the reference path called the contouring error $\hat{e}_c$ (see Figure \ref{fig:contouring}). Based on the geometry of the reference path the costs are given by, 
\begin{align}
    \hat{e}_c(X,Y,\theta) &=  \sin(\Phi(\theta))(X-X_{\text{ref}}(\theta)) - \cos(\Phi(\theta)(Y-Y_{\text{ref}}(\theta)) \, , \nonumber\\
    \hat{e}_l(X,Y,\theta) &=  -\cos(\Phi(\theta))(X-X_{\text{ref}}(\theta)) - \sin(\Phi(\theta)(Y-Y_{\text{ref}}(\theta))\, , \label{eq:cont_error}
\end{align}
with $\Phi(\theta)$ being the angle of the tangent to the reference path. Note that, if both costs are small $\theta$ is a good approximation of $\theta_\mathcal{P}$. 

Finally, to follow the reference path as fast as possible the velocity along the path $v_{\theta}$ is maximized, which corresponds to maximizing the progress over the horizon. Combining these three costs, the cost function of the contouring formulation is defined as,
\begin{align} \label{eq:contouring_cost}
    J(\mathbf{x}_k) = q_c \hat{e}_{c,k}^2 +  q_l \hat{e}_{l,k}^2 - \gamma v_{\theta,k}\, ,
\end{align}
where the individual objectives are weighted with $\gamma \geq 0$, $q_c \geq 0$ and $q_l \geq 0$. This allows to find the desired trade off between path following (minimize  $\hat{e}_{c}$ and $\hat{e}_{l}$, defined in \eqref{eq:cont_error}) and lap time (maximizing $v_{\theta}$). 

\subsubsection*{Track Constraints}
Finally, we consider track constraints, which limit the MPC to stay within the track boundaries. The constraint is formulated by locally approximating the track by circles centered on the middle line,
\begin{align}
(X - X_{\text{cen}}(\theta))^2 + (Y - Y_{\text{cen}}(\theta))^2 \leq R_{\text{Track}}(\theta)^2. \label{eq:track_con}
\end{align}
Compared to the linear constraints used in \cite{liniger_scale_rc_cars_2015}, these circular constraints are a convex inner approximation of the track, which increases the safety of the formulation. To reduce the computation time the position of the circles ($X_{\text{cen}},Y_{\text{cen}}$) is determined given the previous solution of the MPC problem, by fixing $\theta$.

\subsection{MPC Problem Formulation} \label{sec:res_prob_form}
In summary, the dynamics of the problem are given by the vehicle model \eqref{eq:kindynmodel}, the input dynamics, and the arc-length dynamics. To summarize them in one system, we define $\mathbf{u} = [\delta, D, v_\theta]^T$ and $\Delta \mathbf{u} = [\Delta\delta,\Delta D,\Delta v_\theta]^T$. This allows to formulate the full model used in the MPC as,
\begin{align}
    \dot{{\mathbf{x}}} = \begin{bmatrix}
    \dot{\tilde{\mathbf{x}}} \\ 
    \dot{\theta} \\
    \dot{\mathbf{u}}
    \end{bmatrix}
     = 
    \begin{bmatrix}
    \tilde{f}(\mathbf{\tilde{\mathbf{x}}},\mathbf{\tilde{\mathbf{u}}},\Delta \tilde{\mathbf{u}}) \\
    v_\theta \\
    \Delta\mathbf{u}
    \end{bmatrix} 
    = f({\mathbf{x}},\Delta\mathbf{u}) \, ,
    \label{eq:final_model}
\end{align}
giving rise to a model with 10 states and 3 inputs. It is important to note, that in this formulation the control input is $\Delta \mathbf{u}$ and the steering, driving commands as well as $v_\theta$ are states. This final system of differential equations \eqref{eq:final_model} is discretized with Runge-Kutta 4th order integrator, resulting in the discrete time system ${\mathbf{x}}_{k+1} = f(\mathbf{x}_{k},\Delta\mathbf{u}_k)$. 

The MPC cost function consists of the previously defined contouring cost, $J_k(\mathbf{x}_k)$, as well as a control input penalization
\begin{align*}
R(\mathbf{u}_k,\Delta \mathbf{u}_k) = \mathbf{u}^T_k\mathbf{R}_\mathbf{u}\mathbf{u}_k + \Delta\mathbf{u}^T_k\mathbf{R}_{\Delta \mathbf{u}}\Delta\mathbf{u}_k.
\end{align*}
In addition to that we introduce a side slip angle cost which controls the aggressiveness of the MPC, which is defined as,
\begin{align*}
    L(\mathbf{x}_k) = q_\beta(\beta_{\text{kin},k} - \beta_{\text{dyn,k}})^2 \, ,
\end{align*}
where the squared error between the dynamic side slip angle, $\beta_\text{dyn}=\arctan{\left(v_y /v_x \right)}$, and the kinematic side slip angle $\beta_\text{kin}=\arctan{\left(\tan{(\delta)}l_R/(l_R + l_F) \right)}$ is penalized with the weight $q_\beta \geq 0$, . 

The inequality constraints consist of track \eqref{eq:track_con}, tire \eqref{eq:friction_ellipse}, input, and input rate constraints. The track constraints \eqref{eq:track_con} are softened with a slack variable, $S_c$, and the corresponding cost $C(S_{c,k}) = q_s S_{c,k} + q_{ss} S_{c,k}^2$ is added, to guarantee feasibility at all times.  

The final MPC problem with a prediction horizon of $N$ looks as follows,
\begin{align} 
\min_{\Delta \mathbf{u}} \hspace{0.5cm} & \sum_{k = 0}^N J(\mathbf{x}_k) + R(\mathbf{u}_k,\Delta \mathbf{u}_k) + L(\mathbf{x}_k) + C(S_{c,k}) \nonumber\\
s.t .    \hspace{0.5cm} & \mathbf{x}_0 = \mathbf{x}(0),\nonumber \\
                        &  \mathbf{x}_{k+1} = f(\mathbf{x}_k,\Delta\mathbf{u}_k),\nonumber \\
                        &  (X_k - X_{\text{cen},k})^2 + (Y_k - Y_{\text{cen},k})^2  \leq R_{\text{Track},k}^2 + S_{c,k},\nonumber \\ 
                        &  F_{R,y,k}^2 + (p_{\text{long}} F_{R,x,k})^2 \leq (p_{\text{ellipse}} D_R)^2,\nonumber \\
                        &  F_{F,y,k}^2 + (p_{\text{long}} F_{F,x,k})^2 \leq (p_{\text{ellipse}} D_F)^2,\nonumber \\
                        &  \mathbf{u}_{\text{min}} \leq \mathbf{u}_k \leq \mathbf{u}_{\text{max}},\nonumber \\
                        &  \Delta \mathbf{u}_{\text{min}} \leq \Delta \mathbf{u}_k \leq \Delta \mathbf{u}_{\text{max}}. \label{eq:MPC}
\end{align}
The MPC problem \eqref{eq:MPC}, is a nonlinear optimization problem which is solved using the commercial solver ForcesPro NLP \cite{FORCESnlp,FORCESPro} in a receding horizon fashion. For our experiments we use a sampling time of $\SI{50}{\milli\second}$ and a prediction horizon of 40 time steps, which corresponds to a look-ahead of $\SI{2}{\second}$. The predicted horizon of the MPC with corresponding track constraints can be seen in \Cref{fig:mpc_rviz}. One can notice that the predicted horizon is longer than the observed cones, which validates our need of a SLAM map. Furthermore, it can be seen that the MPC is driving a race line and does not only follow a middle line. More results of the achieved driving performance are given in \Cref{sec:overall_system_results}.

\subsection{Runtime Analysis}
\begin{figure}[ht]
    \centering
    	\begin{subfigure}{0.45\textwidth}
        \centering
        \includegraphics[width=0.4\linewidth]{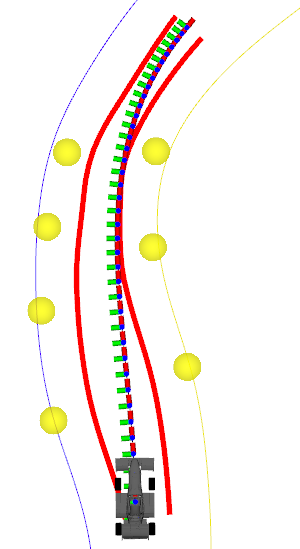}
        \caption{The predicted horizon of the optimization based controller shown with axes. The red lines represent maximal lateral constraints, blue and yellow line the right and left boundary respectively, and yellow larger circles the current observed cones.}
        \label{fig:mpc_rviz}
    \end{subfigure}
	\hspace{1cm}
	\begin{subfigure}{0.45\textwidth}
        \centering
        \setlength{\figureheight}{5cm}
        \setlength{\figurewidth}{7cm}
%
\definecolor{mycolor1}{rgb}{0.00000,0.44700,0.74100}%
\begin{tikzpicture}

\begin{axis}[%
width=0.951\figurewidth,
height=\figureheight,
at={(0\figurewidth,0\figureheight)},
scale only axis,
xmin=0.015,
xmax=0.077,
ymin=0.000,
ymax=1000.000,
xlabel style={font=\color{white!15!black}},
xlabel={$t$[s]},
xticklabel style={
        /pgf/number format/fixed,
        /pgf/number format/precision=2
},
scaled x ticks=false,
axis background/.style={fill=white},
title style={font=\bfseries}
]
\addplot[ybar interval, fill=mycolor1, fill opacity=0.600, draw=black, area legend] table[row sep=crcr] {%
x	y\\
0.018	2.000\\
0.020	21.00\\
0.022	147.000\\
0.024	474.000\\
0.026	785.000\\
0.028	836.000\\
0.030	832.000\\
0.032	668.000\\
0.034	551.000\\
0.036	378.000\\
0.038	258.000\\
0.040	276.000\\
0.042	253.000\\
0.044	230.000\\
0.046	134.000\\
0.048	88.000\\
0.050	139.000\\
0.052	321.000\\
0.054	133.000\\
0.056	69.000\\
0.058	52.000\\
0.060	26.000\\
0.062	9.000\\
0.064	5.000\\
0.066	2.000\\
0.068	4.000\\
0.070	0.000\\
0.072	1.000\\
0.074	1.000\\
};

\addplot [color=red]
  table[row sep=crcr]{%
0.050	0.000\\
0.050	1000.000\\
};

\node[right, align=left, font=\color{red}]
at (axis cs:0.051,600) {$T_s= 0.05$s};

\draw[black,-{Latex[width=3mm]}] (axis cs:0.05,800) -- (axis cs:0.04,800);
\node[right, align=left, font=\color{black}]
at (axis cs:0.041,880) {$90\%$};

\end{axis}
\end{tikzpicture}%
        \caption{Solve time histogram of the used optimization problem.}\label{fig:solvetime}
    \end{subfigure}
  \caption{Evaluation of used Model Predictive Control technique.}
\label{fig:mpc_experiment}
\end{figure}
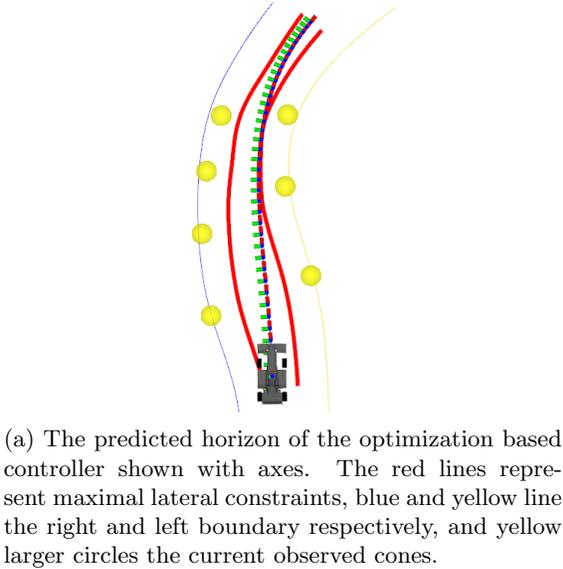
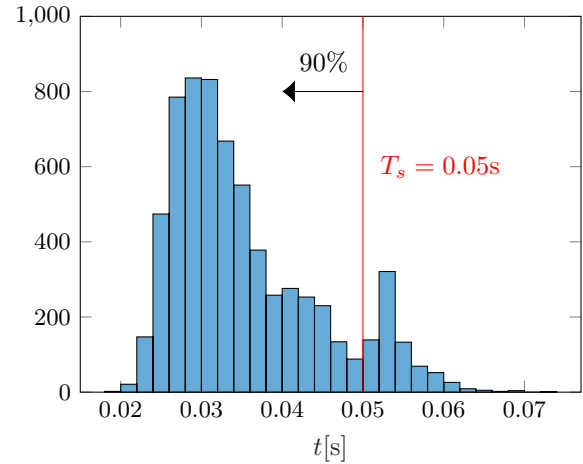

Since the used control technique is optimization based, one of the important factor is the solve time. The histogram of solve times achieved during the FSG trackdrive competition can be seen in \Cref{fig:solvetime}. We can see that $90\%$ of the solve times are below the sampling time, and at most, it takes $\SI{0.065}{\second}$ to solve the nonlinear optimization problem. Due to the soft real-time structure of ROS, computation times above the sampling time do not cause problems, and no performance impact could be noticed in case of over times. 

\section{Testing Framework }\label{sec:testing_framework}

To ensure that all modules work as designed, we formalized our testing procedures and created several tools to verify the correctness of our implementations. We made extensive use of automated simulations to test our code, as described in \Cref{sec:simulation}. In addition, both the simulation and testing generates considerable amounts of data. Our approach to handle the large number of logs and maximize its value is described in \Cref{sec:data-analysis}. We call the framework Automated Testing System\footnote{\url{https://github.com/AMZ-Driverless/rbb_core}} (ATS). 

\subsection{Simulation} \label{sec:simulation}

Real-world testing with the racecar requires time and manpower what drastically limits the number of tests that can be performed. Simulation is used to catch problems before they happen on the race track and thus increase the efficiency of the actual testing. Simulations are ran automatically, triggered by pull requests into our main active code branch and every night several tests are run in simulation.

The used simulator is FSSIM\footnote{\url{https://github.com/AMZ-Driverless/fssim}} which is built upon the popular Gazebo software package \cite{koenig2004design}. FSSIM however does not make use of the Gazebo's included physics engine. Instead, an integrated first principle model is used. This allows us to  match the simulation to the actual vehicle performance. The simulated race track surface is approximated as a flat plane with constant friction and can be automatically generated from log data. Perception sensors are not simulated because of large computational requirements.  Instead, direct cone observations with various noise models are simulated. Simulation results are post processed to give a binary pass/fail and further lap statistics, i.e. lap times. Logged data is uploaded to the data analysis system and immediately processed into a set of visualizations (see \Cref{sec:data-analysis}). In case of a failure, notifications are sent out. This reduces the iteration time and increases the software quality.

\subsection{Data Analysis \& Visualization}\label{sec:data-analysis}

\begin{figure*}[ht]
    \centering
	\begin{subfigure}{0.45\textwidth}
        \includegraphics[width=\textwidth]{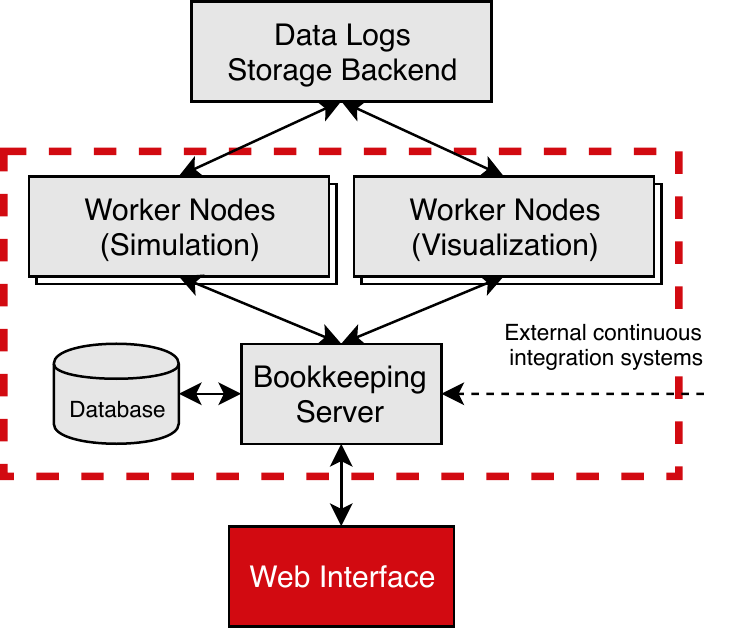}
        \caption{Overview of the architecture of the automated testing framework. Communication is marked with arrows.}
        \label{fig:ats-overview}
    \end{subfigure}
	\hspace{1cm}
	\begin{subfigure}{0.45\textwidth}
       \includegraphics[width=\textwidth]{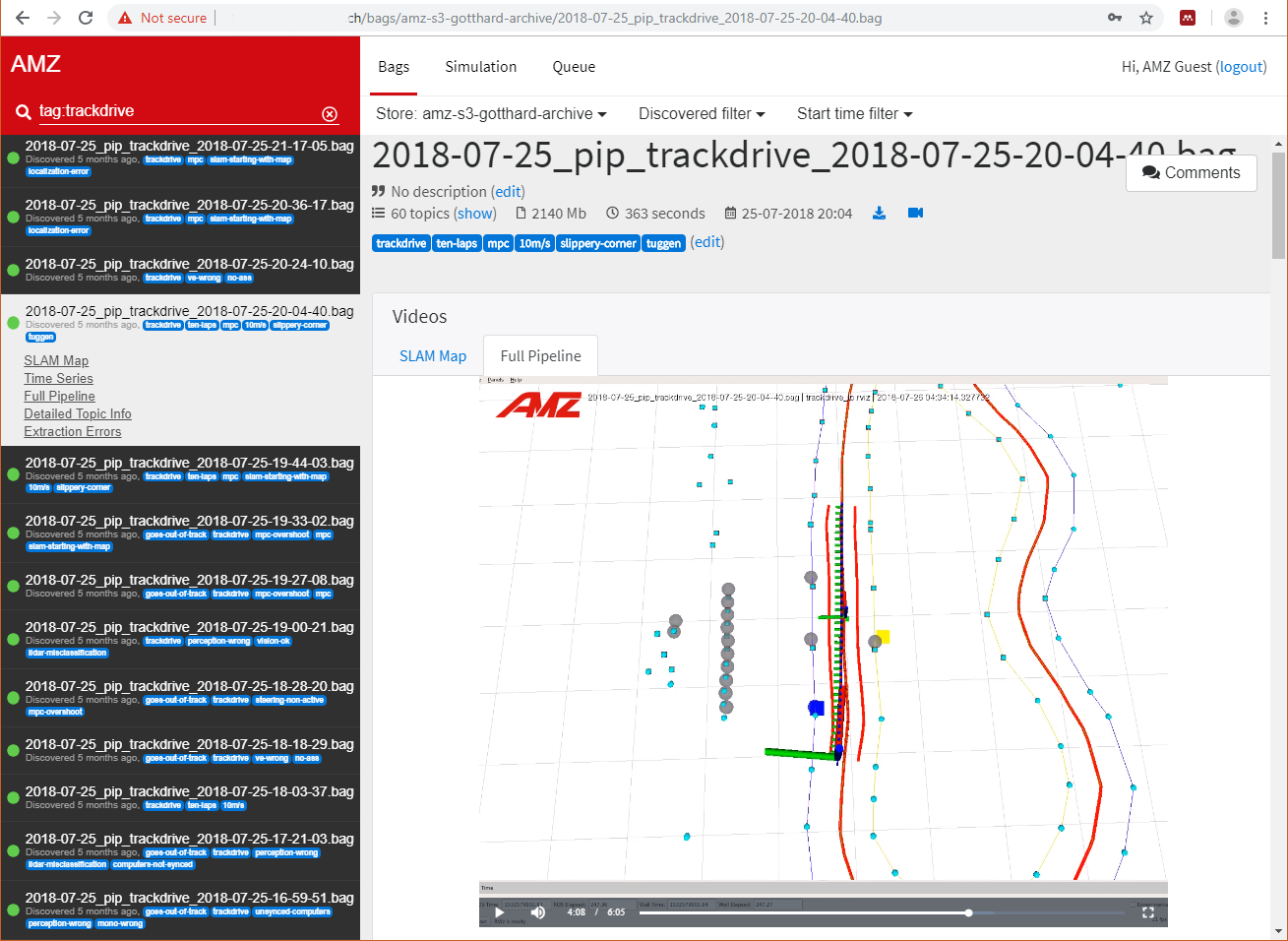}
       \caption{Web-based visualization is easily accessible from everywhere.}
       \label{fig:ats-screenshot}
    \end{subfigure}
  \caption{Automated Testing System (ATS) overview.}
\end{figure*}

During the simulation and testing of the car, large amounts of data are recorded. The archiving, browsing and distribution of these data becomes increasingly difficult due to their volume and size. We designed a framework that automatically visualizes and indexes the logs. The framework is accessed through a web interface, shown in \Cref{fig:ats-screenshot}, which enables everyone to check results quickly. Since the software stack is built around ROS, the popular rosbag format is used for the data logs.

The framework is built in a distributed fashion with architecture depicted in \Cref{fig:ats-overview}. At the heart is a service that does the bookkeeping for all data logs and stores this in a PostgreSQL database keeping a queue of all tasks that need to be performed. These tasks are further distributed to worker nodes. Worker nodes poll the service to request scheduled tasks. This is preferred over a push based system to keep the system simple and stateless. The queue is made for a small number of long running tasks and not optimized for many small short running operations, so polling delay is negligible. The main tasks are simulation (as described in \Cref{sec:simulation}) indexing and visualization.

Visualizations are either manually queued or scheduled upon discovery of new data logs. Each data log can contain different sets of data depending on where it was recorded and which tests were performed. To accommodate this, a matching system is used where sets of data topics are matched against pre-configured plugin setups. For the most important 3D visualization plugin we heavily rely on RViz \footnote{\url{http://wiki.ros.org/rviz}}. The output on a virtual screen is recorded while the logs are being played back. This is a powerful primitive that can be used with any existing visualization software that needs to be automated. Each plugin can output a data object visible in the web interface (e.g., a set of images with plots or a recorded video).

\section{The Race Results}\label{sec:overall_system_results}

\begin{figure*}[ht]
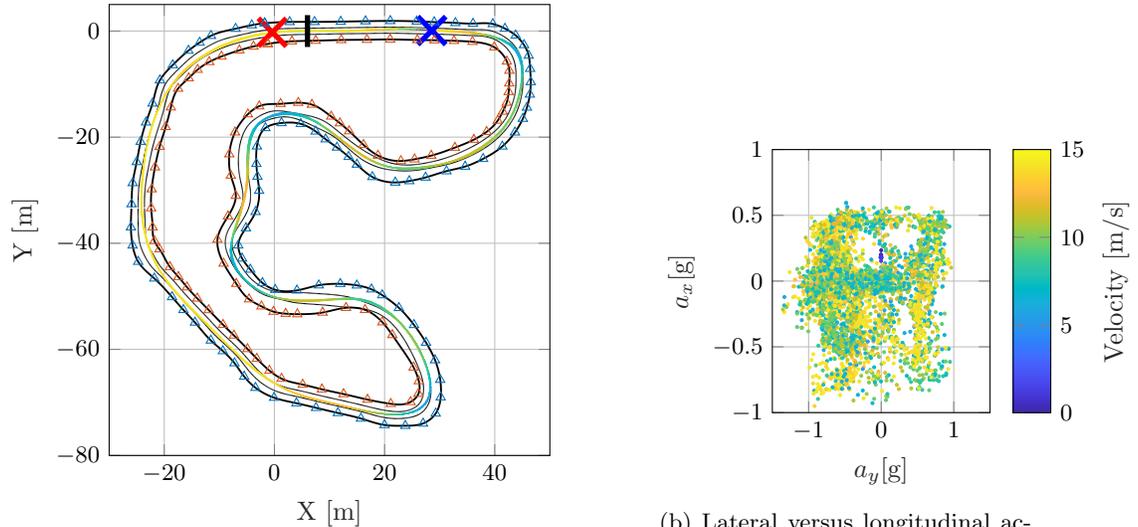

    \centering
	\begin{subfigure}{0.6\textwidth}
        \centering
    	\setlength{\figureheight}{6cm}
        \setlength{\figurewidth}{7cm}
        \input{trackdrive.tex}
        \caption{The driven line from the FSG 2018 competition with corresponding cone positions. The red cross represents the starting position, the blue cross the ending position, and the black thick line the starting line. From the driven path one can see that the vehicle cuts corners while driving slower in corners and faster on straights, as a human driver. Color of the trajectory indicates the velocity of the racecar. }
        \label{fig:fsg_trackdrive_driven_line}
    \end{subfigure}
	\hspace{-0.1cm}
	\begin{subfigure}{0.3\textwidth}
        \centering
        \setlength{\figureheight}{3.5cm}
        \setlength{\figurewidth}{3.5cm}
        \input{gg.tex}
        \caption{Lateral versus longitudinal accelerations measured at the CoG (GG plot) over ten laps. }
        \label{fig:gg}
    \end{subfigure}
  \caption{The results achieved during FSG 2018 trackdrive discipline. }
\end{figure*}

As final results, we show the performance achieved during $10$ consecutive laps at the FSG 2018 competition. These can be seen in the \Cref{fig:fsg_trackdrive_driven_line}. The starting position is indicated by the red cross, located $\SI{6}{\meter}$ before the start line. The vehicle autonomously drove $10$ laps and successfully completed the event by coming to a safe stop (blue cross) within the allowed $\SI{20}{\meter}$ after the starting line. One can see in the shape of the driven path that the car slows down in sharper turns and accelerates at the exit of the turns, similarly to a human driver. Over the course of this run, the vehicle achieved $1.5$g lateral acceleration which is close to the tire limit. The GG plot of the ten laps can be seen in \Cref{fig:gg}. The squared shape is caused by conservative torque limits, which emphasis safety over pure performance. As illustrated in \Cref{tab:lap_times}, the used optimization-based control technique results in fast and consistent lap times.

\begin{table}[ht]
\begin{tabular}{ c|c|c|c|c|c|c|c|c|c|c } 
\toprule
\textbf{Lap Number} & \textbf{1} & \textbf{2} & \textbf{3} & \textbf{4} & \textbf{5} & \textbf{6} & \textbf{7} & \textbf{8} & \textbf{9} & \textbf{10}  \\
\midrule
\textbf{Lap Time} [s] & $28.913$ & $28.608$ & $28.558$ & $28.483$ & $28.625$ & $28.542$ & $28.523$ & $28.571$ & $28.512$ & $28.593$ \\
\end{tabular}
\caption{Achieved lap times on the FSG 2018 competition. Thanks to the optimization nature of the controller the lap times are very consistent. }
\label{tab:lap_times}
\end{table}

\subsection{Lessons Learned}

Over the course of the past two years, the design of our autonomous system grew significantly. However, our design principles largely stayed the same. Parts of these were brought in from previous projects, others resulted from failures. We would therefore like to share these with the reader.

First, we learned that hardware is important, and only real-world testing can thoroughly evaluate the whole system. Thus, testing is the most vital factor for success and it is therefore paramount to keep the development and testing cycles short in order to complete and evaluate features well in time for the competition. Simulation testing is also invaluable, as it allows to make real-world testing more efficient. This realization led to the development of the ATS (\Cref{sec:testing_framework}). Its ability to simulate multiple trials automatically made possible to find good parameters in simulation and only fine tune these at the testing spot. Moreover, ATS' powerful analysis tool helped us to quickly identify problems in both real-world and simulation testing.

Finally, when testing one needs to pay attention to extreme cases. For a racecar this means pushing the car to the limit, using aggressive driving, challenging track layouts, all during good and poor weather conditions. This, helps to better understand the car and allows us to successfully compete in competitions where it is key to go to the limit but not beyond. 

Having a software architecture with fixed, clearly defined interfaces makes it possible to develop, simulate, and test each subsystem on datasets independently. However, the full pipeline should still be tested regularly to verify that it works as a whole and that overall computational constraints are met.

Since the racecar can accelerate from $0$ to $100$km/h in under $2$s and drives at significant speeds, safety is our primary concern when testing. We therefore developed a \textit{safety calculator} to estimate the necessary distance between the track and obstacles, considering the racecar's maximum speed, human reaction time, and even communication delays. The calculated safety margins are used when building test tracks during our testing.

\begin{figure}[ht]
    \centering
    \includegraphics[width=0.3\textwidth]{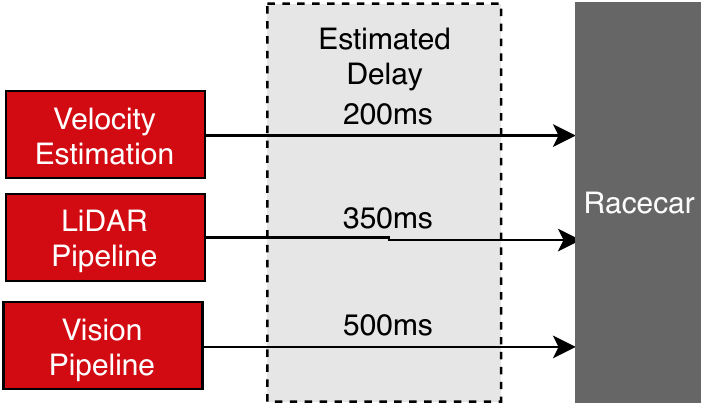}
    \caption{Total delay diagram of the proposed software-hardware architecture.}
    \label{fig:delay}
\end{figure}

Finally, one of the biggest bottlenecks in our current architecture are delays, as shown in \Cref{fig:delay}. The time it takes from a cone detection until a corresponding control command reaches the car sums up to about \SI{300}{\milli\second}. When driving at the limits of handling, this can lead to severe instabilities e.g., the car missing a braking point what might result in leaving the track. Our current workaround is to limit our top speed and wheel torques, which causes us to fall behind human performance.

\section{Conclusion}\label{sec:conclusion}

In this paper we showed that our autonomous racecar design enables autonomous driving on unknown tracks, create a map of them, and further drive faster after loop closure detection. The system was extensively tested and then used in FSD competition. In 2017, out of 15 teams attending the first FSG competition, ours was the only one able to finish all the disciplines. Furthermore, in 2018 we competed in FSI as well as FSG where our system outperformed our previous results and won both competitions beating numerous score records. The results showed the capabilities and scalability of the proposed software stack as well as the potential of the created tools. 

At the time of writing, a new AMZ Driverless racecar is being designed and prepared for the 2019 season. The overall goal of the new team is to match the performance of a human driver. Therefore, the general focus lies on reducing and handling delays in a consistent way throughout the software stack. Furthermore, the perception group's primary aim is to increase the observation radius. The estimation group is working on detecting the track boundaries more reliably and over a longer horizon, creating globally consistent maps, and adding the ability to localize globally. In controls, the focus lies on solving the problem more efficiently, such that more accurate models can be used online, and on rigorous system identification. The software architecture group is working on detecting and recovering from subsystem failures online, as well as automating the testing process further.
Formula Student Driverless is a unique opportunity for engineering students to learn about complex robotic systems and to deploy one under pressure in a competitive environment, just like in industry.

\subsubsection*{Acknowledgments}\label{sec:acknowledgements}

In the name of AMZ Driverless, we would like to thank all our sponsors as we could not have realized this project without their support. Furthermore, we would like to thank Marc Pollefeys, Andrea  Cohen, and  Ian Cherabier from ETH Z\"urich's CVG Group, Dengxin Dai from ETH Z\"urich's CVL Group and Mathias B\"urki from ETH Z\"urich's ASL Group for the knowledge and helpful ideas that they contributed to the project.

\bibliographystyle{apalike}
\bibliography{bibliography}

\appendix

\section{Appendix}
\subsection{Observability Analysis}\label{sec:observability_analysis}

Observability analysis plays a pivotal role for design of a redundant filter. A system is said to be observable if the state $\mathbf{x}(t)$ can be fully determined (reconstructed) using measurements $\mathbf{z}(t)$ and control inputs $\mathbf{u}(t)$. If any sensor fails, it is important to understand if state estimate converges to the true value or not.

In case of non-linear models such as ours, the observability can be determined by performing the rank test on matrix $\mathcal{O}$ using Lie derivative \cite{cit:Herman_observe}, 
which is defined recursively as: 
\noindent\begin{minipage}{.5\linewidth}
\begin{equation}\label{eq:Liederivative}
L^{l+1}_{f} h (\mathbf{x}) = \frac{\partial L^{l}_{f} h}{\partial \mathbf{x}} f(\mathbf{x}, \mathbf{u})\, ,
\end{equation}
\end{minipage}%
\begin{minipage}{.5\linewidth}
\begin{equation}\label{eq:Observavilitymatrix}
\mathcal{O}(\mathbf{x}, \mathbf{u})=\left[ \frac{\partial L^{0}_{f} h (\mathbf{x}) }{\partial \mathbf{x}} , \, \frac{\partial L^{1}_{f} h (\mathbf{x})}{\partial \mathbf{x}}, \, ...   \right]^T\, ,
\end{equation}
\end{minipage}
with $L^{0}_{f} h (\mathbf{x}) = h(\mathbf{x})$. The Observability matrix is defined in \eqref{eq:Observavilitymatrix}.

The system is weakly locally observable if $\mathcal{O}$ is full rank, and not observable otherwise. By iteratively removing measurements and performing the observability analysis it can be concluded that to estimate the full 9 states, a velocity measurement is needed. If all velocity measurement fail, the model is converted from a full dynamic model to a partial kinematic one assuming slip ratio is close enough to zero. In practice this assumption holds for speeds under $\SI{6}{\meter\per\second}$.

\subsection{Slip Ratio Estimation} \label{sec:sr_equation}

The racecar wheels can slip upto 30\% at high accelerations and even higher under heavy braking. This leads to biased velocity estimates from wheel odometry. Wheel slip is therefore estimated and used to correct for misleading wheel odometry information.

The slip ratio is one of the measures to express slipping behavior of the wheel defined as
\begin{align}\label{eq:slipratio_equation}
\text{sr}_{{{ij}}} = \left\{
 	\begin{array}{l l}
 	\frac{\omega_{{{ij}}}  R - v_{{{ij}}} }{v_{{{ij}}}} & \quad \text{if $|v_{ij}|>0$}\\
 	\quad 0 & \quad \text{if $|v_{{{ij}}}|=0$}
 	\end{array} \right. \, , 
\end{align}

$\text{sr}_{{ij}}$ denotes the slip ratio of wheel $ij$, where i $\in \{\text{Front} ,\; \text{Rear} \}$ and j $\in \{\text{Left} ,\; \text{Right} \}$. Here, $\omega_{{{ij}}}$ and $v_{{{ij}}}$ are the angular and linear velocities of wheel $ij$ respectively. $R$ is the radius of the wheel.

Wheelspeed measurement updates are split into two components, linear hub velocity and slipratio in the measurement model given by \eqref{eq:slip_wheelspeed} and both the components are updated simultaneously using wheelspeed measurement in order to update slip ratio ratio in EKF. Here, $b$ denotes the distance between right and left wheels, $\delta$ is steering angle, $l_{F}$ and $l_{R}$ are distance of front and rear axle respectively from the car center of gravity. $r$ is the yaw rate of the car. This method requires the slip ratio to be a part of the state of the EKF. 

\begin{align}
\label{eq:slip_wheelspeed}
 \begin{bmatrix} \omega_{FL} \\ \omega_{FR} \\ \omega_{RL} \\ \omega_{RR} \end{bmatrix}
 =
  \begin{bmatrix}
   ((v_{x}-r  \frac {b}{2}) \, \cos(\delta)+(v_{y}+r \, l_{F}) \, \sin(\delta))  \frac {(sr_{FL}+1) }{R}\\
   ((v_{x}+r  \frac {b}{2}) \, \cos(\delta)+(v_{y}+r \, l_{F}) \, \sin(\delta))  \frac {(sr_{FR}+1) }{R}\\
   (v_{x}-r  \frac {b}{2})  \frac {(sr_{RL}+1) }{R}\\
   (v_{x}+r  \frac {b}{2}) \frac {(sr_{RR}+1)}{R}\\
   \end{bmatrix}
   +
   \mathbf{n_{\omega}}.
\end{align}

\subsection{FastSLAM 2.0 Algorithm Details}\label{sec:slam_algo_details}
The goal of the mapping phase is to create a 2D map of cones and to simultaneously localize within it. Each landmark $\mathbf{\lambda}_n$ consists of a Gaussian position estimate $[\mathbf{\mu}_n, \mathbf{\Sigma}_n]$, a color estimate $c_n \in \{ \text{yellow}, \text{blue}, \text{orange}, \text{unknown} \}$ and two counters $[n_{s,n}, n_{m,n}]$ which indicate how often the landmark is seen or not seen (missing) respectively. A map is then defined as a set of $N$ landmarks:
\begin{equation}
 \mathbf{\Lambda} = \big([\bm{\mu}_1, \bm{\Sigma}_1, c_1, n_{s,1}, n_{m,1}], ..., [\bm{\mu}_N, \bm{\Sigma}_N, c_N, n_{s,N}, n_{m,N}] \big) \, .
 	\label{eq:map}
\end{equation}
Each particle $S^\textit{m}_k$ at time step $\textit{k}$ with particle index $m \in 1, ..., M$ carries a map and a pose, both defined in a global frame which is initialized at the beginning of the mapping phase.

\subsubsection*{Observations and Data Association}

The map is updated when a new set of observations is received. 
For each particle, a new pose is sampled from a Gaussian distribution obtained using an EKF approach. 
The prior $[\bm{\mu}_{s_k,0}, \mathbf{\Sigma}_{s_k,0}]$ of this filter is the composition of the pose at the previous time step and the odometry estimate. 
Additionally, each observation's delay is compensated using the integrated odometry between the time of observation and the current time of the filter. Subsequently, the landmark observations are associated to already existing landmarks in the map using the nearest neighbour method. 
The Mahalanobis distance of an observation $\mathbf{z}_k$ to the Gaussian distribution of a landmark $\lambda_n$ is used as a measure for the likelihood of correlation. 
If the maximum likelihood of an observation correlating to any landmark is below a threshold $l$, the observation is assumed to belong to a new landmark~\cite{paper_fluela}.

\subsubsection*{Proposal Distribution and Map Update}

After data association, the pose proposal EKF is iteratively refined by incorporation of the matched observations as measurements~\cite{Montemerlo2007}:
\begin{align}
    \mathbf{\Sigma}_{s_k,n} &= (\mathbf{G}_{s,n}^T\mathbf{Z}_{n}^{-1}\mathbf{G}_{s,n}+\mathbf{\Sigma}_{s_k,n-1}^{-1})^{-1} \, , \\
    \bm{\mu}_{s_k,n} &= \bm{\mu}_{s_{k},{n-1}} + \mathbf{\Sigma}_{s_k,n}\mathbf{G}_{s,n}^T\mathbf{Z}_{n}^{-1}(\mathbf{z}_{k,n}-\mathbf{\hat{z}}_{n}) \, ,
\end{align}
where 
\begin{equation}
    \mathbf{Z}_{n} = \mathbf{R} + \mathbf{G}_{\theta,n}\mathbf{\Sigma}_{s_k,n-1}\mathbf{G}_{\theta,n}^T \, ,
\end{equation}
and $\mathbf{R}$ is the measurement noise, and $\mathbf{G}_{\theta,n}$ and $\textbf{G}_{s,n}$ are the Jacobians with respect to the landmark position and the particle pose respectively. A new pose is then sampled from the Gaussian distribution given by:
\begin{align}
    \mathbf{s}_k \sim \mathcal{N}(\bm{\mu}_{s_k},\mathbf{\Sigma}_{s_k})\, .
\end{align}
Subsequently, the landmark positions are updated in a straight forward way using the EKF equations. 
Observations that were classified as ``new landmarks'' are initialized at the respective location. The landmark color is described by a categorical distribution with K = 3 possible outcomes. 
Bayesian inference on the color is applied using a Dirichlet distribution as conjugate prior.
A discrete, counter based system updates the probabilities of each possible color for each landmark. 
This process happens separately for each sensor, and is tuned to suit the expected color accuracy of them.

\subsubsection*{Particle Weighting and Resampling}

Each particle is weighted according to how well its map is in coherency with new landmark observations~\cite{cit:gotthard-icra-paper}
\begin{equation} 
    w_{k} = {w_{k-1}l^{\nu}} {w_{b}^{\kappa}} {w_{c}^{\gamma}} {\prod_{n \in \mathbf{a}_k} w_{k,n}}\, ,
\end{equation}
where $w_{k-1}$ is the particle weight at the previous time step, $l$ the weight assigned to new landmarks and $\nu$ is the number of new landmarks, $w_{b}$ a penalty for landmarks that are in sensor range but were not observed and $\kappa$ is the number of not observed landmarks, $w_{c}$ penalizes a color mismatch for all $\gamma$ landmarks whom color estimate don't agree with the observations and
\begin{equation}
    w_{k,n} = \frac{1}{2 \pi \sqrt{\text{det}(\mathbf{L}_n)}}\exp{\left(-\frac{1}{2}(\mathbf{z}_{k,n}-\mathbf{\hat{z}}_{n})^T \mathbf{L}_{n}^{-1}(\mathbf{z}_{k,n}-\mathbf{\hat{z}}_{n})\right)}\, ,
\end{equation}
are the importance weights of all matched landmarks where $\mathbf{L}_n$ is the special innovation covariance incorporating both measurement and motion models as well as the landmark uncertainty~\cite{Montemerlo2007} and $\text{det}( \cdot )$ is the determinant operator.
The particle weight variance naturally increases over time and therefore resampling is enforced once the effective sample size $N_{\text{eff}}$ of the particles drops below a certain threshold, $N_{\text{eff}} < 0.6N$.

\end{document}